\documentclass{article}

\usepackage{microtype}
\usepackage{graphicx}
\usepackage{subcaption}
\usepackage{booktabs} 

\usepackage{hyperref}

\usepackage[preprint]{icml2026}

\usepackage{amsmath}
\usepackage{amssymb}
\usepackage{mathtools}
\usepackage{amsthm}

\usepackage[capitalize,noabbrev]{cleveref}

\theoremstyle{plain}

\theoremstyle{definition}

\theoremstyle{remark}

\usepackage[textsize=tiny]{todonotes}

\icmltitlerunning{Explaining the Explainer: Understanding the Inner Workings of Transformer-based Symbolic Regression Models}

\newcommand{\patches}{\textsc{patches}}

\usepackage{cleveref}
\newcommand{\fullcref}[1]{%
  \hyperref[#1]{\Cref*{#1}}%
}
\usepackage{multirow}
\usepackage{fontawesome}

\usepackage{dsfont} 
\usepackage{array} 
\usepackage{forest} 
\usepackage{parskip} 
\usepackage{xcolor} 
\usepackage[table]{xcolor} 
\newcommand{\red}[1]{{\color{red}{#1}}}
\definecolor{darkgreen}{RGB}{0,150,0} 
\newcolumntype{P}[1]{>{\hspace{5mm}}p{#1}}
\definecolor{darkgreen}{RGB}{0,150,0} 
\newcommand{\green}[1]{{\color{darkgreen}#1}}
\usepackage{svg}
\usepackage{hyperref}
\usepackage{xurl}

\begin{document}

\twocolumn[
  \icmltitle{Explaining the Explainer: Understanding the Inner Workings of Transformer-based Symbolic Regression Models}



  \icmlsetsymbol{equal}{*}

  \begin{icmlauthorlist}
    \icmlauthor{Arco van Breda}{yyy}
    \icmlauthor{Erman Acar}{yyy}
  \end{icmlauthorlist}

  \icmlaffiliation{yyy}{University of Amsterdam, Amsterdam, The Netherlands}

  \icmlcorrespondingauthor{Arco van Breda}{a.vanbreda@uva.nl}

  \icmlkeywords{Mechanistic Interpretability, MI, Symbolic Regression, SR, Circuit Discovery, Transformer Circuits, Neurosymbolic AI, Activation Patching}

  \vskip 0.3in
]



\printAffiliationsAndNotice{}  


\begin{abstract}
Following their success across many domains, transformers have proven effective for symbolic regression (SR); however, the internal mechanisms underlying operator generation remain largely unexplored. Although mechanistic interpretability has successfully identified circuits in language and vision models, it has not yet been applied to SR. We introduce \patches, an evolutionary circuit discovery algorithm that identifies compact and correct circuits for SR. Using \patches, we isolate 28 circuits, providing the first circuit-level characterisation of an SR transformer. We validate these findings through a causal evaluation framework based on key notions such as \emph{faithfulness, completeness, and minimality}. Our analysis shows that mean patching with performance-based evaluation most reliably isolates functionally correct circuits. In contrast, we demonstrate that direct logit attribution and probing classifiers primarily capture correlational features rather than causal ones, limiting their utility for circuit discovery. Overall, these results establish SR as a high-potential application domain for mechanistic interpretability and propose a principled methodology for circuit discovery.
\end{abstract}

\section{Introduction}
As deep learning models increasingly automate scientific discovery and reasoning \cite{Wang_Fu_et_al._2023, Cui_Qi_Zhou_Yu_Wang_Zhang_Zhang_Wang_Liu_2025}, interpreting their outputs becomes critical. \textit{Symbolic Regression} (SR) stands out in scientific discovery by producing explicit mathematical equations describing the data. Unlike post-hoc techniques such as LIME or SHAP, which approximate model behaviour locally without capturing relations between variables \cite{Ribeiro2016WhyTrust, Molnar2019InterpretableML, LundbergLee2017SHAP}, SR models aim to uncover the exact input-output relationship over the entire dataset.

\begin{figure*}
    \centering
    \includegraphics[width=1\linewidth, trim=0 0 4.73cm 0, clip]{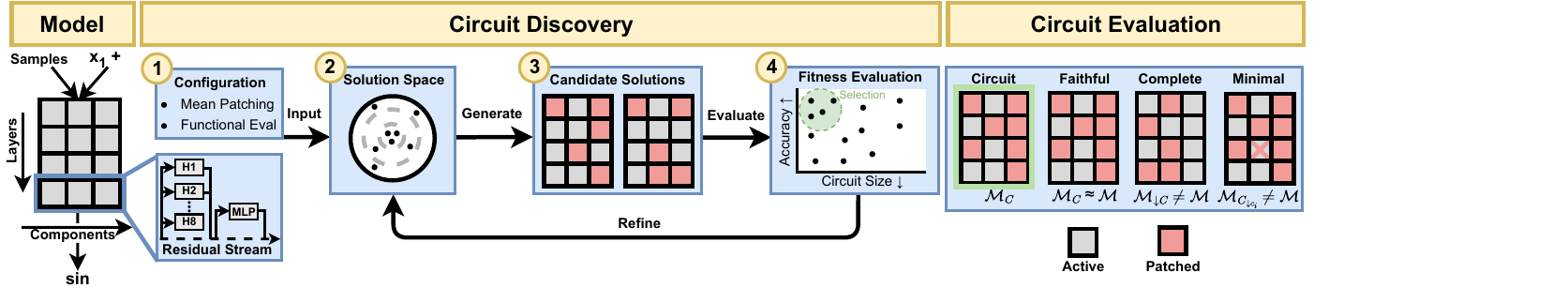}
    \caption{\textbf{The PATCHES Framework.} \textit{Left:} Model schematic: equation samples and previously decoded samples are processed by layers (rows) and components (columns) to predict the target token (e.g., $\sin$). \textit{Center:} Discovery loop. We (1) configure the patching strategy, (2-3) sample candidate masks via CMA-ES, and (4) refine the search distribution based on a fitness trade-off between performance and circuit size. \textit{Right:} Validation criteria based on the best circuit (green). $\mathcal{M}$: Full model; $\mathcal{M}_C$: Only circuit active; $\mathcal{M}_{\downarrow C}$: Only circuit complement active; $\mathcal{M}_{C_{\downarrow c_i}}$: Only circuit active without circuit component $c_i$.}
    \label{fig:FIGURE 1}
\end{figure*}
Despite this transparency, modern NeuroSymbolic SR models that generate these expressions often remain black boxes themselves. Their internal workings and decision-making processes remain poorly understood, which is paradoxical: SR is valued for its interpretability, yet the mechanism that produces explanations for the data itself lacks transparency. This contradiction highlights the need for interpretability methods to uncover how these models arrive at their symbolic outputs.

To resolve the interpretability gap, we turn to \textit{mechanistic interpretability} (MI), a field dedicated to reverse-engineering neural networks into human-understandable forms \cite{introduction_to_circuits}. A key technique within MI is \textit{circuit discovery}, which seeks to identify circuits; interpretable subgraphs of a model that explain a particular behaviour. While MI has successfully uncovered algorithms within language \cite{factual_recall_on_the_neuron_level, ACDC, hanna2023doesgpt2computegreaterthan, FunctionalFaitfullness, iterativepatching, hanna2024faithfaithfulnessgoingcircuit} and vision models \cite{introduction_to_circuits, Automatic_Discovery_of_Visual_Circuits, Going_Deeper_with_Convolutions}, it has not yet been applied to SR. Bridging this gap requires identifying the specific components, e.g., attention heads or multilayer perceptron blocks (MLPs), responsible for generating symbolic operators, such as unary (e.g., \texttt{sin}, \texttt{exp}) and binary (e.g., \(+\), \(\times\)) functions.

Applying MI to SR presents significant methodological challenges. The standard technique for isolating circuits, \textit{activation patching}, lacks consensus, with strategies such as \textit{Mean} and \textit{Resample} patching often yielding conflicting results \cite{Heimersheim_Nanda_2024}. Validation standards also vary between \textit{model-based} metrics (e.g., logits) and \textit{functional-based} metrics (e.g., accuracy). Finally, circuit discovery methods differ in their approach to causality, ranging from \textit{cumulative} searches \cite{ACDC, iterativepatching} to \textit{Direct Logit Attribution} \cite{Liu_Mao_Wen_2025}.

To address these challenges, we introduce a unified framework for circuit discovery in SR, visualised in \fullcref{fig:FIGURE 1}. The first stage of the pipeline isolates specific symbolic behaviours (such as the generation of a \texttt{sin} token) within the Transformer encoder. Then we employ a Probabilistic Algorithm for Tuning Circuits through Heuristic Evolution and Search (\patches), our novel circuit discovery algorithm, to optimise a sparse mask over the model's components, identifying candidate circuits that maximise performance while minimising size. Finally, to ensure circuit correctness, we evaluate these circuits against three formal criteria:
\begin{enumerate}
    \item \textit{Faithfulness}: requiring the circuit to accurately reproduce the model’s target token \cite{hanna2024faithfaithfulnessgoingcircuit}
    \item \textit{Completeness}: all necessary components are included in the circuit \cite{IOI-gpt2},
    \item \textit{Minimalilty}: no redundant components are present in the circuit \cite{ACDC}. 
\end{enumerate}

We summarise our main contributions below: 

\begin{itemize}

    \item We apply circuit discovery to transformer-based SR, isolating circuits responsible for specific unary (e.g., \texttt{sin}, \texttt{exp}) and binary (e.g., \(+\), \(\times\)) operator behaviours. To the best of our knowledge, our work is the first such investigation.  

     \item In doing so, we present a novel circuit discovery method; \patches, which employs the Covariance Matrix Adaptation Evolution Strategy (CMA-ES) that yields smaller circuits than traditional iterative patching for SR.

     \item We establish formal definitions for \textit{faithfulness}, \textit{completeness}, and \textit{minimality} to evaluate circuit correctness. We validate the sufficiency of these definitions through extensive experiments, providing a model-agnostic pipeline designed for easy adaptation across domains.

    \item We systematically compare circuit discovery techniques, evaluating trade-offs between patching strategies (i.e., \textit{Mean} and \textit{Resample} patching). We further distinguish \textit{model-based} from \textit{functional-based} evaluation metrics, and show that \textit{Direct Logit Attribution} and probing are not ideal for identifying causal relations.

\end{itemize}

\section{Related Work}

\paragraph{Mechanistic interpretability and circuit discovery.}
Mechanistic interpretability (MI) aims to reverse engineer neural networks into interpretable algorithms, often formalised as \emph{circuits}, minimal subgraphs sufficient for generating a target token \cite{Miller_Chughtai_Saunders_2024}. Early work analysed individual neurons and attention patterns, later evolving into circuit-level analyses that trace causal pathways through transformers \citep{introduction_to_circuits, elhage2021mathematical}. Circuit discovery has since been applied to tasks such as induction heads, arithmetic, and algorithmic reasoning in language models \citep{Park_Lee_2025, ACDC}. However, MI has not yet been applied to SR models to the best of our knowledge, despite their structured outputs and inherent interpretability. Our work aims to fulfil this gap.

\paragraph{Patching and causal interventions.}
A central tool in MI is activation patching, which tests causal involvement of a component by replacing internal activations during a forward pass and measuring output changes \citep{zhang2024bestpracticesactivationpatching}. To avoid out-of-distribution effects from zeroing or noising, two strategies are common: \emph{mean patching} \citep{IOI-gpt2}, replacing activations with their dataset mean, and \emph{resample patching}, replacing them with those from a corrupted input \citep{Meng_Bau_Andonian_Belinkov_2023}. These strategies differ in assumptions and stability; mean patching is input-independent but dataset-sensitive, while resample patching is instance-specific. Despite widespread use, systematic comparisons between them remain underexplored.

\paragraph{Functional vs model-based evaluation.}
Circuit correctness is typically evaluated using either functional-based or model-based metrics. Functional evaluations measure task-level behaviour such as accuracy or top-$k$ correctness \citep{FunctionalFaitfullness}, while model-based evaluations measure changes in internal quantities such as logit differences or normalised logit scores \citep{zhang2024bestpracticesactivationpatching}.We employ both methods to systematically compare their efficacy.

\paragraph{Symbolic regression models.}
We analyse Neuro Symbolic Regression that Scales (NeSymReS) \citep{Neural_symbolic_regression_that_scales}, a foundational framework that introduced the transformer-based sequence modelling to SR. Its architecture inspires newer models such as SymFormer \cite{SymFormer}, DGSR \cite{DGSR}, and TPSR \cite{Shojaee_Meidani_Farimani_Reddy_2023}. Given its role in establishing this paradigm, NeSymReS is particularly well-suited for mechanistic analysis.

\begin{figure}[h!]
    \centering
    \includegraphics[width=1.00\linewidth]{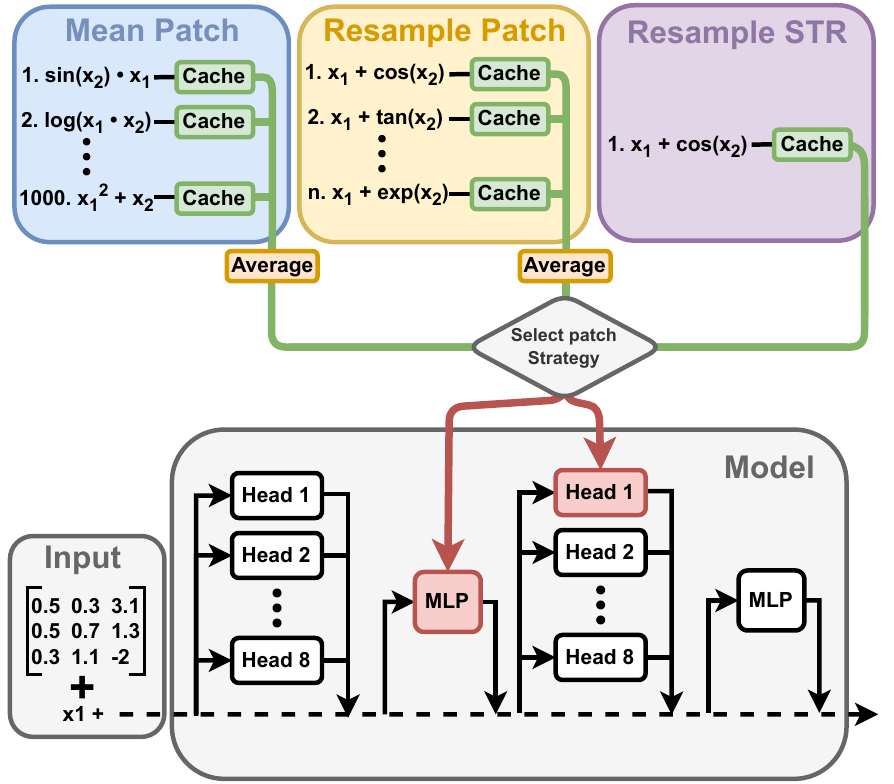}
    \caption{Diagram illustrating three patching strategies applied to the formula $x_1+\sin(x_2)$, target being  $\sin$. The patching strategy is selected manually. Corresponding patches are cached, averaged, and used to modify the selected parts of the model; in this case, Layer 1's feedforward block and Layer 2's Head 1.}
    \label{fig:diagram_patching}
\end{figure}

NeSymReS uses an encoder--decoder architecture. Numerical datasets $(X, y)$ are encoded into a latent representation, which the decoder then uses to autoregressively generate mathematical expressions in prefix notation (e.g., $\sin(x)$ as $[\texttt{sin}, x]$). \hyperref[fig:diagram_patching]{Figure~\ref{fig:diagram_patching} left} provides a visualisation, with further details in \hyperref[APP:NeSymReS]{Appendix~\ref{APP:NeSymReS}}. Numerical constants are represented by a placeholder token $c$, whose values are optimised post-hoc using BFGS \cite{fletcher1987practical}.


\paragraph{Probing classifiers.}
Probing determines if specific information is decodable from a component's activations \cite{hewitt-manning-2019-structural}, but it cannot distinguish between information the model actually uses (causation) and mere artefacts (correlation) \cite{probing_important}. While prior work cites this limitation to justify choosing mechanistic analysis over probing \cite{Probing_correlation_2, Probing_correlation_3}, we take the inverse approach. We apply probes to components already verified as causally essential by our circuit discovery, using these circuits as a ground truth to test if probing accuracy truly reflects functional utility.

\section{Methodology}
\subsection{Patching}
Patching isolates causal influence by replacing the activations of specific components (attention heads or MLPs) with counterfactual values \cite{ACDC}. We perform this intervention during the forward pass of the target token generation, as shown in \hyperref[fig:FIGURE 1]{Figure~\ref*{fig:FIGURE 1} (Model)} for the \texttt{sin} operator. We leverage the NNSight library from \citeauthor{nnsight} to intercept and overwrite the activations during inference. To avoid the out-of-distribution shifts caused by zero- or noise-patching \cite{zhang2024bestpracticesactivationpatching, Miller_Chughtai_Saunders_2024}, we employ three distribution-aware strategies shown in \hyperref[fig:diagram_patching]{Figure~\ref*{fig:diagram_patching}} which we explain next. 

\textbf{Mean patching} (\hyperref[fig:diagram_patching]{Blue panel}) replaces a component's activation with a fixed mean computed over a dataset average \cite{IOI-gpt2}. This patch is input-independent; we select 1000 equations seen by the model during training and compute the average activation for every component.

\textbf{Resample patching} (\hyperref[fig:diagram_patching]{Yellow panel}) calculates a patch using specific counterfactuals generated for every equation in the dataset. For each input, we create corrupted versions by replacing the target token with all alternative unary or binary tokens except for semantically related tokens. We then replace the component's activation with the mean activation computed across these corrupted inputs \cite{factual_recall_on_the_neuron_level}.

\textbf{Resample Symmetric Token Replacement (STR)} (\hyperref[fig:diagram_patching]{Purple panel}) is a single-sample variant of resample patching where the activation is replaced by that of the closest semantically related token \cite{zhang2024bestpracticesactivationpatching, Vig_Gehrmann_Belinkov_Qian_Nevo_Singer_Shieber, Heimersheim_Nanda_2024}.

While mean patching requires tuning a hyper-parameter related to what samples are included in the mean patch, with resample patching, we select appropriate corrupted variations by defining what operators can replace the current operation. Given these trade-offs and the lack of prior comparisons between these methods, we will compare both.

\subsection{Circuit Discovery Pipeline with PATCHES}
To discover the circuits, we propose \patches\ as an alternative to traditional iterative patching methods for circuit discovery. Unlike other approaches, which rely on the sequence in which components are patched, risking larger or incorrect circuits due to local dependencies, \patches\ performs global optimisation over all components simultaneously. Our approach makes use of three key elements: target token-specific datasets for activation patching, the \patches\ algorithm itself, and the well-defined circuit evaluation criteria.

\paragraph{Dataset Selection.}
\label{Dataset Creation and Expression Criteria}
To ensure reliable and interpretable circuit discovery, we generate separate datasets for each target token, such as \texttt{add}, \texttt{log}, or \texttt{sin}, based on three strict criteria. First, we select only expressions that the model correctly reconstructs to ensure failures are due only to patching. Second, to prevent attribution ambiguity, samples must contain the target token but exclude semantically related excluded tokens (e.g., excluding \texttt{cos} when targeting \texttt{sin}). Finally, to ensure fair comparison across strategies, we use the same fixed set of 500 expressions per target token: 100 for discovery and evaluation, and 400 to test generalisation.

\paragraph{PATCHES}
employs the Covariance Matrix Adaptation Evolution Strategy (CMA-ES) algorithm, which optimises stochastic, non-convex spaces by iteratively sampling from a multivariate Gaussian distribution \cite{auger2005restart}. Unlike reinforcement learning, CMA-ES requires no gradients or intermediate rewards, relying solely on fitness evaluations to efficiently discover sparse circuits (see \hyperref[Appendix:CMA-ES]{Appendix~\ref*{Appendix:CMA-ES}} for more details).

The discovery pipeline follows the workflow illustrated in \hyperref[fig:FIGURE 1]{Figure \ref{fig:FIGURE 1} (Circuit Discovery)}. First (\hyperref[fig:FIGURE 1]{1}), we initialise the configuration with the target token, accompanying dataset, patching method, and evaluation strategy. To enable continuous optimisation of discrete structures (\hyperref[fig:FIGURE 1]{2--3}), we model candidate solutions as probabilistic masks. CMA-ES samples these as vectors $\mathbf{x} \in [0,1]^d$, where $x_i$ represents the exclusion probability for component $i$ (head or MLP) and $d$ is the number of components in the model; any component with $x_i > 0.5$ is removed from the circuit. Finally (\hyperref[fig:FIGURE 1]{4}), the search minimises a dual-objective fitness function that balances circuit size $|C|$ against performance penalties:
\begin{equation*}
    F(C) = |C| + \lambda \sum_{i=1}^{m} \max(0, T_i - S_i(C))
\end{equation*}    
\noindent where $\lambda=100$ is a penalty constant, and $\max(0, T_i - S_i(C))$ quantifies the failure to meet the performance threshold $T_i$ for the evaluation metric $S_i$. We initialise the probability of a component being part of the candidate circuit at 0.5 (std 0.1) to encourage exploration near the exclusion threshold. After the evolutionary phase, we apply iterative patching from \cite{iterativepatching} to ensure minimality, which will be formally defined in the following section\footnote{In practice, this step rarely finds smaller circuits. Thus \patches\ is often able to find minimal circuits on its own.}. This approach yields small, faithful circuits that preserve the model's target token. 

\paragraph{Circuit Evaluation Criteria.}
\label{Circuit Evaluation Criteria}
After identifying candidate circuits through \patches, the next step is to assess whether these circuits genuinely explain the target token. This assessment is performed using three key criteria: \textit{faithfulness}, \textit{completeness}, and \textit{minimality} as shown in \hyperref[fig:FIGURE 1]{Figure \ref{fig:FIGURE 1} (Circuit Evaluation)}. We distinguish between functional-level evaluation \cite{FunctionalFaitfullness}, and model-level evaluation metrics \cite{zhang2024bestpracticesactivationpatching}. Functional-level metrics compare the top-$k$ performance of the full model, $\mathcal{M}$, and the circuit, $C$, we use accuracy as the performance measure. Model-level evaluation measures the difference in the target token’s logit score between the full model and the circuit.

We allow for a small performance degradation, $\delta$, in the circuits as a trade-off between circuit correctness and sparsity. $\delta$ is monotonic: if a circuit satisfies a property for some \(\delta_1\), it necessarily satisfies it for any \(\delta_2 \in (\delta_1, 1]\).

\begin{table*}[t]
\centering
\begin{small}
\setlength{\tabcolsep}{4pt} 
\caption{ 
\textbf{Baseline and Circuit Discovery Results} on the generalisation set. 
Performance is measured per target token (Tgt) in Top-$k$ accuracy (T1--T3) and normalised Logit Score (Lgt).
\textit{Baseline:} Scores for the Full Model and the Patched Model using Mean, Resample, or STR patching.
\textit{Circuit:} Discovered Circuit Size (CS), Faithfulness, and Completeness evaluated via Functional Accuracy (Acc) or Model Logits (Lgt).
\faCheck\ indicates the circuit satisfies all 3 correctness criteria. Green: above full model performance, red: below threshold.
}

\begin{tabular}{lll r ccc cccc r c cccc cc c}

\toprule
 & & & &
\multicolumn{7}{c}{\textbf{BASELINE}} & &
\multicolumn{8}{c}{\textbf{CIRCUIT}} \\
\cmidrule(lr){5-11} \cmidrule(lr){13-20}
& & & &
\multicolumn{3}{c}{\textbf{Full Model $\uparrow$}} & 
\multicolumn{4}{c}{\textbf{Patched Model $\downarrow$}} &  & & 
\multicolumn{4}{c}{\textbf{Faithful $\uparrow$}} &
\multicolumn{2}{c}{\textbf{Comp. $\downarrow$}} &  \\
\cmidrule(lr){5-7} \cmidrule(lr){8-11} \cmidrule(lr){14-17} \cmidrule(lr){18-19}
 \textbf{Tgt} & \textbf{Patch} & \textbf{Eval} & &
\textbf{T1} & \textbf{T2} & \textbf{Lgt} & 
\textbf{T1} & \textbf{T2} & \textbf{T3} & \textbf{Lgt} & &
\textbf{CS $\downarrow$}  & \textbf{T1} & \textbf{T2} & \textbf{T3} & \textbf{Lgt} & \textbf{T3} & \textbf{Lgt} & \textbf{\faCheck}\\
\midrule

\multirow{2}{*}{\textbf{Add}} 
 & Res. & Acc & & 0.93 & 0.94 & 0.92 & 0.16 & \red{0.86} & \red{0.99} & 0.21 & & 57 & 0.85 & 0.94 & 0.99 & \red{0.72} & \red{0.99} & 0.23 & \faTimes \\
 & Res. & Lgt & & 0.93 & 0.94 & 0.92 & 0.16 & \red{0.86} & \red{0.99} & 0.21 & & 67 & 0.92 & 0.94 & 1.00 & 0.92 & \red{0.99} & 0.21 & \faTimes \\
\midrule

\multirow{5}{*}{\textbf{Log}} 
 & Mean & Acc & & 0.31 & 0.63 & 0.31 & 0.00 & 0.00 & 0.12 & 0.00 & & 50 & \green{0.66} & \green{1.00} & 1.00 & \green{0.44} & 0.15 & 0.01 & \faCheck \\
 & Mean & Lgt & & 0.31 & 0.63 & 0.31 & 0.00 & 0.00 & 0.12 & 0.00 & & 47 & \red{0.00} & \green{0.96} & 1.00 & 0.23 & 0.12 & 0.00 & \faCheck \\
 & Res. & Acc & & 0.31 & 0.63 & 0.31 & 0.00 & 0.06 & \red{0.54} & 0.00 & & 84 & 0.26 & 0.58 & 0.90 & 0.28 & \red{0.52} & 0.02 & \faTimes \\
 & Res. & Lgt & & 0.31 & 0.63 & 0.31 & 0.00 & 0.06 & \red{0.54} & 0.00 & & 47 & 0.23 & \red{0.41} & \red{0.79} & 0.21 & \red{0.53} & 0.05 & \faTimes \\
 & STR  & Exp & & 0.31 & 0.63 & 0.31 & 0.00 & 0.20 & \red{0.33} & 0.00 & & 70 & 0.24 & 0.60 & 0.92 & 0.30 & \red{0.45} & \red{0.63} & \faTimes \\
\midrule

\multirow{5}{*}{\textbf{Sin}} 
 & Mean & Acc & & 0.73 & 0.83 & 0.70 & 0.00 & 0.00 & 0.00 & 0.00 & & 57 & 0.70 & 0.79 & 1.00 & \red{0.51} & 0.00 & 0.00 & \faCheck \\
 & Mean & Lgt & & 0.73 & 0.83 & 0.70 & 0.00 & 0.00 & 0.00 & 0.00 & & 52 & 0.79 & 0.83 & 1.00 & 0.63 & 0.00 & 0.00 & \faCheck \\
 & Res. & Acc & & 0.73 & 0.83 & 0.70 & 0.03 & 0.12 & 0.24 & 0.04 & & 63 & 0.65 & 0.81 & \red{0.87} & \red{0.59} & \red{0.28} & 0.04 & \faTimes \\
 & Res. & Lgt & & 0.73 & 0.83 & 0.70 & 0.03 & 0.12 & 0.24 & 0.04 & & 62 & 0.65 & 0.80 & \red{0.86} & 0.60 & 0.24 & 0.04 & \faCheck \\
 & STR  & Cos & & 0.73 & 0.83 & 0.70 & 0.00 & \red{0.50} & \red{0.70} & 0.03 & & 69 & \red{0.55} & 0.74 & 0.91 & 0.39 & \red{0.70} & 0.03 & \faTimes \\

\bottomrule
\end{tabular}
\label{tab:baseline+circuit_results}
\end{small}
\end{table*}
\textbf{Faithfulness} assesses whether a circuit, \( C \), reproduces the full model \( \mathcal{M} \)'s behavior for a target token. We evaluate this by patching the circuits complement: \( \mathcal{M}_C\), thereby leaving only the circuit intact.

A circuit \( C \) is \emph{functionally faithful} up to $\delta_f$  for a model \( \mathcal{M} \):
\begin{equation*}
    \left| \mathcal{T}_{k,t}(\mathcal{M}_C |  \mathcal{D} , \hat{y}_{<t})- \mathcal{T}_{k,t}(\mathcal{M} |  \mathcal{D}, \hat{y}_{<t}) \right| \leq \delta_f \quad \forall k,
\end{equation*}

and, a circuit \( C \) is \emph{model faithful} up to $\delta_f$ \phantomsection\label{deltaf} for a model \( \mathcal{M} \):
\begin{equation*}
\frac{1}{|\mathcal{D}|} \sum_{x \in \mathcal{D}} \left( \mathcal{M}(x) - \mathcal{M}_C(x) \right) \leq \delta_f,
\end{equation*}

where \( \mathcal{T}_{k,t}(\mathcal{M} \mid \mathcal{D}, \hat{y}_{<t}) \) denotes the top-\(k\) accuracy of model \( \mathcal{M} \) over dataset \( \mathcal{D} \) at timestep \( t \), given previously predicted tokens \( \hat{y}_{<t} \) and \(\delta_f \in [0, 1]\) is the threshold controlling acceptable degradation in performance of the full model compared to the circuit of the target token.

\textbf{Completeness} verifies that the circuit contains all essential components. We test this by patching the circuit from the full model: \(\mathcal{M}_{\downarrow C}\), and ensuring that the model's ability to predict the target token degrades significantly.

A circuit \( C \) is \emph{functionally complete} up to $\delta_c$ \phantomsection\label{deltac} for a model \( \mathcal{M} \):
\begin{equation*}
\mathcal{T}_{k,t}(\mathcal{M}_{\downarrow C} | \mathcal{D}, \hat{y}_{<t}) \leq \delta_c \quad \forall k
\end{equation*}
A circuit \( C \) is \emph{model complete} up to $\delta_c$ for a model \( \mathcal{M} \):
\begin{equation*}
\frac{1}{|\mathcal{D}|} \sum_{x \in \mathcal{D}} \mathcal{M}_{\downarrow C}(x) \leq \delta_c
\end{equation*}


\textbf{Minimality} ensures that the circuit contains no redundant components \cite{IOI-gpt2, iterativepatching}. We evaluate minimality by first patching the circuit complement, $\mathcal{M}_c$, then we patch each component \( c_i \in C \) individually; a circuit is minimal if the performance of the modified circuit \( \mathcal{M}_{C \downarrow c_i} \) drops by at least \(\delta_f\) compared to the unpatched model:

A circuit \( C \) is \emph{functionally minimal} up to $\delta_f$ for a model \( \mathcal{M} \):
\begin{equation*}
\left| \mathcal{T}_{k,t}(\mathcal{M} |  \mathcal{D}, \hat{y}_{<t}) - \mathcal{T}_{k,t}(\mathcal{M}_{C \downarrow c_i} |  \mathcal{D}, \hat{y}_{<t})\right| \geq \delta_f,
 \quad \forall i \forall k
\end{equation*}
A circuit \( C \) is \emph{model minimal} up to $\delta_f$ for a model \( \mathcal{M} \):
\begin{equation*}
\frac{1}{|\mathcal{D}|} \sum_{x \in \mathcal{D}} \left( \mathcal{M}(x) - \mathcal{M}_{C \downarrow c_i}(x) \right) \geq \delta_f,
\quad \forall i
\end{equation*}

\section{Circuit Discovery Results}

We now turn to experimental results, starting with baseline performance before analysing \patches\ circuits. We focus discovery on the encoder of our autoregressive model to isolate latent representation generation.

With \patches, we explore the discovery of three operators; \texttt{add}, \texttt{log}, \texttt{sin} and illustrate an additional five operators in \hyperref[APP: Additional Circuit Discovery Results]{Appendix \ref{APP: Additional Circuit Discovery Results}}. We compare circuits found with Resample Patching, Mean patching, Functional and Model-based performance metrics\footnote{All code will be made available upon publication.}.

\subsection{Model Performance}
Before analysing circuit discovery, we first assess the unpatched model to establish its baseline performance and limitations. A more detailed evaluation of NeSymReS is provided in \hyperref[APP: Model Performance]{Appendix~\ref{APP: Model Performance}}. To isolate model behaviour, we exclude formulas requiring constants, thereby avoiding reliance on the external BFGS optimiser.

Our model can reconstruct 61.3\% of formulas without constants (see \fullcref{fig:MP with and without constants}), is best in predicting addition and multiplication and struggles with logarithms (see \fullcref{fig:MP element distributions}). We also show that our model has trouble with longer equations (see \hyperref[fig:MP element distributions]{Figure \ref{fig:MP element distributions} a-c}) and verify correct use by testing on the same dataset used by the original authors (see \fullcref{fig:Feyman AI Datset}).

\begin{figure*}[ht!]
    \centering
    \begin{subfigure}{0.49\textwidth}
        \centering
        \includegraphics[width=\linewidth]{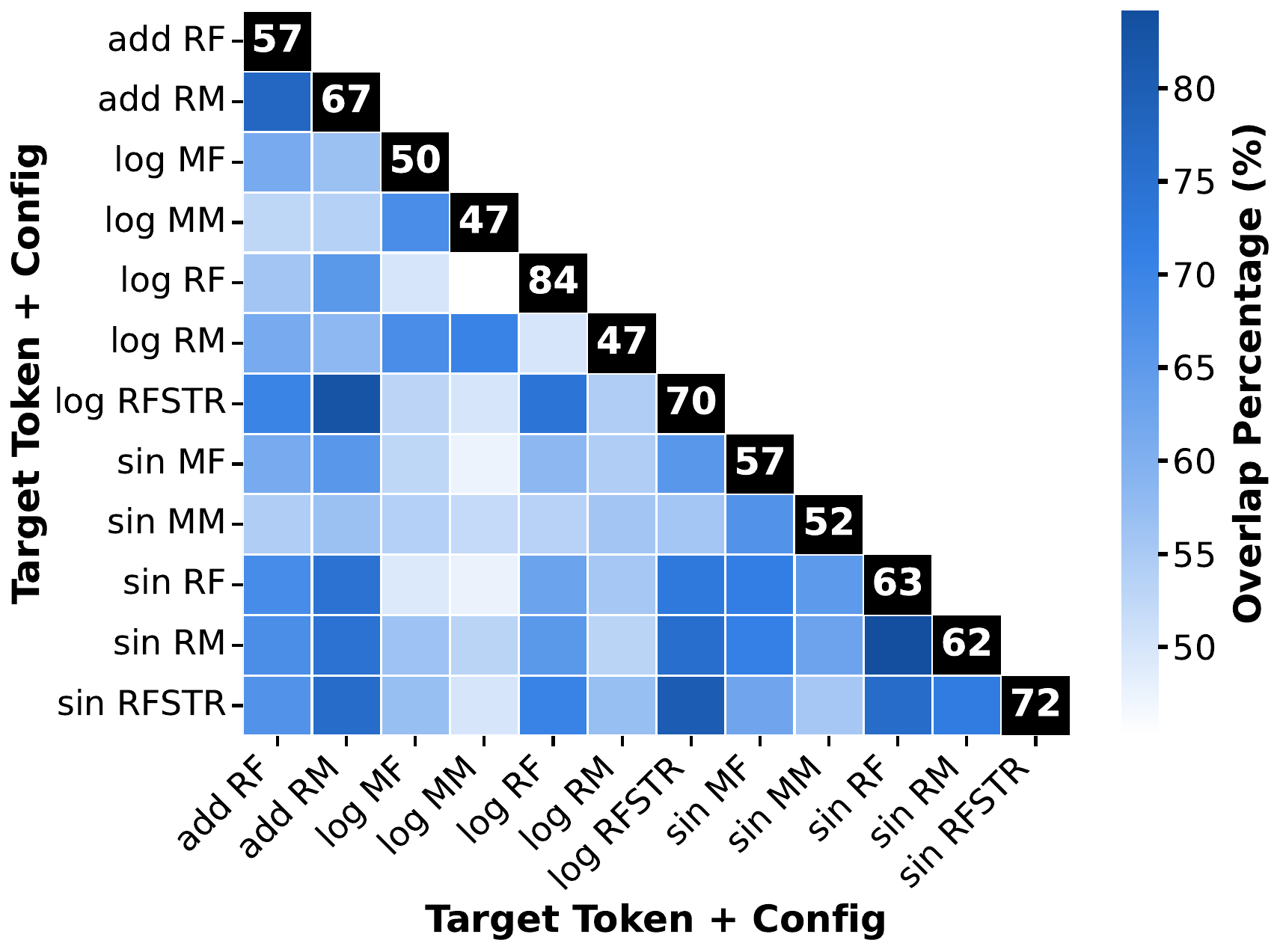}
        \caption{}
        \label{fig:overlap_matrix}
    \end{subfigure}
    \begin{subfigure}{0.49\textwidth}
        \centering
        \includegraphics[width=\linewidth]{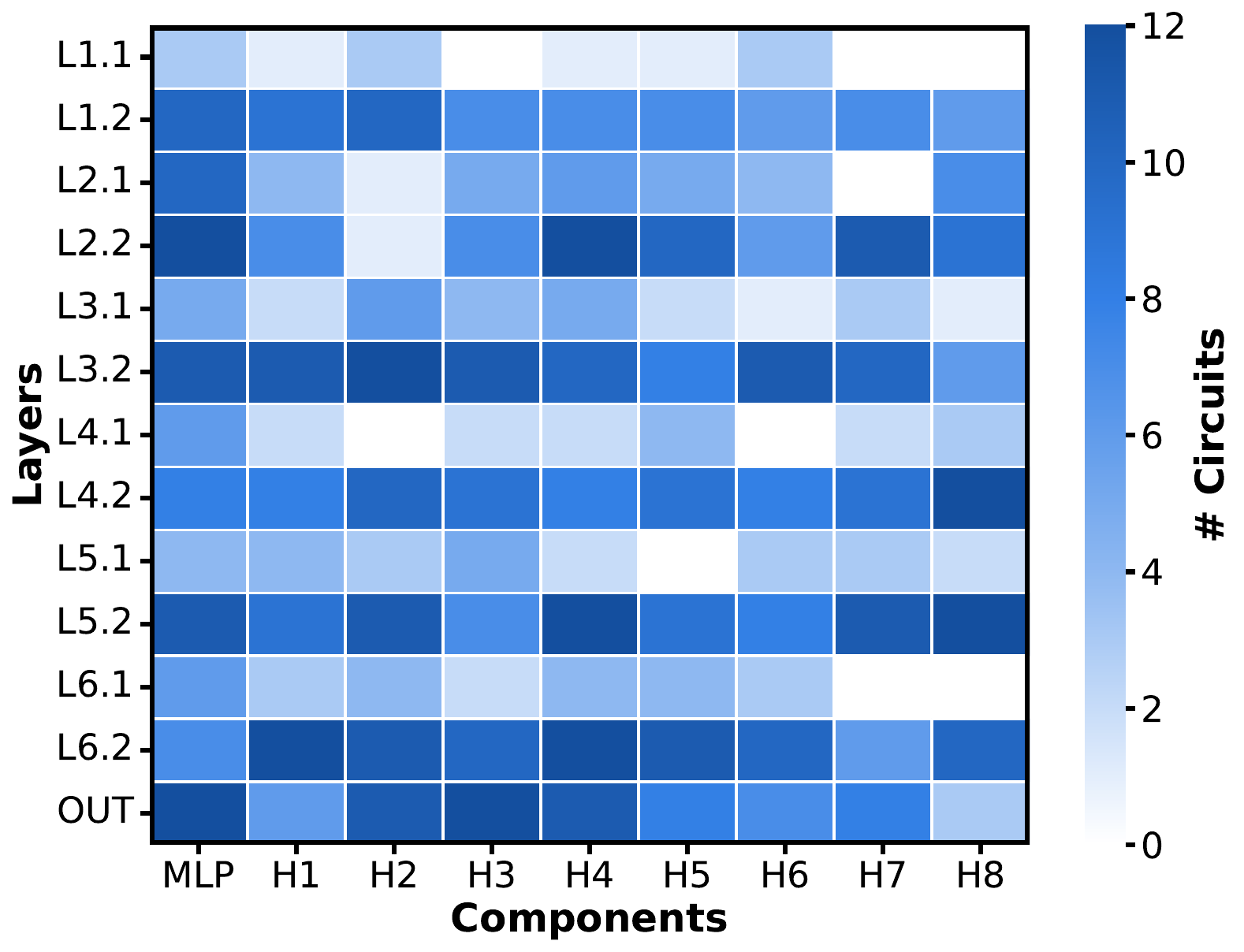}        
        \caption{}
        \label{fig:importance_map}
    \end{subfigure}
    
    \caption{
\textbf{Confusion plots illustrating the diversity of circuits identified in the model.} \textit{(a)} Overlap percentages between operators; the diagonal (black) indicates circuit length. RF: Resample Functional; RM: Resample Model; MF Mean Functional; MM: Mean Model; RFSTR: Resample Functional STR \textit{(b)} Usage frequency of components in the selected circuits. L\emph{x.x} denotes layer \emph{x.x}; OUT the output projection; MLP the multilayer perceptron block; H\textit{x} attention head \emph{x}. Full tables and figures in \hyperref[Additional Verification Experiments]{Appendix \ref{Additional Verification Experiments}}.}
    \label{fig:additional_verification_experiments}
\end{figure*}

\subsection{Baselines}
Establishing a baseline is vital for circuit discovery: we must ensure the full model reliably performs the task, and conversely, that patching disrupts the target token for causal isolation. \hyperref[tab:baseline+circuit_results]{Table \ref{tab:baseline+circuit_results} (BASELINE)} shows baselines across operations. We evaluate top-$k$ accuracy with $k=3$, reflecting the model's beam search behavior (See \fullcref{fig:MP distribution}).

Ideally, patching all components should substantially degrade model performance. However, resample patching fails to suppress the \texttt{sin} operator, retaining high performance. This indicates that the intervention does not fully erase target token information, thereby obscuring causal effects and hindering circuit discovery. 

This issue is even more pronounced with STR patching; as shown in \hyperref[tab:baselines]{Appendix \ref{APP: Additional Circuit Discovery Results} Table \ref{tab:baselines}}, all tested operators retain excessively high baseline scores. Consequently, we caution against using resample STR without explicitly verifying that they sufficiently disrupt the model's behaviour.

Similarly, addition and multiplication retain high baselines due to dataset overrepresentation, causing the model to default to them even when patched. To mitigate this, we restrict analysis to resample patching for these tokens, which more effectively removes correlated information.

\subsection{Circuits}
\label{circuit_results}
\paragraph{The Sine Mean Functional Circuit.}
We illustrate the discovery process with the \texttt{sin} operator using Mean Patching and Functional evaluation. First, we generate 500 equations containing \texttt{sin}, excluding \texttt{cos} to ensure specificity. Second, the baseline: the full model achieves 100\% top-3 accuracy, while the fully mean-patched model drops to 0.00\%, see \hyperref[tab:baseline+circuit_results]{Table \ref{tab:baseline+circuit_results} (BASELINE)}, confirming the patch suppresses the target token. Third, \patches\ finds a minimal subgraph of 57 components. Finally, evaluation confirms correctness by recovering full accuracy (Faithful: 1.00), and its complement fails to generate the target token (Complete: 0.00), see \hyperref[tab:baseline+circuit_results]{Table \ref{tab:baseline+circuit_results} (CIRCUIT)}.

\paragraph{Overall results} shown in \fullcref{tab:baseline+circuit_results} present the 12 circuits discovered across different configurations on the generalisation set with all 28 circuits observed in \fullcref{tab:circuits test}. Discovery set results are shown in \fullcref{tab:circuits Train}. Discovery set scores are similar to the generalisation results, confirming that the discovered circuits generalise well to unseen equations. 

To evaluate the circuits, we require faithfulness to be within 10 percentage points of the baseline (\hyperref[deltaf]{$\delta_f$}$=0.1$), and patched circuit performance to fall below 25\% for completeness (\hyperref[deltac]{$\delta_c$}$=0.25)$. These thresholds were chosen empirically to balance model flexibility and behavioural specificity.

Out of the 12 evaluated circuits, one does not meet the faithfulness threshold under its respective evaluation strategy. For completeness, seven circuits exceed the 25\% cutoff. However, this is often due to high baseline scores, making it harder to suppress the behaviour through patching. Excluding cases where patched model baseline performance exceeds 25\%, only a single circuit, \texttt{sin}-RF, remains incomplete. We report both successful and unsuccessful cases to emphasise the importance of strong baselines and careful consideration of patching strategies.

In total, 13 of the 28 circuits satisfy all three criteria: faithfulness, completeness, and minimality, and are therefore considered correct\footnote{We omit minimality from \fullcref{tab:baseline+circuit_results} as it is enforced by \patches, which inherently solves for the minimal subset of components required by the chosen evaluation strategy.}. We illustrate that our approach can be expand upon easily to multi-token circuits based on the work from \citeauthor{Garc_a_Carrasco_2024} in \hyperref[Multi-Token Circuits]{Appendix \ref{Multi-Token Circuits}}.

Circuits comprise 40\%--78\% of the model. We attribute this density to SR's constraints: unlike NLP, where vast vocabularies enable sparsity, our $\sim$20-token vocabulary necessitates a distributed, polysemantic encoding of complex syntax (e.g., nesting). This high utilization is not unique to SR; similar circuit densities are observed in sequence continuation tasks \cite{iterativepatching}. Consequently, usage variations appear driven by patching and evaluation strategies rather than intrinsic complexity; we analyse these effects below.

\paragraph{Functional Vs. Model-based performance.} Functional-based evaluation yields shorter circuits for most operators (\fullcref{tab:circuits test}), though neither method is strictly superior. While logit-metrics capture subtle internal effects, they may not reflect meaningful output changes, which are critical in autoregressive settings where small shifts flip predictions. Since we are interested in explaining the overall behaviour to users rather than fine-tuning internal states, we advocate for functional evaluation.

\begin{figure*}[t]
    \centering
    \begin{subfigure}{0.49\textwidth}
        \centering
        \includegraphics[width=\linewidth]{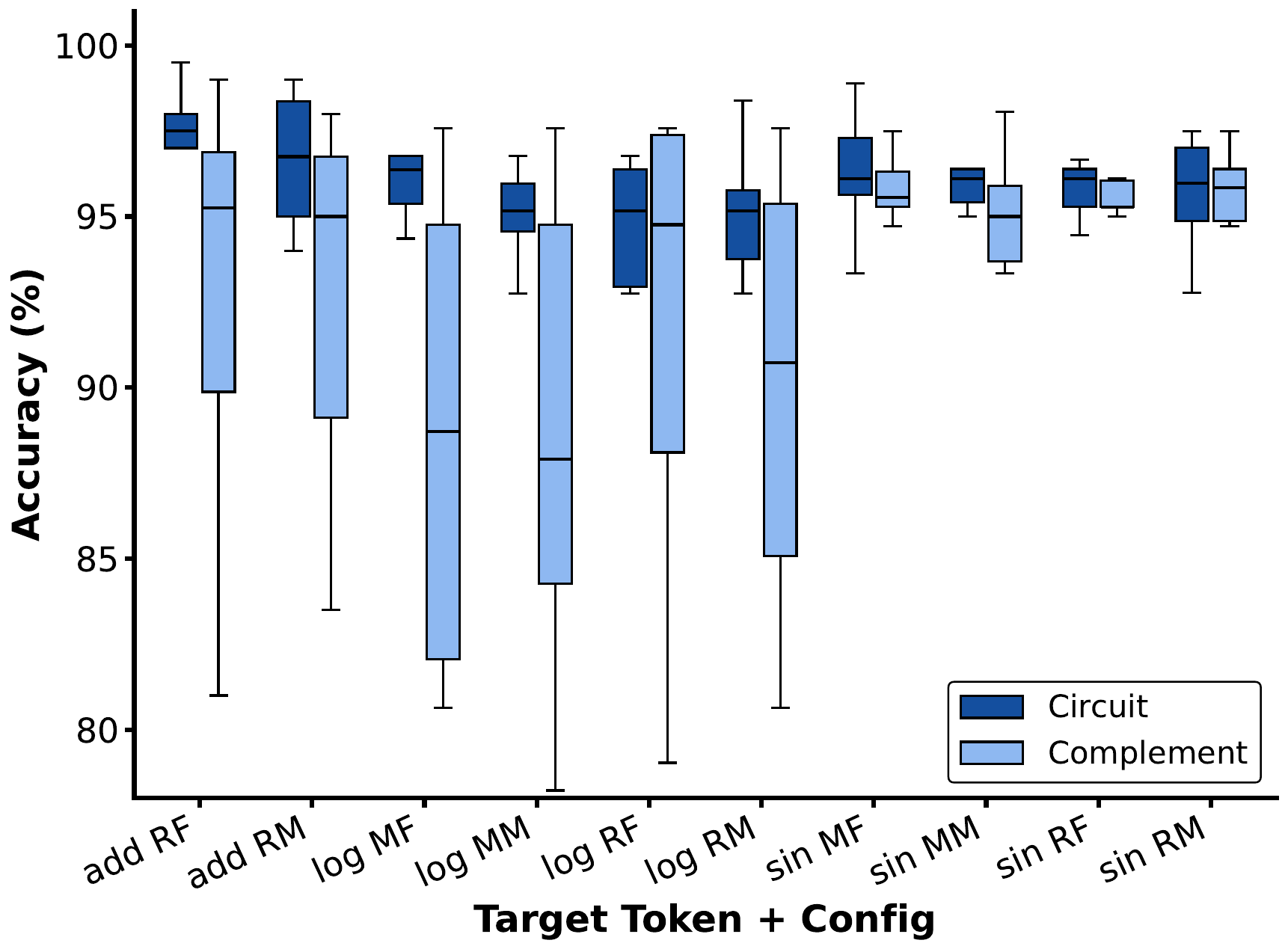}
        \caption{}
        \label{fig:probe_main}
    \end{subfigure}
    \begin{subfigure}{0.49\textwidth}
        \centering
        \includegraphics[width=\linewidth]{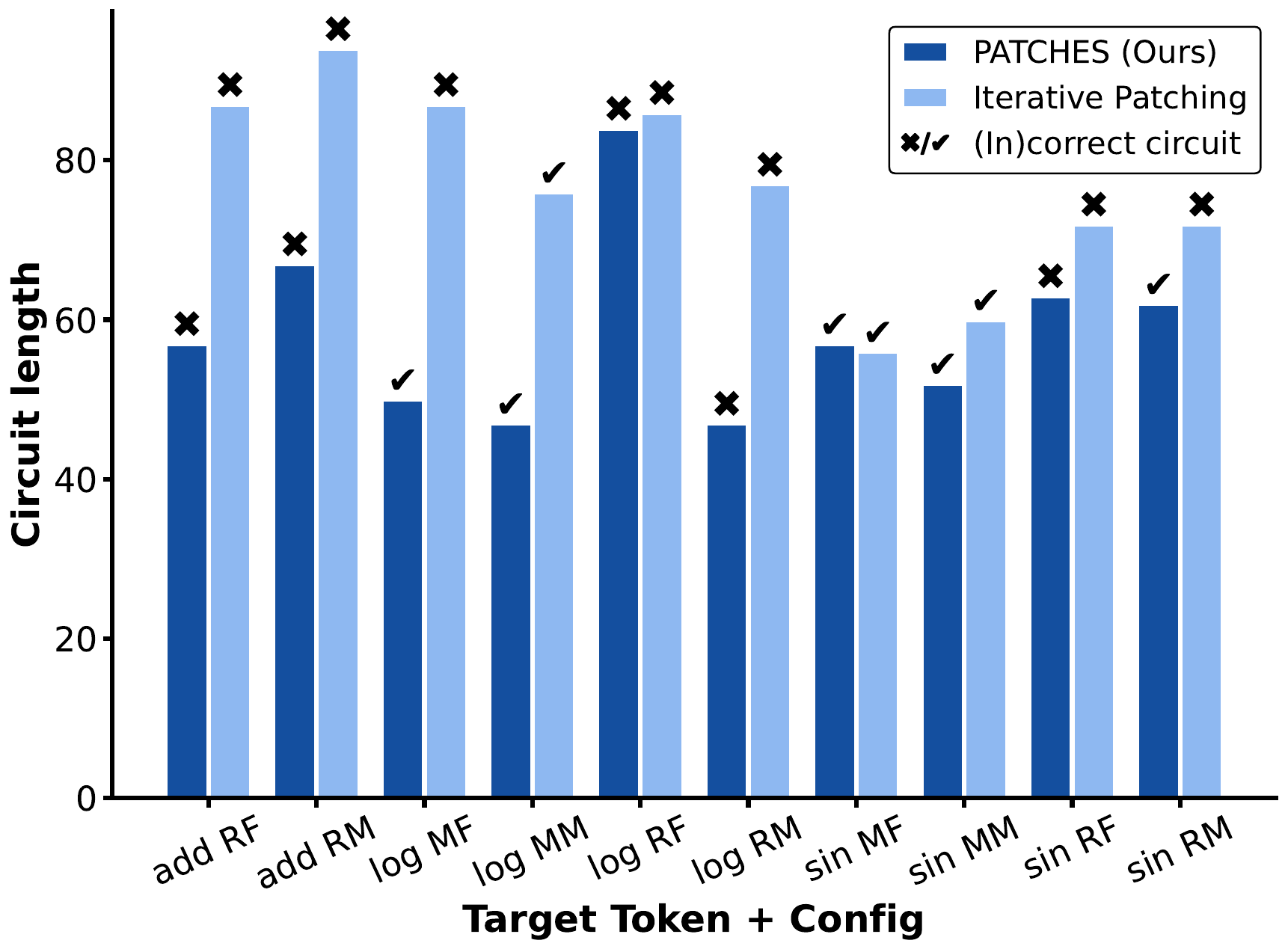}
        \caption{}
        \label{fig:circuit_lengths}
    \end{subfigure}
    \caption{\textbf{Probing and Iterative Patching Results.} RF: Resample Functional, RM: Resample Model, MF Mean Functional, MM: Mean Model, RFSTR: Resample Functional STR. \textit{(a)} Probe accuracy comparisons between circuit and complement components. Full Table in \hyperref[tab:probing]{Appendix \ref{APP:Probing}}. \textit{(b)} Circuit length comparison between \patches\ and Iterative Patching. \faTimes/\faCheck \; indicate incorrect and correct circuits respectively. Full Table in \hyperref[Iterative Patching: Additional Results]{Appendix \ref{Iterative Patching: Additional Results}}.}
    \label{fig:placeholder}
\end{figure*}

\paragraph{Mean Vs. Resample Patching} results illustrate that for \texttt{sin} and \texttt{log} (and \texttt{cos}, \texttt{tan}), resample patching consistently produces longer circuits. This is expected, as it excludes the target operation and requires more components to preserve performance. While faithfulness remain similar, completeness often worsens. The same is true for resample STR patching, which also produces longer circuits and fails the completeness threshold in three of four cases. We therefore recommend mean patching, which offers correct circuits more often, and is computationally more efficient while generating smaller, more interpretable circuits.

\subsection{Verification Experiments} 
We verify circuits via three experiments. First, we check component usage and overlap to ensure distinctness. Next, we illustrate that probing circuit components contain more linearly decodable information.

The first two verification steps are displayed in \fullcref{fig:additional_verification_experiments}. To confirm that circuits are not merely subsets of one another, we report overlap percentages in \fullcref{fig:overlap_matrix}. As expected, higher correlations appear for longer circuits and when the same operators are compared. \fullcref{fig:importance_map} illustrates which components are utilised across layers, revealing a well-distributed usage pattern in which the full model contributes to these operators. Full results for all circuits, along with an additional verification experiment, are provided in \hyperref[Additional Verification Experiments]{Appendix~\ref{Additional Verification Experiments}}.

\begin{figure*}[t!]
    \centering
    \begin{subfigure}{0.49\textwidth}
        \centering
        \includegraphics[width=\linewidth]{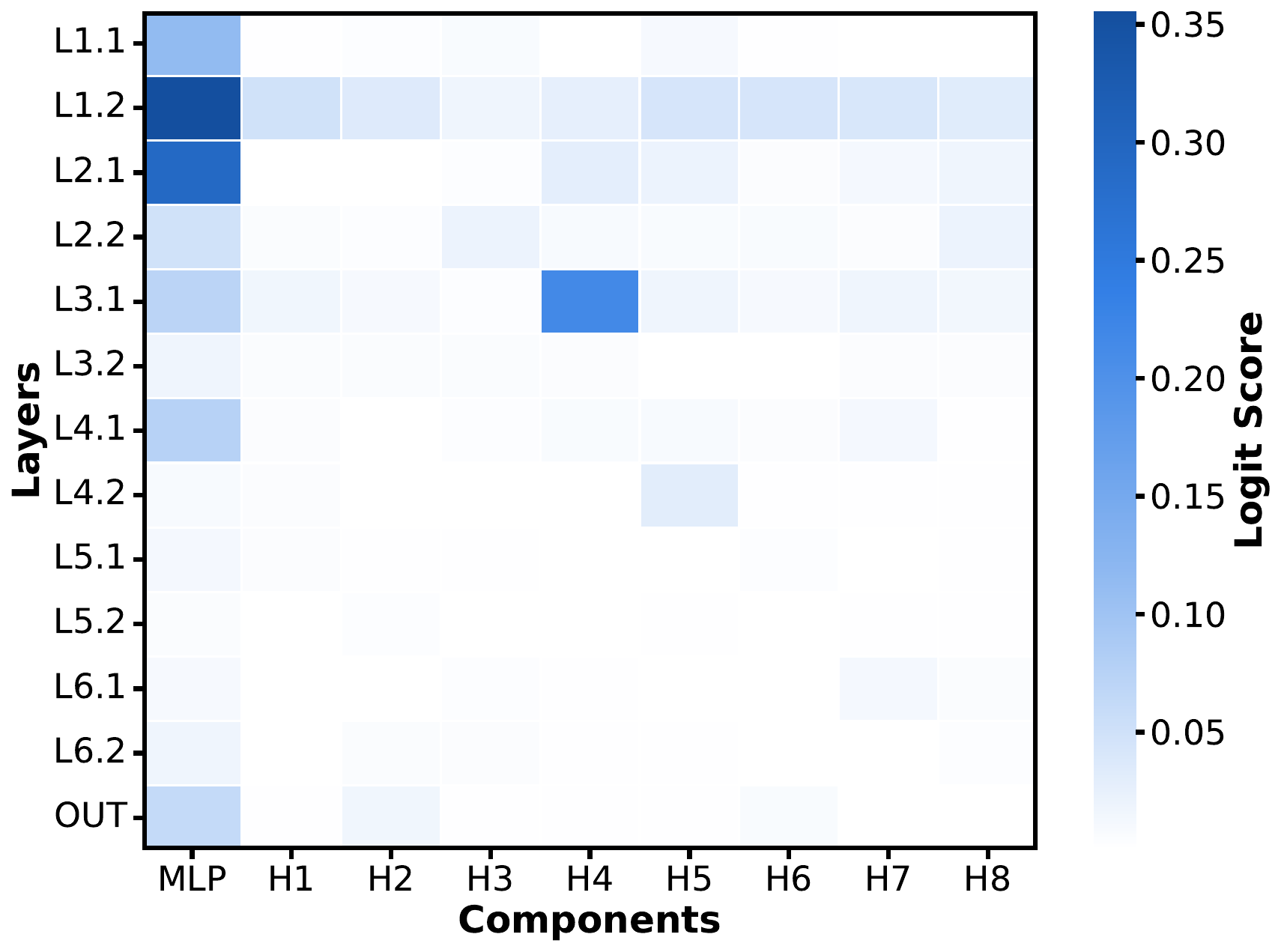}
        \caption{}
        \label{fig:DLA Sin Mean attention}
    \end{subfigure}
    \begin{subfigure}{0.49\textwidth}
        \centering
        \includegraphics[width=\linewidth]{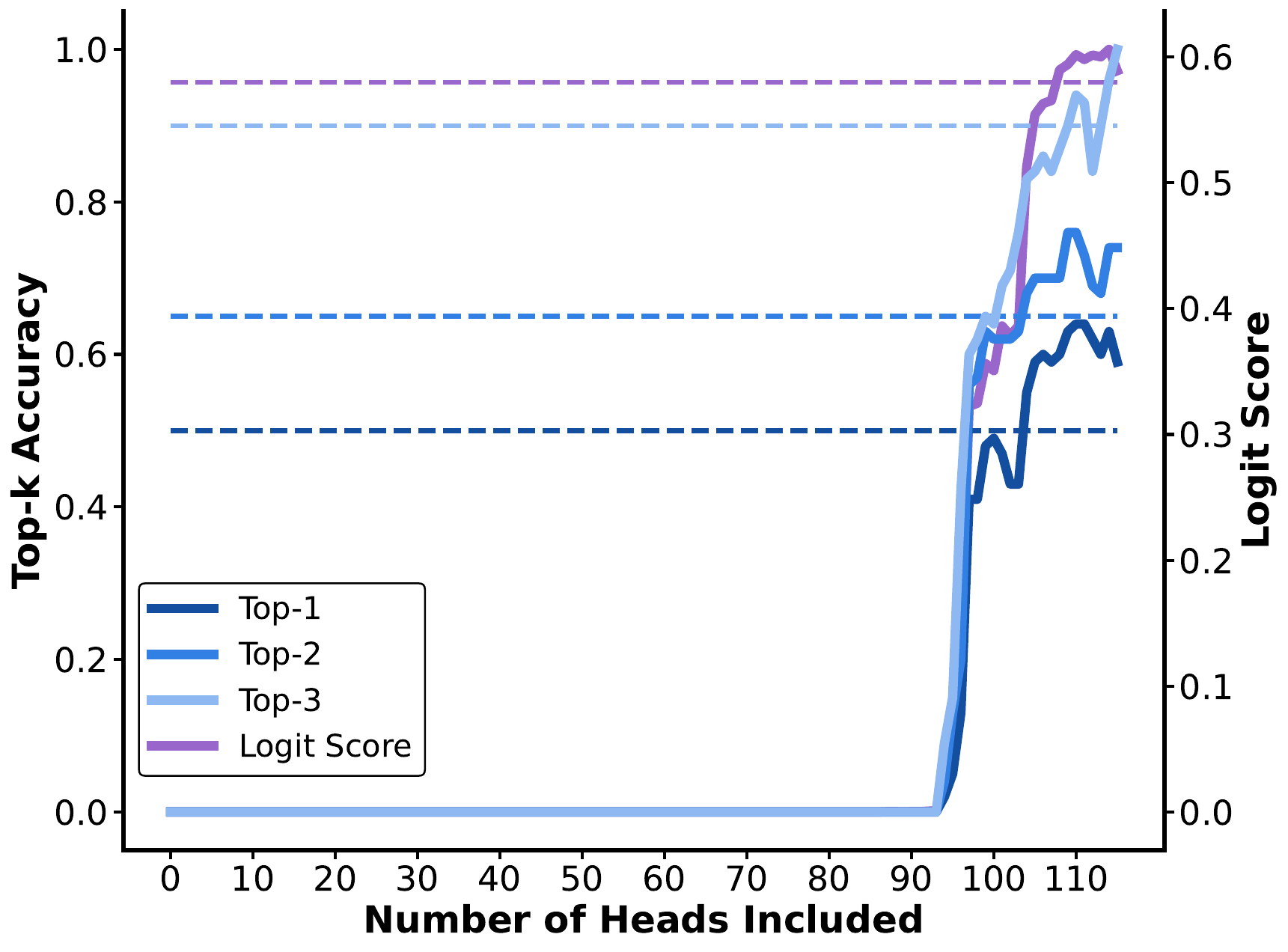}        
        \caption{}
         \label{fig:DLA Sin Mean top}
    \end{subfigure}
    \caption{\textbf{Direct attribution and faithfulness evaluation}.
\textit{(a)} Change in logit score vs.\ \texttt{sin} mean when patching individual heads, averaged over 100 samples. L\emph{x.x} denotes layer \emph{x.x}; OUT the output projection; MLP the multilayer perceptron block; H\textit{x} attention head \emph{x}.
\textit{(b)} Faithfulness evaluation of \texttt{sin} mean importance ranking; thresholds (dashed lines) from \fullcref{tab:baseline+circuit_results}. }
\label{fig:DLAPaper}
\end{figure*}

\vspace{-5pt}
\paragraph{Probing.}\label{section:Experiments_Probing}
We use probing as an additional verification experiment to assess whether components identified by circuit discovery also exhibit stronger linear decodability, and to situate our findings within the broader probing literature. We compare probing performance between components inside the circuit and components outside the circuit.

To do so, we randomly sample 10 components from the circuit and 10 components from the complement of the circuit. For each of these 20 components, we train a probing model using 10 different random seeds, resulting in a total of 200 trained models per evaluation and patching configuration (see \hyperref[APP:Probing]{Appendix \ref{APP:Probing}} for hyperparameter settings). Subsequently, we calculate the average performance across seeds for both circuit and circuit complement components. A paired t-test is conducted to assess the significance of the performance difference between groups.

Across all target tokens, probing accuracy is generally higher for circuit components than for components in the circuit complement, as shown in \fullcref{fig:probe_main}. This suggests that circuit components tend to encode more linearly decodable information about the model’s output behaviour. In addition, circuit components exhibit similar or lower standard deviation in probe accuracy, indicating more consistent information encoding across sampled components.

Despite these trends, differences in probing accuracy are rarely statistically significant. Only the \texttt{log} operator shows a significant gap between circuit and out-of-circuit components, likely due to its smaller and more isolated circuit.

In addition to verifying that more information is stored in circuit components, these results reinforce prior findings that high probing performance does not necessarily imply causal relevance \cite{Probing_correlation_1, Probing_correlation_2, Probing_correlation_3}. While circuit components more reliably represent behaviourally relevant information, probing alone fails to confirm that the model uses this information during prediction, as probing accuracy remains high even for complement components.

\subsection{Alternative Circuit discovery techniques.}
To illustrate our circuits are compact we compare our method to Iterative Patching \cite{iterativepatching, ACDC}, which initialises a candidate circuit as the full model and sequentially patches components layer-by-layer. If performance does not decrease significantly, the component is deemed non-essential and removed from the candidate circuit. The algorithm alternates between backward and forward sweeps until convergence. However, empirical testing reveals that this greedy, order-dependent approach often fails to find small circuits, as early pruning decisions can lock the search into local optima.

We compare \patches\ to Iterative Patching by applying Iterative Patching across the same operations as in \hyperref[circuit_results]{Section 4.2}; additional results are provided in \hyperref[Iterative Patching: Additional Results]{Appendix \ref{Iterative Patching: Additional Results}}.

\fullcref{fig:circuit_lengths} illustrates that circuits found by Iterative Patching are overall larger. In addition to being larger circuits, these circuits do not generalise as well as the circuits found by \patches\ as the \texttt{log} Mean Functional circuit and \texttt{sin} Resample Model circuits are found to be incorrect on the generalisation set, which is not the case for \patches. These examples illustrate that \patches\ produces smaller circuits and that  more components does not guarantee better performance and may introduce confounding heads.

\section{Direct Logit Attribution}
Unlike cumulative circuit discovery methods such as \patches\ and Iterative Patching, Direct Logit Attribution evaluates components individually by patching a component out and measuring their logit change. Components with the highest impact are selected to form circuits, but these methods often omit standard circuit correctness evaluations. We evaluate circuits by first testing for faithfulness, followed by minimality and completeness.

To find a circuit via Direct Logit Attribution, we iteratively reintroduce components to a fully patched model, starting with the component showing the largest individual logit or performance difference when patched. We continue this process until either the logit score or the Top-$k$ accuracies meet the respective thresholds. 

Results displayed in \fullcref{fig:DLAPaper} suggest that Direct Logit Attribution struggles to produce a faithful circuit, requiring almost all components to meet model and functional faithfulness, creating non-minimal circuits. We illustrate that this ordering and reintroducing of components has little correlation with model usage as we see no score improvement for the first 50 components. Additional results in \hyperref[fig:DLAAPP]{Appendix \ref{Direct Logit Attribution Additional Results} Figure \ref{fig:DLAAPP}} show that resample patching, despite earlier gains, still requires many components to achieve faithfulness.

These results illustrate that this method does not produce circuits compatible with our evaluation strategies, as it fails the minimality criterion required to isolate specific target tokens. While the circuits are technically complete, since nearly the full model is needed to reach high performance, they meet neither faithfulness nor minimality standards.

\section{Conclusion}
In this work, we presented the first mechanistic interpretability analysis of transformer-based symbolic regression. By introducing \patches, an evolutionary circuit discovery algorithm, we identified 28 circuits from which 13 were found to generalise well to unseen equations. These circuits were found to be faithful, complete and minimal for unary and binary operators. Our systematic comparison reveals that standard greedy search methods often fail to find small circuits for symbolic regression, and that mean-patching combined with functional evaluation offers a more reliable framework for circuit discovery. Crucially, our findings challenge the utility of Direct Logit Attribution and probing classifiers for causal analysis, as neither reliably correlates with the components actually used by the model.




\section*{Impact Statement}
``This paper presents work whose goal is to advance the field of mechanistic interpretability. There are many potential societal consequences of our work, none of which we feel must be specifically highlighted here.''


\bibliography{mybib}

@article{Neural_symbolic_regression_that_scales, title={Neural Symbolic Regression that Scales}, url={http://arxiv.org/abs/2106.06427}, DOI={10.48550/arXiv.2106.06427}, abstractNote={Symbolic equations are at the core of scientiﬁc discovery. The task of discovering the underlying equation from a set of input-output pairs is called symbolic regression. Traditionally, symbolic regression methods use hand-designed strategies that do not improve with experience. In this paper, we introduce the ﬁrst symbolic regression method that leverages large scale pre-training. We procedurally generate an unbounded set of equations, and simultaneously pre-train a Transformer to predict the symbolic equation from a corresponding set of input-output-pairs. At test time, we query the model on a new set of points and use its output to guide the search for the equation. We show empirically that this approach can rediscover a set of well-known physical equations, and that it improves over time with more data and computational resources.}, note={arXiv:2106.06427 [cs]}, number={arXiv:2106.06427}, publisher={arXiv}, author={Biggio, Luca and Bendinelli, Tommaso and Neitz, Alexander and Lucchi, Aurelien and Parascandolo, Giambattista}, year={2021}, month=jun, language={en} }

@article{set_encoder, title={Set Transformer: A Framework for Attention-based Permutation-Invariant Neural Networks}, abstractNote={Many machine learning tasks such as multiple instance learning, 3D shape recognition and fewshot image classiﬁcation are deﬁned on sets of instances. Since solutions to such problems do not depend on the order of elements of the set, models used to address them should be permutation invariant. We present an attention-based neural network module, the Set Transformer, speciﬁcally designed to model interactions among elements in the input set. The model consists of an encoder and a decoder, both of which rely on attention mechanisms. In an effort to reduce computational complexity, we introduce an attention scheme inspired by inducing point methods from sparse Gaussian process literature. It reduces computation time of self-attention from quadratic to linear in the number of elements in the set. We show that our model is theoretically attractive and we evaluate it on a range of tasks, demonstrating increased performance compared to recent methods for set-structured data.}, author={Lee, Juho and Lee, Yoonho and Kim, Jungtaek and Kosiorek, Adam R and Choi, Seungjin and Teh, Yee Whye}, language={en} }

@article{func_gen, title={Deep Learning for Symbolic Mathematics}, url={http://arxiv.org/abs/1912.01412}, DOI={10.48550/arXiv.1912.01412}, abstractNote={Neural networks have a reputation for being better at solving statistical or approximate problems than at performing calculations or working with symbolic data. In this paper, we show that they can be surprisingly good at more elaborated tasks in mathematics, such as symbolic integration and solving differential equations. We propose a syntax for representing mathematical problems, and methods for generating large datasets that can be used to train sequence-to-sequence models. We achieve results that outperform commercial Computer Algebra Systems such as Matlab or Mathematica.}, note={arXiv:1912.01412 [cs]}, number={arXiv:1912.01412}, publisher={arXiv}, author={Lample, Guillaume and Charton, François}, year={2019}, month=dec, language={en} }

@article{Miller_Chughtai_Saunders_2024, title={Transformer Circuit Faithfulness Metrics are not Robust}, url={http://arxiv.org/abs/2407.08734}, DOI={10.48550/arXiv.2407.08734}, abstractNote={Mechanistic interpretability work attempts to reverse engineer the learned algorithms present inside neural networks. One focus of this work has been to discover ‘circuits’ – subgraphs of the full model that explain behaviour on specific tasks. But how do we measure the performance of such circuits? Prior work has attempted to measure circuit ‘faithfulness’ – the degree to which the circuit replicates the performance of the full model. In this work, we survey many considerations for designing experiments that measure circuit faithfulness by ablating portions of the model’s computation. Concerningly, we find existing methods are highly sensitive to seemingly insignificant changes in the ablation methodology. We conclude that existing circuit faithfulness scores reflect both the methodological choices of researchers as well as the actual components of the circuit - the task a circuit is required to perform depends on the ablation used to test it. The ultimate goal of mechanistic interpretability work is to understand neural networks, so we emphasize the need for more clarity in the precise claims being made about circuits. We open source a library at this https URL that includes highly efficient implementations of a wide range of ablation methodologies and circuit discovery algorithms.}, note={arXiv:2407.08734 [cs]}, number={arXiv:2407.08734}, publisher={arXiv}, author={Miller, Joseph and Chughtai, Bilal and Saunders, William}, year={2024}, month=jul, language={en} }

@article{IOI-gpt2, title={INTERPRETABILITY IN THE WILD: A CIRCUIT FOR INDIRECT OBJECT IDENTIFICATION IN GPT-2 SMALL}, author={Wang, Kevin and Variengien, Alexandre and Conmy, Arthur and Shlegeris, Buck and Steinhardt, Jacob}, year={2023}, language={en} }

@misc{factual_recall_on_the_neuron_level,
  author = {Neel Nanda and Senthooran Rajamanoharan and Janos Kramar and Rohin Shah},
  title = {Fact finding: Attempting to reverse-engineer factual recall on the neuron level},
  year = {2023},
  month = {Dec},
  url = {https://www.alignmentforum.org/posts/iGuwZTHWb6DFY3sKB/fact-finding-attempting-to-reverse-engineer-factual-recall}
}

@article{introduction_to_circuits,
  author = {Chris Olah and Nick Cammarata and Ludwig Schubert and Gabriel Goh and Michael Petrov and Shan Carter},
  title = {Zoom in: An introduction to circuits},
  journal = {Distill},
  year = {2020},
  doi = {10.23915/distill.00024.001},
  url = {https://distill.pub/2020/circuits/zoom-in}
}

@article{ACDC, title={Towards Automated Circuit Discovery for Mechanistic Interpretability}, url={http://arxiv.org/abs/2304.14997}, DOI={10.48550/arXiv.2304.14997}, abstractNote={Through considerable effort and intuition, several recent works have reverseengineered nontrivial behaviors of transformer models. This paper systematizes the mechanistic interpretability process they followed. First, researchers choose a metric and dataset that elicit the desired model behavior. Then, they apply activation patching to find which abstract neural network units are involved in the behavior. By varying the dataset, metric, and units under investigation, researchers can understand the functionality of each component.}, note={arXiv:2304.14997 [cs]}, number={arXiv:2304.14997}, publisher={arXiv}, author={Conmy, Arthur and Mavor-Parker, Augustine N. and Lynch, Aengus and Heimersheim, Stefan and Garriga-Alonso, Adrià}, year={2023}, month=oct, language={en} }

@article{Automatic_Discovery_of_Visual_Circuits, title={Automatic Discovery of Visual Circuits}, url={http://arxiv.org/abs/2404.14349}, DOI={10.48550/arXiv.2404.14349}, abstractNote={To date, most discoveries of network subcomponents that implement humaninterpretable computations in deep vision models have involved close study of single units and large amounts of human labor. We explore scalable methods for extracting the subgraph of a vision model’s computational graph that underlies recognition of a specific visual concept. We introduce a new method for identifying these subgraphs: specifying a visual concept using a few examples, and then tracing the interdependence of neuron activations across layers, or their functional connectivity. We find that our approach extracts circuits that causally affect model output, and that editing these circuits can defend large pretrained models from adversarial attacks. Our code and data are available at https://github.com/ multimodal-interpretability/visual-circuits.}, note={arXiv:2404.14349 [cs]}, number={arXiv:2404.14349}, publisher={arXiv}, author={Rajaram, Achyuta and Chowdhury, Neil and Torralba, Antonio and Andreas, Jacob and Schwettmann, Sarah}, year={2024}, month=apr, language={en} }

@article{Going_Deeper_with_Convolutions, title={Going Deeper with Convolutions}, url={http://arxiv.org/abs/1409.4842}, DOI={10.48550/arXiv.1409.4842}, abstractNote={We propose a deep convolutional neural network architecture codenamed Inception, which was responsible for setting the new state of the art for classiﬁcation and detection in the ImageNet Large-Scale Visual Recognition Challenge 2014 (ILSVRC14). The main hallmark of this architecture is the improved utilization of the computing resources inside the network. This was achieved by a carefully crafted design that allows for increasing the depth and width of the network while keeping the computational budget constant. To optimize quality, the architectural decisions were based on the Hebbian principle and the intuition of multi-scale processing. One particular incarnation used in our submission for ILSVRC14 is called GoogLeNet, a 22 layers deep network, the quality of which is assessed in the context of classiﬁcation and detection.}, note={arXiv:1409.4842 [cs]}, number={arXiv:1409.4842}, publisher={arXiv}, author={Szegedy, Christian and Liu, Wei and Jia, Yangqing and Sermanet, Pierre and Reed, Scott and Anguelov, Dragomir and Erhan, Dumitru and Vanhoucke, Vincent and Rabinovich, Andrew}, year={2014}, month=sep, language={en} }

@misc{zhang2024bestpracticesactivationpatching,
      title={Towards Best Practices of Activation Patching in Language Models: Metrics and Methods}, 
      author={Fred Zhang and Neel Nanda},
      year={2024},
      eprint={2309.16042},
      archivePrefix={arXiv},
      primaryClass={cs.LG},
      url={https://arxiv.org/abs/2309.16042}, 
}

@misc{hanna2023doesgpt2computegreaterthan,
      title={How does GPT-2 compute greater-than?: Interpreting mathematical abilities in a pre-trained language model}, 
      author={Michael Hanna and Ollie Liu and Alexandre Variengien},
      year={2023},
      eprint={2305.00586},
      archivePrefix={arXiv},
      primaryClass={cs.CL},
      url={https://arxiv.org/abs/2305.00586}, 
}

@article{AI_Feynman, title={AI Feynman: a Physics-Inspired Method for Symbolic Regression}, url={http://arxiv.org/abs/1905.11481}, DOI={10.48550/arXiv.1905.11481}, abstractNote={A core challenge for both physics and artificial intellicence (AI) is symbolic regression: finding a symbolic expression that matches data from an unknown function. Although this problem is likely to be NP-hard in principle, functions of practical interest often exhibit symmetries, separability, compositionality and other simplifying properties. In this spirit, we develop a recursive multidimensional symbolic regression algorithm that combines neural network fitting with a suite of physics-inspired techniques. We apply it to 100 equations from the Feynman Lectures on Physics, and it discovers all of them, while previous publicly available software cracks only 71; for a more difficult test set, we improve the state of the art success rate from 15% to 90%.}, note={arXiv:1905.11481 [physics]}, number={arXiv:1905.11481}, publisher={arXiv}, author={Udrescu, Silviu-Marian and Tegmark, Max}, year={2020}, month=apr, language={en} }

@article{nnsight, title={NNsight and NDIF: Democratizing Access to Open-Weight Foundation Model Internals}, url={http://arxiv.org/abs/2407.14561}, DOI={10.48550/arXiv.2407.14561}, abstractNote={The enormous scale of state-of-the-art foundation models has limited their accessibility to scientists, because customized experiments on large models require costly hardware and complex engineering that is impractical for most researchers. To alleviate these problems, we introduce NNsight, an open-source Python package with a simple, flexible API that can express interventions on any PyTorch model by building computation graphs. We also introduce NDIF, a collaborative research platform providing researchers access to foundation-scale LLMs via the NNsight API. Code, documentation, and tutorials are available at https://nnsight.net/.}, note={arXiv:2407.14561 [cs]}, number={arXiv:2407.14561}, publisher={arXiv}, author={Fiotto-Kaufman, Jaden and Loftus, Alexander R. and Todd, Eric and Brinkmann, Jannik and Pal, Koyena and Troitskii, Dmitrii and Ripa, Michael and Belfki, Adam and Rager, Can and Juang, Caden and Mueller, Aaron and Marks, Samuel and Sharma, Arnab Sen and Lucchetti, Francesca and Prakash, Nikhil and Brodley, Carla and Guha, Arjun and Bell, Jonathan and Wallace, Byron C. and Bau, David}, year={2025}, month=jan, language={en} }

@article{DGSR, title={Deep Generative Symbolic Regression}, url={http://arxiv.org/abs/2401.00282}, DOI={10.48550/arXiv.2401.00282}, abstractNote={Symbolic regression (SR) aims to discover concise closed-form mathematical equations from data, a task fundamental to scientific discovery. However, the problem is highly challenging because closed-form equations lie in a complex combinatorial search space. Existing methods, ranging from heuristic search to reinforcement learning, fail to scale with the number of input variables. We make the observation that closed-form equations often have structural characteristics and invariances (e.g., the commutative law) that could be further exploited to build more effective symbolic regression solutions. Motivated by this observation, our key contribution is to leverage pre-trained deep generative models to capture the intrinsic regularities of equations, thereby providing a solid foundation for subsequent optimization steps. We show that our novel formalism unifies several prominent approaches of symbolic regression and offers a new perspective to justify and improve on the previous ad hoc designs, such as the usage of cross-entropy loss during pre-training. Specifically, we propose an instantiation of our framework, Deep Generative Symbolic Regression (DGSR). In our experiments, we show that DGSR achieves a higher recovery rate of true equations in the setting of a larger number of input variables, and it is more computationally efficient at inference time than state-of-the-art RL symbolic regression solutions.}, note={arXiv:2401.00282 [cs]}, number={arXiv:2401.00282}, publisher={arXiv}, author={Holt, Samuel and Qian, Zhaozhi and Schaar, Mihaela van der}, year={2023}, month=dec, language={en} }

@article{FunctionalFaitfullness, title={Functional Faithfulness in the Wild: Circuit Discovery with Differentiable Computation Graph Pruning}, url={http://arxiv.org/abs/2407.03779}, DOI={10.48550/arXiv.2407.03779}, abstractNote={In this paper, we introduce a comprehensive reformulation of the task known as Circuit Discovery, along with DiscoGP, a novel and effective algorithm based on differentiable masking for discovering circuits. Circuit discovery is the task of interpreting the computational mechanisms of language models (LMs) by dissecting their functions and capabilities into sparse subnetworks (circuits). We identified two major limitations in existing circuit discovery efforts: (1) a dichotomy between weight-based and connection-edge-based approaches forces researchers to choose between pruning connections or weights, thereby limiting the scope of mechanistic interpretation of LMs; (2) algorithms based on activation patching tend to identify circuits that are neither functionally faithful nor complete. The performance of these identified circuits is substaintially reduced, often resulting in near-random performance in isolation. Furthermore, the complement of the circuit—i.e., the original LM with the identified circuit removed — still retains adequate performance, indicating that essential components of a complete circuits are missed by existing methods. DiscoGP successfully addresses the two aforementioned issues and demonstrates state-of-theart faithfulness, completeness, and sparsity. The effectiveness of the algorithm and its novel structure open up new avenues of gathering new insights into the internal workings of generative AI.}, note={arXiv:2407.03779 [cs]}, number={arXiv:2407.03779}, publisher={arXiv}, author={Yu, Lei and Niu, Jingcheng and Zhu, Zining and Penn, Gerald}, year={2024}, month=jul, language={en} }

@article{iterativepatching, title={Towards Interpretable Sequence Continuation: Analyzing Shared Circuits in Large Language Models}, url={http://arxiv.org/abs/2311.04131}, DOI={10.48550/arXiv.2311.04131}, abstractNote={While transformer models exhibit strong capabilities on linguistic tasks, their complex architectures make them difficult to interpret. Recent work has aimed to reverse engineer transformer models into human-readable representations called circuits that implement algorithmic functions. We extend this research by analyzing and comparing circuits for similar sequence continuation tasks, which include increasing sequences of Arabic numerals, number words, and months. By applying circuit interpretability analysis, we identify a key sub-circuit in both GPT-2 Small and Llama-2-7B responsible for detecting sequence members and for predicting the next member in a sequence. Our analysis reveals that semantically related sequences rely on shared circuit subgraphs with analogous roles. Additionally, we show that this sub-circuit has effects on various mathrelated prompts, such as on intervaled circuits, Spanish number word and months continuation, and natural language word problems. Overall, documenting shared computational structures enables better model behavior predictions, identification of errors, and safer editing procedures. This mechanistic understanding of transformers is a critical step towards building more robust, aligned, and interpretable language models.}, note={arXiv:2311.04131 [cs]}, number={arXiv:2311.04131}, publisher={arXiv}, author={Lan, Michael and Torr, Philip and Barez, Fazl}, year={2024}, month=oct, language={en} }

@article{probing_important, title={Probing Classifiers: Promises, Shortcomings, and Advances}, volume={48}, ISSN={0891-2017, 1530-9312}, DOI={10.1162/coli_a_00422}, abstractNote={Abstract
            Probing classifiers have emerged as one of the prominent methodologies for interpreting and analyzing deep neural network models of natural language processing. The basic idea is simple—a classifier is trained to predict some linguistic property from a model’s representations—and has been used to examine a wide variety of models and properties. However, recent studies have demonstrated various methodological limitations of this approach. This squib critically reviews the probing classifiers framework, highlighting their promises, shortcomings, and advances.}, number={1}, journal={Computational Linguistics}, author={Belinkov, Yonatan}, year={2022}, month=apr, pages={207–219}, language={en} }

@misc{hanna2024faithfaithfulnessgoingcircuit,
      title={Have Faith in Faithfulness: Going Beyond Circuit Overlap When Finding Model Mechanisms}, 
      author={Michael Hanna and Sandro Pezzelle and Yonatan Belinkov},
      year={2024},
      eprint={2403.17806},
      archivePrefix={arXiv},
      primaryClass={cs.LG},
      url={https://arxiv.org/abs/2403.17806}, 
}

@article{elhage2021mathematical,
   title={A Mathematical Framework for Transformer Circuits},
   author={Elhage, Nelson and Nanda, Neel and Olsson, Catherine and Henighan, Tom and Joseph, Nicholas and Mann, Ben and Askell, Amanda and Bai, Yuntao and Chen, Anna and Conerly, Tom and DasSarma, Nova and Drain, Dawn and Ganguli, Deep and Hatfield-Dodds, Zac and Hernandez, Danny and Jones, Andy and Kernion, Jackson and Lovitt, Liane and Ndousse, Kamal and Amodei, Dario and Brown, Tom and Clark, Jack and Kaplan, Jared and McCandlish, Sam and Olah, Chris},
   year={2021},
   journal={Transformer Circuits Thread},
   note={https://transformer-circuits.pub/2021/framework/index.html}
}

@inproceedings{Garc_a_Carrasco_2024, series={IJCAI-2024},
   title={Detecting and Understanding Vulnerabilities in Language Models via Mechanistic Interpretability},
   url={http://dx.doi.org/10.24963/ijcai.2024/43},
   DOI={10.24963/ijcai.2024/43},
   booktitle={Proceedings of the Thirty-ThirdInternational Joint Conference on Artificial Intelligence},
   publisher={International Joint Conferences on Artificial Intelligence Organization},
   author={García-Carrasco, Jorge and Maté, Alejandro and Trujillo, Juan},
   year={2024},
   month=aug, pages={385–393},
   collection={IJCAI-2024} }

@article{Probing_correlation_1, title={Probing the Probing Paradigm: Does Probing Accuracy Entail Task Relevance?}, url={http://arxiv.org/abs/2005.00719}, DOI={10.48550/arXiv.2005.00719}, abstractNote={Although neural models have achieved impressive results on several NLP benchmarks, little is understood about the mechanisms they use to perform language tasks. Thus, much recent attention has been devoted to analyzing the sentence representations learned by neural encoders, through the lens of `probing’ tasks. However, to what extent was the information encoded in sentence representations, as discovered through a probe, actually used by the model to perform its task? In this work, we examine this probing paradigm through a case study in Natural Language Inference, showing that models can learn to encode linguistic properties even if they are not needed for the task on which the model was trained. We further identify that pretrained word embeddings play a considerable role in encoding these properties rather than the training task itself, highlighting the importance of careful controls when designing probing experiments. Finally, through a set of controlled synthetic tasks, we demonstrate models can encode these properties considerably above chance-level even when distributed in the data as random noise, calling into question the interpretation of absolute claims on probing tasks.}, note={arXiv:2005.00719 [cs]}, number={arXiv:2005.00719}, publisher={arXiv}, author={Ravichander, Abhilasha and Belinkov, Yonatan and Hovy, Eduard}, year={2021}, month=mar, language={en} }

@article{Probing_correlation_2, title={Amnesic Probing: Behavioral Explanation with Amnesic Counterfactuals}, volume={9}, ISSN={2307-387X}, DOI={10.1162/tacl_a_00359}, abstractNote={Abstract
            A growing body of work makes use of probing in order to investigate the working of neural models, often considered black boxes. Recently, an ongoing debate emerged surrounding the limitations of the probing paradigm. In this work, we point out the inability to infer behavioral conclusions from probing results, and offer an alternative method that focuses on how the information is being used, rather than on what information is encoded. Our method, Amnesic Probing, follows the intuition that the utility of a property for a given task can be assessed by measuring the influence of a causal intervention that removes it from the representation. Equipped with this new analysis tool, we can ask questions that were not possible before, for example, is part-of-speech information important for word prediction? We perform a series of analyses on BERT to answer these types of questions. Our findings demonstrate that conventional probing performance is not correlated to task importance, and we call for increased scrutiny of claims that draw behavioral or causal conclusions from probing results.1}, journal={Transactions of the Association for Computational Linguistics}, author={Elazar, Yanai and Ravfogel, Shauli and Jacovi, Alon and Goldberg, Yoav}, year={2021}, month=mar, pages={160–175}, language={en} }

@article{Probing_correlation_3, title={The Cognitive Revolution in Interpretability: From Explaining Behavior to Interpreting Representations and Algorithms}, url={http://arxiv.org/abs/2408.05859}, DOI={10.48550/arXiv.2408.05859}, abstractNote={Artiﬁcial neural networks have long been understood as “black boxes”: though we know their computation graphs and learned parameters, the knowledge encoded by these weights and functions they perform are not inherently interpretable. As such, from the early days of deep learning, there have been efforts to explain these models’ behavior and understand them internally; and recently, mechanistic interpretability (MI) has emerged as a distinct research area studying the features and implicit algorithms learned by foundation models such as large language models. In this work, we aim to ground MI in the context of cognitive science, which has long struggled with analogous questions in studying and explaining the behavior of “black box” intelligent systems like the human brain. We leverage several important ideas and developments in the history of cognitive science to disentangle divergent objectives in MI and indicate a clear path forward. First, we argue that current methods are ripe to facilitate a transition in deep learning interpretation echoing the “cognitive revolution” in 20th-century psychology that shifted the study of human psychology from pure behaviorism toward mental representations and processing. Second, we propose a taxonomy mirroring key parallels in computational neuroscience to describe two broad categories of MI research, semantic interpretation (what latent representations are learned and used) and algorithmic interpretation (what operations are performed over representations) to elucidate their divergent goals and objects of study. Finally, we elaborate the parallels and distinctions between various approaches in both categories, analyze the respective strengths and weaknesses of representative works, clarify underlying assumptions, outline key challenges, and discuss the possibility of unifying these modes of interpretation under a common framework.}, note={arXiv:2408.05859 [cs]}, number={arXiv:2408.05859}, publisher={arXiv}, author={Davies, Adam and Khakzar, Ashkan}, year={2024}, month=aug, language={en} }

@book{boyd2007tutorial,
  title={A tutorial on geometric programming},
  author={Boyd, Stephen and Kim, Seung-Jean and Vandenberghe, Lieven and Hassibi, Ali},
  year={2007},
  publisher={Stanford University}
}

@misc{tyto2023flighttime,
  author       = {{Tyto Robotics}},
  title        = {How to Increase Drone Flight Time and Lift Capacity},
  year         = {2023},
  howpublished = {\url{https://www.tytorobotics.com/blogs/articles/how-to-increase-drone-flight-time-and-lift-capacity}},
  note         = {Accessed: 2025-05-25}
}

@book{openstax2023pendulum,
  title        = {University Physics Volume 1},
  author       = {{OpenStax}},
  year         = {2023},
  publisher    = {OpenStax},
  url          = {https://openstax.org/books/university-physics-volume-1/pages/15-4-pendulums},
  note         = {Accessed: 2025-05-25}
}

@misc{hansen2023cmaevolutionstrategytutorial,
      title={The CMA Evolution Strategy: A Tutorial}, 
      author={Nikolaus Hansen},
      year={2023},
      eprint={1604.00772},
      archivePrefix={arXiv},
      primaryClass={cs.LG},
      url={https://arxiv.org/abs/1604.00772}, 
}

@inproceedings{auger2005restart,
  title={Restart CMA evolution strategies with increasing population size},
  author={Auger, Anne and Hansen, Nikolaus},
  booktitle={2005 IEEE Congress on Evolutionary Computation},
  volume={2},
  pages={1769--1776},
  year={2005},
  organization={IEEE}
}

@book{fletcher1987practical,
  title     = {Practical Methods of Optimization},
  author    = {Fletcher, R.},
  edition   = {2},
  year      = {1987},
  publisher = {John Wiley \& Sons},
  address   = {New York, NY, USA}
}

@article{Vig_Gehrmann_Belinkov_Qian_Nevo_Singer_Shieber, title={Investigating Gender Bias in Language Models Using Causal Mediation Analysis}, abstractNote={Many interpretation methods for neural models in natural language processing investigate how information is encoded inside hidden representations. However, these methods can only measure whether the information exists, not whether it is actually used by the model. We propose a methodology grounded in the theory of causal mediation analysis for interpreting which parts of a model are causally implicated in its behavior. The approach enables us to analyze the mechanisms that facilitate the ﬂow of information from input to output through various model components, known as mediators. As a case study, we apply this methodology to analyzing gender bias in pre-trained Transformer language models. We study the role of individual neurons and attention heads in mediating gender bias across three datasets designed to gauge a model’s sensitivity to gender bias. Our mediation analysis reveals that gender bias effects are concentrated in speciﬁc components of the model that may exhibit highly specialized behavior.}, author={Vig, Jesse and Gehrmann, Sebastian and Belinkov, Yonatan and Qian, Sharon and Nevo, Daniel and Singer, Yaron and Shieber, Stuart}, language={en} }

@article{Heimersheim_Nanda_2024, title={How to use and interpret activation patching}, url={http://arxiv.org/abs/2404.15255}, DOI={10.48550/arXiv.2404.15255}, abstractNote={Activation patching is a popular mechanistic interpretability technique, but has many subtleties regarding how it is applied and how one may interpret the results. We provide a summary of advice and best practices, based on our experience using this technique in practice. We include an overview of the different ways to apply activation patching and a discussion on how to interpret the results. We focus on what evidence patching experiments provide about circuits, and on the choice of metric and associated pitfalls.}, note={arXiv:2404.15255 [cs]}, number={arXiv:2404.15255arXiv:2404.15255}, publisher={arXiv}, author={Heimersheim, Stefan and Nanda, Neel}, year={2024}, month=apr, language={en} }

@incollection{Ribeiro2016WhyTrust,
  author    = {Marco Tulio Ribeiro and Sameer Singh and Carlos Guestrin},
  title     = {“Why Should I Trust You?”: Explaining the Predictions of Any Classifier},
  booktitle = {Proceedings of the 22nd ACM SIGKDD International Conference on Knowledge Discovery and Data Mining},
  year      = {2016},
  pages     = {1135--1144},
  publisher = {ACM},
  doi       = {10.1145/2939672.2939778},
  url       = {https://doi.org/10.1145/2939672.2939778}
}

@book{Molnar2019InterpretableML,
  author    = {Christoph Molnar},
  title     = {Interpretable Machine Learning: A Guide for Making Black Box Models Explainable},
  publisher = {Lulu.com},
  year      = {2019},
  note      = {3rd edition available online at https://christophm.github.io/interpretable-ml-book/}
}

@inproceedings{LundbergLee2017SHAP,
  author    = {Scott M. Lundberg and Su-In Lee},
  title     = {A Unified Approach to Interpreting Model Predictions},
  booktitle = {Advances in Neural Information Processing Systems 30 (NeurIPS 2017)},
  year      = {2017},
  pages     = {4765--4774},
  url       = {https://arxiv.org/abs/1705.07874}
}

@article{meurer2017sympy,
  title={SymPy: symbolic computing in Python},
  author={Meurer, Aaron and Smith, Christopher P and Paprocki, Mateusz and Čertík, Ondřej and Kirpichev, Sergey B and Rocklin, Matthew and Kumar, Amit and Ivanov, Sergiu and Moore, Jason K and Singh, Sartaj and Rathnayake, Tal and Vig, Sylvain L and Granger, Brian E and Muller, Richard and Bonazzi, Francesco and Gupta, Harsh and Vats, Shivam and Johansson, Fredrik},
  journal={PeerJ Computer Science},
  volume={3},
  pages={e103},
  year={2017},
  publisher={PeerJ Inc.},
  doi={10.7717/peerj-cs.103}
}

@article{Meng_Bau_Andonian_Belinkov_2023, title={Locating and Editing Factual Associations in GPT}, url={http://arxiv.org/abs/2202.05262}, DOI={10.48550/arXiv.2202.05262}, abstractNote={We analyze the storage and recall of factual associations in autoregressive transformer language models, ﬁnding evidence that these associations correspond to localized, directly-editable computations. We ﬁrst develop a causal intervention for identifying neuron activations that are decisive in a model’s factual predictions. This reveals a distinct set of steps in middle-layer feed-forward modules that mediate factual predictions while processing subject tokens. To test our hypothesis that these computations correspond to factual association recall, we modify feedforward weights to update speciﬁc factual associations using Rank-One Model Editing (ROME). We ﬁnd that ROME is effective on a standard zero-shot relation extraction (zsRE) model-editing task. We also evaluate ROME on a new dataset of difﬁcult counterfactual assertions, on which it simultaneously maintains both speciﬁcity and generalization, whereas other methods sacriﬁce one or another. Our results conﬁrm an important role for mid-layer feed-forward modules in storing factual associations and suggest that direct manipulation of computational mechanisms may be a feasible approach for model editing. The code, dataset, visualizations, and an interactive demo notebook are available at https://rome.baulab.info/.}, note={arXiv:2202.05262 [cs]}, number={arXiv:2202.05262}, publisher={arXiv}, author={Meng, Kevin and Bau, David and Andonian, Alex and Belinkov, Yonatan}, year={2023}, month=jan, language={en} }

@inproceedings{hewitt-manning-2019-structural,
    title = "{A} Structural Probe for Finding Syntax in Word Representations",
    author = "Hewitt, John  and
      Manning, Christopher D.",
    editor = "Burstein, Jill  and
      Doran, Christy  and
      Solorio, Thamar",
    booktitle = "Proceedings of the 2019 Conference of the North {A}merican Chapter of the Association for Computational Linguistics: Human Language Technologies, Volume 1 (Long and Short Papers)",
    month = jun,
    year = "2019",
    address = "Minneapolis, Minnesota",
    publisher = "Association for Computational Linguistics",
    url = "https://aclanthology.org/N19-1419/",
    doi = "10.18653/v1/N19-1419",
    pages = "4129--4138"
}

@article{Wang_Fu_et_al._2023, title={Scientific discovery in the age of artificial intelligence}, volume={620}, ISSN={0028-0836, 1476-4687}, DOI={10.1038/s41586-023-06221-2}, number={7972}, journal={Nature}, author={Wang, Hanchen and Fu, Tianfan and Du, Yuanqi and Gao, Wenhao and Huang, Kexin and Liu, Ziming and Chandak, Payal and Liu, Shengchao and Van Katwyk, Peter and Deac, Andreea and Anandkumar, Anima and Bergen, Karianne and Gomes, Carla P. and Ho, Shirley and Kohli, Pushmeet and Lasenby, Joan and Leskovec, Jure and Liu, Tie-Yan and Manrai, Arjun and Marks, Debora and Ramsundar, Bharath and Song, Le and Sun, Jimeng and Tang, Jian and Veličković, Petar and Welling, Max and Zhang, Linfeng and Coley, Connor W. and Bengio, Yoshua and Zitnik, Marinka}, year={2023}, month=aug, pages={47–60}, language={en} }

@article{Cui_Qi_Zhou_Yu_Wang_Zhang_Zhang_Wang_Liu_2025, title={Artificial intelligence and food flavor: How AI models are shaping the future and revolutionary technologies for flavor food development}, volume={24}, ISSN={1541-4337, 1541-4337}, DOI={10.1111/1541-4337.70068}, abstractNote={The food flavor science, traditionally reliant on experimental methods, is now entering a promising era with the help of artificial intelligence (AI). By integrating existing technologies with AI, researchers can explore and develop new flavor substances in a digital environment, saving time and resources. More and more research will use AI and big data to enhance product flavor, improve product quality, meet consumer needs, and drive the industry toward a smarter and more sustainable future. In this review, we elaborate on the mechanisms of flavor recognition and their potential impact on nutritional regulation. With the increase of data accumulation and the development of internet information technology, food flavor databases and food ingredient databases have made great progress. These databases provide detailed information on the nutritional content, flavor molecules, and chemical properties of various food compounds, providing valuable data support for the rapid evaluation of flavor components and the construction of screening technology. With the popularization of AI in various fields, the field of food flavor has also ushered in new development opportunities. This review explores the mechanisms of flavor recognition and the role of AI in enhancing food flavor analysis through high-throughput omics data and screening technologies. AI algorithms offer a pathway to scientifically improve product formulations, thereby enhancing flavor and customized meals. Furthermore, it discusses the safety challenges of integrating AI into the food flavor industry.}, number={1}, journal={Comprehensive Reviews in Food Science and Food Safety}, author={Cui, Zhiyong and Qi, Chengliang and Zhou, Tianxing and Yu, Yanyang and Wang, Yueming and Zhang, Zhiwei and Zhang, Yin and Wang, Wenli and Liu, Yuan}, year={2025}, month=jan, pages={e70068}, language={en} }

@article{Liu_Mao_Wen_2025, title={How do Large Language Models Understand Relevance? A Mechanistic Interpretability Perspective}, url={http://arxiv.org/abs/2504.07898}, DOI={10.48550/arXiv.2504.07898}, abstractNote={Recent studies have shown that large language models (LLMs) can assess relevance and support information retrieval (IR) tasks such as document ranking and relevance judgment generation. However, the internal mechanisms by which off-the-shelf LLMs understand and operationalize relevance remain largely unexplored. In this paper, we systematically investigate how different LLM modules contribute to relevance judgment through the lens of mechanistic interpretability. Using activation patching techniques, we analyze the roles of various model components and identify a multi-stage, progressive process in generating either pointwise or pairwise relevance judgment. Specifically, LLMs first extract query and document information in the early layers, then process relevance information according to instructions in the middle layers, and finally utilize specific attention heads in the later layers to generate relevance judgments in the required format. Our findings provide insights into the mechanisms underlying relevance assessment in LLMs, offering valuable implications for future research on leveraging LLMs for IR tasks.}, note={arXiv:2504.07898 [cs]}, number={arXiv:2504.07898arXiv:2504.07898}, publisher={arXiv}, author={Liu, Qi and Mao, Jiaxin and Wen, Ji-Rong}, year={2025}, month=apr, language={en} }

@article{Park_Lee_2025, title={ICLR: In-Context Learning of Representations}, url={http://arxiv.org/abs/2501.00070}, DOI={10.48550/arXiv.2501.00070}, abstractNote={Recent work has demonstrated that semantics specified by pretraining data influence how representations of different concepts are organized in a large language model (LLM). However, given the open-ended nature of LLMs, e.g., their ability to in-context learn, we can ask whether models alter these pretraining semantics to adopt alternative, context-specified ones. Specifically, if we provide in-context exemplars wherein a concept plays a different role than what the pretraining data suggests, do models reorganize their representations in accordance with these novel semantics? To answer this question, we take inspiration from the theory of conceptual role semantics and define a toy “graph tracing” task wherein the nodes of the graph are referenced via concepts seen during training (e.g., apple, bird, etc.) and the connectivity of the graph is defined via some predefined structure (e.g., a square grid). Given exemplars that indicate traces of random walks on the graph, we analyze intermediate representations of the model and find that as the amount of context is scaled, there is a sudden re-organization from pretrained semantic representations to in-context representations aligned with the graph structure. Further, we find that when reference concepts have correlations in their semantics (e.g., Monday, Tuesday, etc.), the context-specified graph structure is still present in the representations, but is unable to dominate the pretrained structure. To explain these results, we analogize our task to energy minimization for a predefined graph topology, providing evidence towards an implicit optimization process to infer context-specified semantics. Overall, our findings indicate scaling context-size can flexibly re-organize model representations, possibly unlocking novel capabilities.}, note={arXiv:2501.00070 [cs]}, number={arXiv:2501.00070}, publisher={arXiv}, author={Park, Core Francisco and Lee, Andrew and Lubana, Ekdeep Singh and Yang, Yongyi and Okawa, Maya and Nishi, Kento and Wattenberg, Martin and Tanaka, Hidenori}, year={2025}, month=may, language={en} }

@article{Shojaee_Meidani_Farimani_Reddy_2023, title={Transformer-based Planning for Symbolic Regression}, url={http://arxiv.org/abs/2303.06833}, DOI={10.48550/arXiv.2303.06833}, abstractNote={Symbolic regression (SR) is a challenging task in machine learning that involves finding a mathematical expression for a function based on its values. Recent advancements in SR have demonstrated the effectiveness of pre-trained transformerbased models in generating equations as sequences, leveraging large-scale pretraining on synthetic datasets and offering notable advantages in terms of inference time over classical Genetic Programming (GP) methods. However, these models primarily rely on supervised pre-training goals borrowed from text generation and overlook equation discovery objectives like accuracy and complexity. To address this, we propose TPSR, a Transformer-based Planning strategy for Symbolic Regression that incorporates Monte Carlo Tree Search into the transformer decoding process. Unlike conventional decoding strategies, TPSR enables the integration of non-differentiable feedback, such as fitting accuracy and complexity, as external sources of knowledge into the transformer-based equation generation process. Extensive experiments on various datasets show that our approach outperforms state-of-the-art methods, enhancing the model’s fitting-complexity trade-off, extrapolation abilities, and robustness to noise 2 .}, note={arXiv:2303.06833 [cs]}, number={arXiv:2303.06833}, publisher={arXiv}, author={Shojaee, Parshin and Meidani, Kazem and Farimani, Amir Barati and Reddy, Chandan K.}, year={2023}, month=oct, language={en} }

@article{SymFormer, title={SymFormer: End-to-end symbolic regression using transformer-based architecture}, url={http://arxiv.org/abs/2205.15764}, DOI={10.48550/arXiv.2205.15764}, abstractNote={Many real-world problems can be naturally described by mathematical formulas. The task of ﬁnding formulas from a set of observed inputs and outputs is called symbolic regression. Recently, neural networks have been applied to symbolic regression, among which the transformer-based ones seem to be the most promising. After training the transformer on a large number of formulas (in the order of days), the actual inference, i.e., ﬁnding a formula for new, unseen data, is very fast (in the order of seconds). This is considerably faster than state-of-the-art evolutionary methods. The main drawback of transformers is that they generate formulas without numerical constants, which have to be optimized separately, so yielding suboptimal results. We propose a transformer-based approach called SymFormer, which predicts the formula by outputting the individual symbols and the corresponding constants simultaneously. This leads to better performance in terms of ﬁtting the available data. In addition, the constants provided by SymFormer serve as a good starting point for subsequent tuning via gradient descent to further improve the performance. We show on a set of benchmarks that SymFormer outperforms two state-of-the-art methods while having faster inference.}, note={arXiv:2205.15764 [cs]}, number={arXiv:2205.15764}, publisher={arXiv}, author={Vastl, Martin and Kulhánek, Jonáš and Kubalík, Jiří and Derner, Erik and Babuška, Robert}, year={2022}, month=oct, language={en} }
\bibliographystyle{icml2026}

\newpage
\appendix
\onecolumn

\newpage
\section{Neural Symbolic Regression that Scales}
\label{APP:NeSymReS}
Recent work has extended the Transformer architecture to SR. Neural Symbolic Regression that Scales (NeSymReS), introduced by Biggio et al.~\cite{Neural_symbolic_regression_that_scales}, is the first large-scale pre-trained Transformer model designed for SR. This thesis uses NeSymReS due to its pre-trained weights and well-documented open-source code. While newer models have been proposed, many build on NeSymReS and adopt similar architectures, often with less accessible or poorly documented codebases. To give a solid understanding of the model, the next paragraphs will describe the pipeline, encoder-decoder architecture, as well as the training setup of NeSymReS.

\paragraph{Pipeline} To uncover equations that represent the input \((X, y)\) pairs, the NeSymReS pipeline follows four stages for each equation. First, a ground-truth infix expression \( f^* \) is generated using the framework from Lample et al. \cite{func_gen}. Based on this framework, a skeleton representing the symbolic structure of the expression is sampled. The skeleton is the full equation with constants abstracted as variables (e.g., \( \texttt{sin}(x_1) + c_1 \cdot x_2 \)). These constants are then instantiated with sampled values from predefined ranges.  
Second, the completed expression is evaluated over a range of input values to generate input-output pairs \((X, y)\).
Third, these observations are encoded by NeSymReS into a latent representation, which the decoder uses to auto-regressively generate a prefix skeleton formula.  
Finally, during inference, beam search identifies the most likely candidate equations, and the constants are refined using the Broyden–Fletcher–Goldfarb–Shanno (BFGS) algorithm, to produce the final predicted expression \( f^{\texttt{pred}} \).

\paragraph{Encoder Architecture}\label{Encoder Architecture} The encoder used in NeSymReS is a \textbf{set encoder}, introduced by \citeauthor{set_encoder}, designed to process unordered sets. Set encoders ensure \textit{permutation equivariance}: reordering the input rows does not affect the output representation. These encoders are particularly well-suited to symbolic regression tasks, where the order of input examples should not influence the model’s predictions.

To address the quadratic complexity \(\mathcal{O}(n^2)\) of standard self-attention (with respect to the input size \(n\)), the set encoder uses a more efficient attention mechanism based on \textit{trainable inducing points}. These inducing points act as a learned, fixed-size bottleneck through which input elements interact. By projecting the input set into a smaller, intermediate set of \(m\) inducing points, the attention computation is reduced to \(\mathcal{O}(n \cdot m)\), which significantly improves scalability. 

The NeSymReS encoder is composed of six stacked layers, each implementing an \textit{Induced Set Attention Block (ISAB)}. ISABs introduce the set of trainable \textit{inducing points} \(\mathbf{I} \in \mathbb{R}^{m \times d}\), with each ISAB containing two \textit{Multihead Attention Blocks (MABs)}, applied sequentially:
\begin{align*}
    \mathrm{ISAB}(\mathbf{X}) = \mathrm{MAB}(\mathbf{X},\, \mathbf{H}) \quad \text{where} \quad \mathbf{H} = \mathrm{MAB}(\mathbf{I},\, \mathbf{X})
\end{align*}
First, the inducing points attend to the inputs to produce a condensed representation \(\mathbf{H}\). Then, each input token attends to \(\mathbf{H}\), allowing information to flow indirectly between all input elements via the shared inducing set.

Each MAB uses eight attention heads followed by a feedforward block. The encoder operates with a hidden dimensionality of 512 and uses 50 inducing points. See \fullcref{fig:set_encoder_diagram} for an illustration of a single ISAB layer.

\begin{figure}[ht!]
    \centering
    \includegraphics[width=1.00\linewidth]{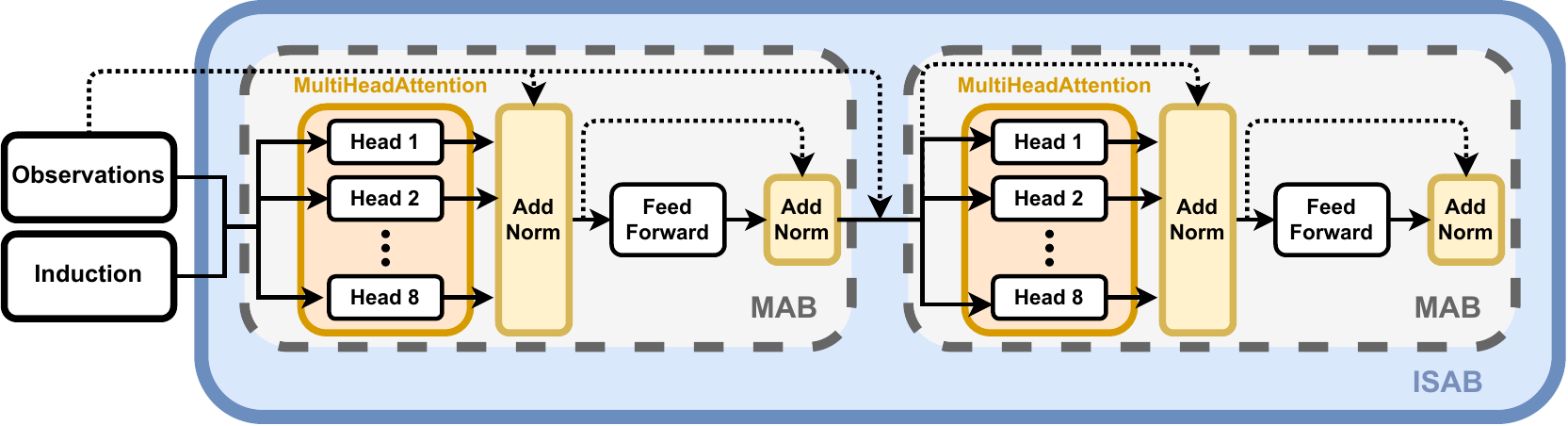}
    \caption{Diagram of a set-encoder architecture. Each ISAB layer (blue) contains two MAB blocks (grey), which consist of MultiHeadAttention (orange) and Feed Forward modules with residual connections (dashed arrows).}
    \label{fig:set_encoder_diagram}
\end{figure}

\paragraph{Decoder Architecture} The decoder in NeSymReS is a standard transformer decoder that autoregressively generates mathematical expressions. During inference, at each timestep, $n$ candidate solutions are generated, where $n$ is twice the beam size. For each candidate, the probability of the most likely character (from the set of all output symbols) is computed and added to the current beam score, which represents the cumulative logit score of the sequence decoded so far. The $n / 2$ highest-scoring candidates are retained for the next generation step, while the rest are discarded. If the end token (\texttt{<F>}) is generated, the corresponding equation is checked for validity and, if valid, added to a list of potential hypotheses. Once the maximum equation length is reached, constants in the hypotheses are optimised using the BFGS algorithm and compared against the true $y$-values. Formulas with the smallest deviation between predicted ($y_{\text{pred}}$) and true ($y_{\text{true}}$) values are returned to the user.

\paragraph{Pre-Training} NeSymReS is trained on 100 million equations\footnote{While NeSymReS was also trained on smaller datasets, this particular model achieved the best performance according to Biggio et al. \cite{Neural_symbolic_regression_that_scales} and is therefore used in this work.}. The training dataset contains 1.2 million unique skeletons with an average length of 8.2 operators. The model’s task is to predict the prefix-notated formula that corresponds to the given observations. During training, the decoder receives the ground truth previous tokens rather than generating the sequence autoregressively. The model is trained with minibatches of size 150, using a negative log-likelihood loss function for 23 epochs over three days on an RTX 2080.

\paragraph{Gradient-Based Search with BFGS} \label{BFGS_explaination}  
In SR, models often predict expressions containing both structural components (e.g., operators and variables) and numerical constants. While structure can be learned using neural networks such as transformers, the precise values of constants typically cannot be predicted directly due to the limited output vocabulary. To address this, NeSymReS treats constants as free parameters \(c\) and optimises them post-hoc using gradient-based methods.

One common approach is to apply the BFGS (Broyden–Fletcher–Goldfarb–Shanno) algorithm, a quasi-Newton method for unconstrained non-linear optimization by \citeauthor{fletcher1987practical}. BFGS is particularly well-suited for this task due to its efficiency and ability to approximate the Hessian matrix, enabling faster convergence than basic gradient descent.

Given a symbolic expression $f(x; \theta)$, where $\theta$ denotes the set of constant parameters in the expression, we define a loss function over a dataset $\{(x_i, y_i)\}_{i=1}^N$:

\[
\mathcal{L}(\theta) = \frac{1}{N} \sum_{i=1}^{N} \left(f(x_i; \theta) - y_i\right)^2
\]

This loss measures how well the expression with current constants fits the target data. The BFGS algorithm iteratively updates $\theta$ to minimize $\mathcal{L}(\theta)$, using both gradient information and an approximation of the inverse Hessian:

\[
\theta_{k+1} = \theta_k - H_k \nabla \mathcal{L}(\theta_k)
\]

where $H_k$ is an approximation to the inverse of the Hessian at step $k$, and $\nabla \mathcal{L}(\theta_k)$ is the gradient of the loss with respect to the constants.
\newpage
\section{Covariance Matrix Adaptation Evolution Strategy}\label{Appendix:CMA-ES}
This section further introduces the Covariance Matrix Adaptation Evolution Strategy (CMA-ES), a core component of our \patches\;circuit discovery method. To provide the necessary context, we briefly review the foundations of evolutionary computing and explain the core principles of CMA-ES.

Evolutionary computing (EC) is a family of optimalisation algorithms based on natural selection. A typical evolutionary algorithm is iterative and returns a population of candidate solutions after every generation. These get evaluated on their fitness (performance on the task) and with this information a selection is made on the best candidates. These candidates are then recombined with each other and mutated to create a new generation of candidates. Unlike reinforcement learning which sometimes requires gradients or intermediate rewards, evolutionary strategies solely operate on the fitness evaluation making them particularly suitable for optimising complex, non-differentiable objectives, such as discovering sparse circuits in large models. 

One such EC algorithm is CMA-ES, a stochastic optimisation method for non-convex, high-dimensional search spaces. The algorithm adapts a covariance matrix $\mathbf{C}$ to capture parameter correlations \cite{hansen2023cmaevolutionstrategytutorial, auger2005restart}. CMA-ES maintains a multivariate Gaussian distribution $\mathcal{N}(\mathbf{m}, \sigma^2 \mathbf{C})$, where $\mathbf{m}$ is the mean vector representing the current estimate of the optimal solution, $\sigma$ is the global step size\footnote{The step size in CMA-ES serves a similar purpose to the learning rate in deep learning; it determines the scale of updates during optimisation.}, and $\mathbf{C}$ encodes the shape and orientation of the search distribution.

In each generation, $\lambda$ candidate solutions $\mathbf{x}_i$ are sampled from the distribution. After evaluating their fitness, the mean vector $\mathbf{m}$ is updated through a weighted average of the top-performing candidates. This guides the search towards better regions of the solution space. The covariance matrix $\mathbf{C}$ is then updated as:

\begin{equation*}
\mathbf{C} = (1 - c_\text{cov}) \mathbf{C} + c_\text{cov} \mathbf{p}_c \mathbf{p}_c^\top + c_\text{cov} \sum_{i=1}^\lambda w_i \frac{(\mathbf{x}_i - \mathbf{m}')(\mathbf{x}_i - \mathbf{m}')^\top}{\sigma^2}
\end{equation*}

Here, $c_\text{cov}$ controls how quickly the covariance adapts, and $\mathbf{p}_c$ is the evolution path, or momentum. It accumulates the directions over generations and is updated as:

\begin{equation*}
\mathbf{p}_c = (1 - c_c) \mathbf{p}_c + \sqrt{c_c(2 - c_c)} \cdot \frac{\mathbf{m} - \mathbf{m}'}{\sigma}
\end{equation*}
Additionally, the global step size $\sigma$ is also adapted, based on how far the search is progressing. This is controlled by the conjugate evolution path $\mathbf{p}_\sigma$:
\begin{equation*}
\sigma = \sigma \cdot \exp \left( 
\frac{\|\mathbf{p_\sigma}\|}{\sqrt{1 - \left(1 - c_\sigma \right)^{2 \cdot (t + 1)}}} 
- 1 \right) \cdot \frac{1}{d_\sigma}
\end{equation*} 
\begin{equation*}
\mathbf{p}_\sigma = (1 - c_\sigma) \mathbf{p}_\sigma + \sqrt{c_\sigma(2 - c_\sigma)} \cdot \mathbf{C}^{-1/2}\frac{\mathbf{m} - \mathbf{m}'}{\sigma}
\end{equation*}
This reflects how CMA-ES adapts its step size: increasing it when progress is consistent, and reducing when the search needs to focus on fine-tuning.

\newpage
\section{Model Performance}
\label{APP: Model Performance}
To evaluate NeSymReS performance, we use the uploaded 100Mil pretrained model weights\footnote{Retrieved from: https://drive.google.com/drive/folders/1LTKUX-KhoUbW-WOx-ZJ8KitxK7Nov41G}. Generating the equations is supported by both beam search and the BFGS algorithm, whose individual contributions to performance are assessed. To validate the findings reported by  \citeauthor{Neural_symbolic_regression_that_scales}, we replicate their methodology as closely as possible and evaluate performance on the Feynman dataset \cite{AI_Feynman}.

\begin{table}[ht!]
\centering
\caption{Comparison of true and model-generated equations by reconstruction strategies.}
\label{tab:True vs recreated equations}
\begin{tabular}{@{} l l l @{}}
\toprule
Evaluation Type & True Equation & Recreated Equation \\ \midrule
Skeleton      & $2 \cdot x_1 - \dfrac{x_2}{x_3}$       & $2 \cdot x_1 - \dfrac{x_2}{x_3}$ \\[3mm]
Inserting 0   & $\dfrac{x_1}{x_2^4}$                     & $\dfrac{x_1}{x_2^4 \cdot (c+1)}$ \\[5mm]
Inserting 1   & $-\sin(x_2) + \tan(x_1^2)$              & $-\sin(c \cdot x_2) + \tan(x_1^2)$ \\[1mm]
Inserting 0/1 & $\cos\Bigl(x_1 \cdot (x_2+1)\Bigr)$       & $\cos\Bigl(x_1 \cdot (c \cdot x_1 + c + x_2)\Bigr)$ \\[2mm]
Point eval    & $x_1 \cdot \bigl(x_1 \cdot x_2 + x_1 + 1\bigr)$ & $x_1 \cdot \bigl(x_1 \cdot (x_2+1) + 1\bigr)$ \\ \bottomrule
\end{tabular}
\end{table}

\label{explaining model performance stages}
Model performance is evaluated in four stages. First, we assess the raw skeleton output, equations where constants are replaced by variables, by simplifying both the predicted and true equations using Sympy \cite{meurer2017sympy}, without applying BFGS. Second, manual inspection displayed that the model often captures the correct structure but tends to over-generate constants; replacing these with zero or one and simplifying frequently recovers the correct formula. If the first two steps do not recover the correct formula, we perform pointwise evaluation by providing 200 inputs into both the true and predicted expressions to test for equivalence. Finally, we apply BFGS to optimise constants, although this step is slow, CPU-bound, and does not reflect the model’s actual generative capabilities. Examples are shown in \fullcref{tab:True vs recreated equations}.

\subsection{Performance on Equations With and Without Constants}

To evaluate model performance, we use two datasets: one without constants, where the model can recover the correct formula without BFGS, and one with floating-point constants, which often requires BFGS to fill in missing values. The beam size (32) and BFGS setup (10 restarts) follow the configuration from NeSymReS~\cite{func_gen, Neural_symbolic_regression_that_scales}. Due to BFGS’s computational cost, we evaluated 10\,000 samples for the no-constant dataset and 1\,000 for the constant dataset.

\begin{figure}[h]
    \centering
    \begin{subfigure}{0.49\textwidth}
        \centering
        \includegraphics[width=\linewidth]{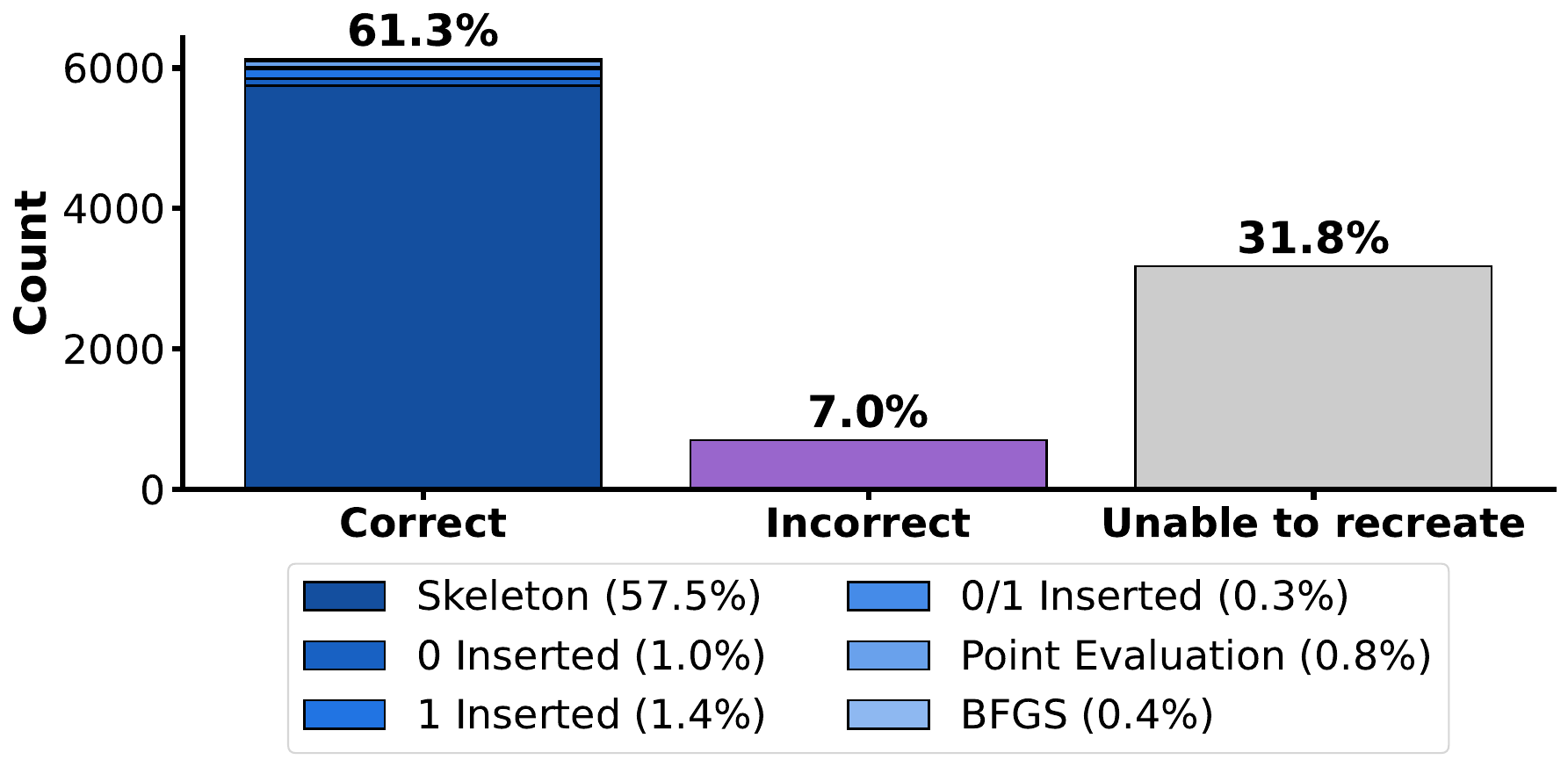}
        \caption{}
        \label{fig:MP 10.000 samples}
    \end{subfigure}
    \begin{subfigure}{0.49\textwidth}
        \centering
        \includegraphics[width=\linewidth]{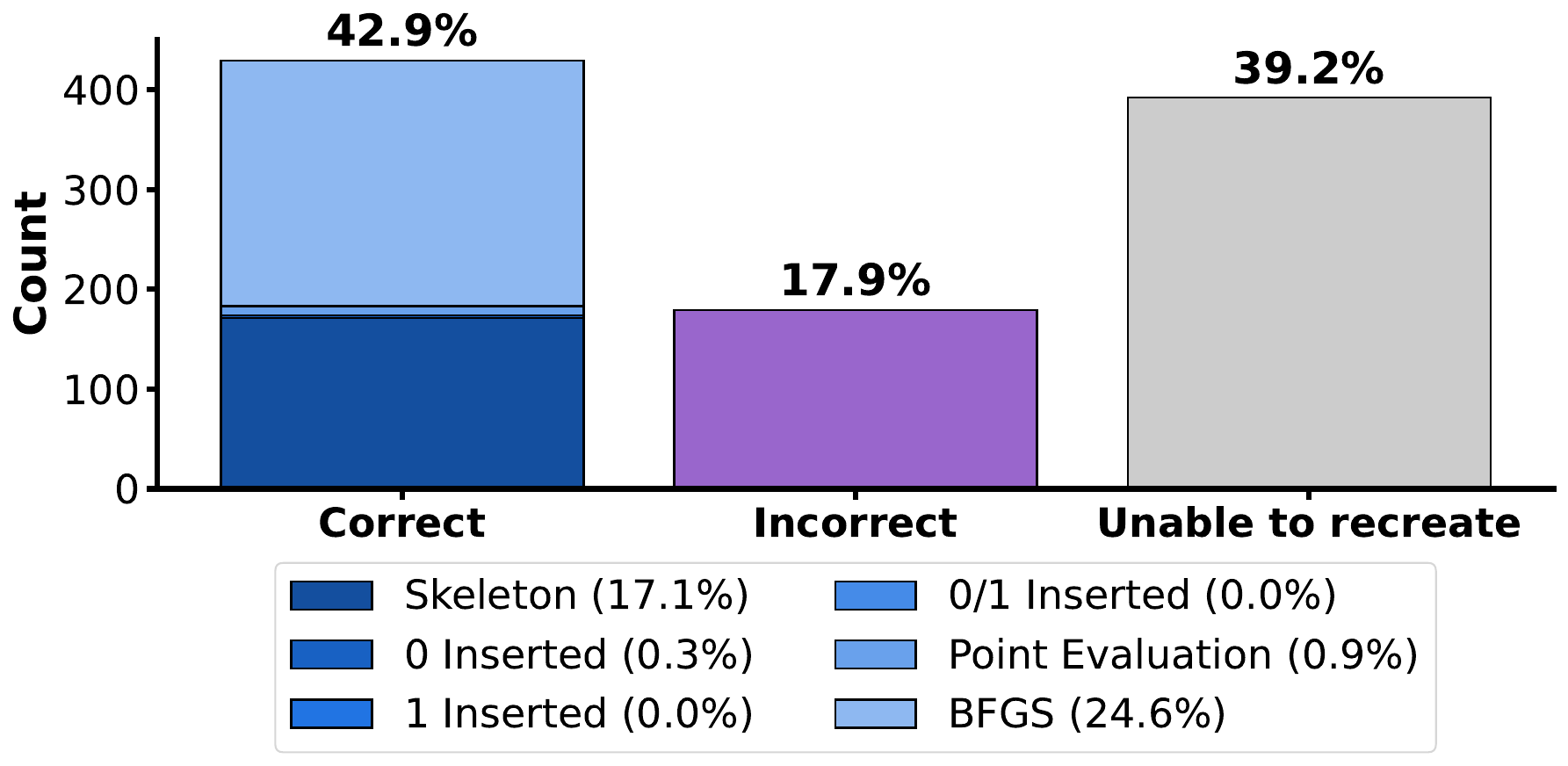}        
        \caption{}
        \label{fig:MP 1000 Samples}
    \end{subfigure}
    \caption{Comparison of model performance with beam size 32, showing distributions of correct, incorrect, and unrecreatable samples. The correct category is subdivided by reconstruction strategy. \textbf{(a)} 10\,000 samples without constants. \textbf{(b)} 1\,000 samples with constants.
}
    \label{fig:MP with and without constants}
\end{figure}
The results in \fullcref{fig:MP with and without constants} display that a substantial portion of true formulas could not be recreated: \(31.8\%\) without constants and \(39.2\%\) with constants. This suggests that in many cases the decoder returned only the start token \texttt{<S>} or invalid prefix notation. \fullcref{fig:MP 10.000 samples} shows that \(61.3\%\) of the 10\,000 formulas were correctly reconstructed, with strong skeleton performance. Substituting constants with zero or one yields a modest \(2.7\%\) improvement, reflecting the model’s tendency to overgenerate constants.

\fullcref{fig:MP 1000 Samples} shows performance drops with floating-point constants: only \(42.9\%\) of formulas are correctly recreated, and the error rate among incorrect outputs increases. BFGS accounts for over half of the final performance, as the model cannot generate most constants directly. These results indicate that while the model effectively recovers structural skeletons, its reliance on BFGS to generate constants reveals a gap between NeSymReS constant generation and the refining of BFGS. Since BFGS is a post-processing step and not part of the model, its contribution reflects an external correction rather than learned numerical understanding. Therefore for further experiments we will utilize the dataset without constants.

\subsection{Effect of Beam Size on Model Performance}

To assess the impact of beam search, the model was evaluated across beam sizes ranging from \(2^0\) to \(2^5\). Due to computational constraints, this experiment was conducted on 1\,000 datapoints. The results are shown in \fullcref{fig:MP different beam sizes.}, with a more detailed view of the \(2^5\) beam size in \fullcref{fig:MP with and without constants}.

\begin{figure}[ht!]
    \centering
    \begin{subfigure}{0.48\textwidth}
        \centering
        \includegraphics[width=\linewidth]{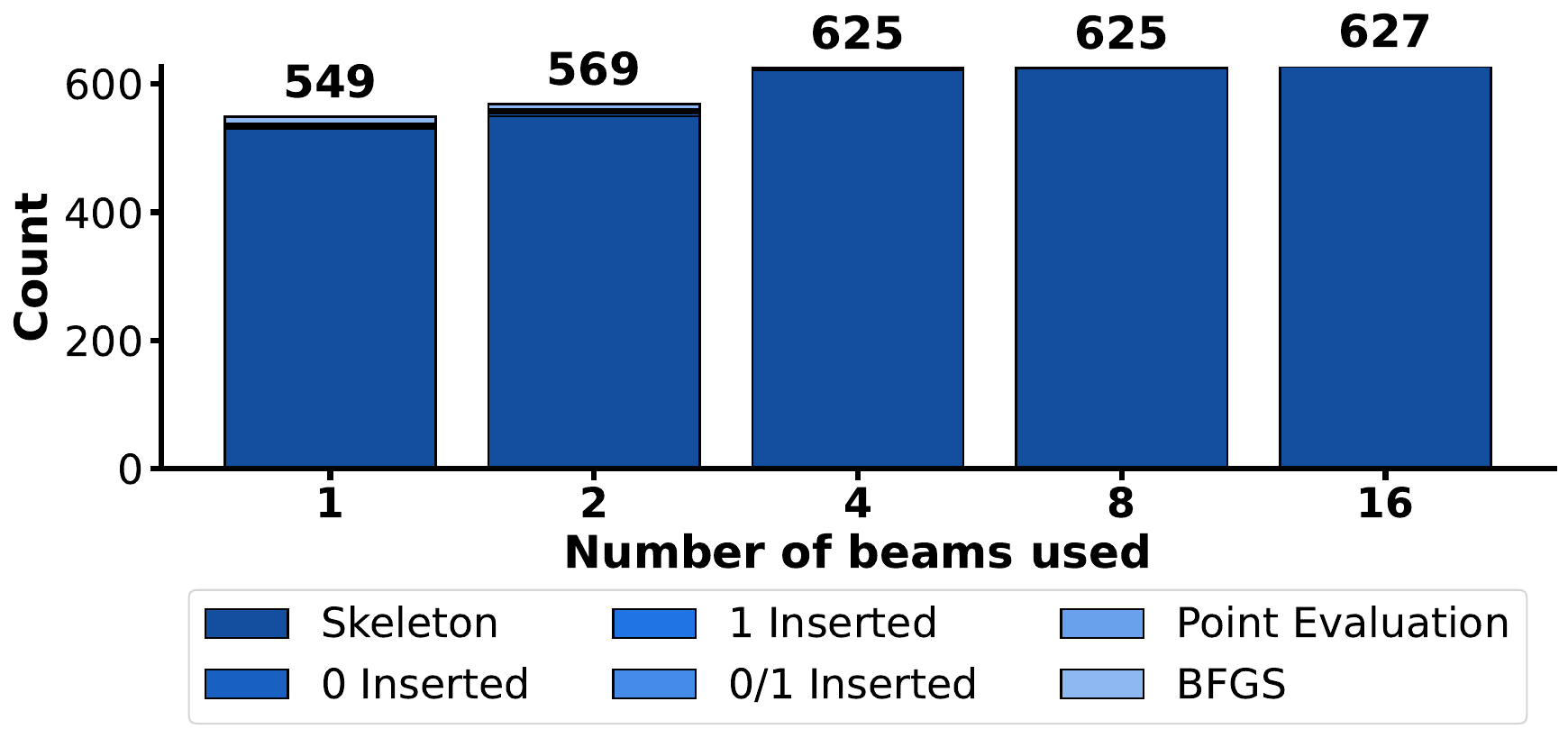}
        \caption{}
        \label{fig:MP across beam sizes 1000 samples}
    \end{subfigure}
    \begin{subfigure}{0.48\textwidth}
        \centering
        \includegraphics[width=\linewidth]{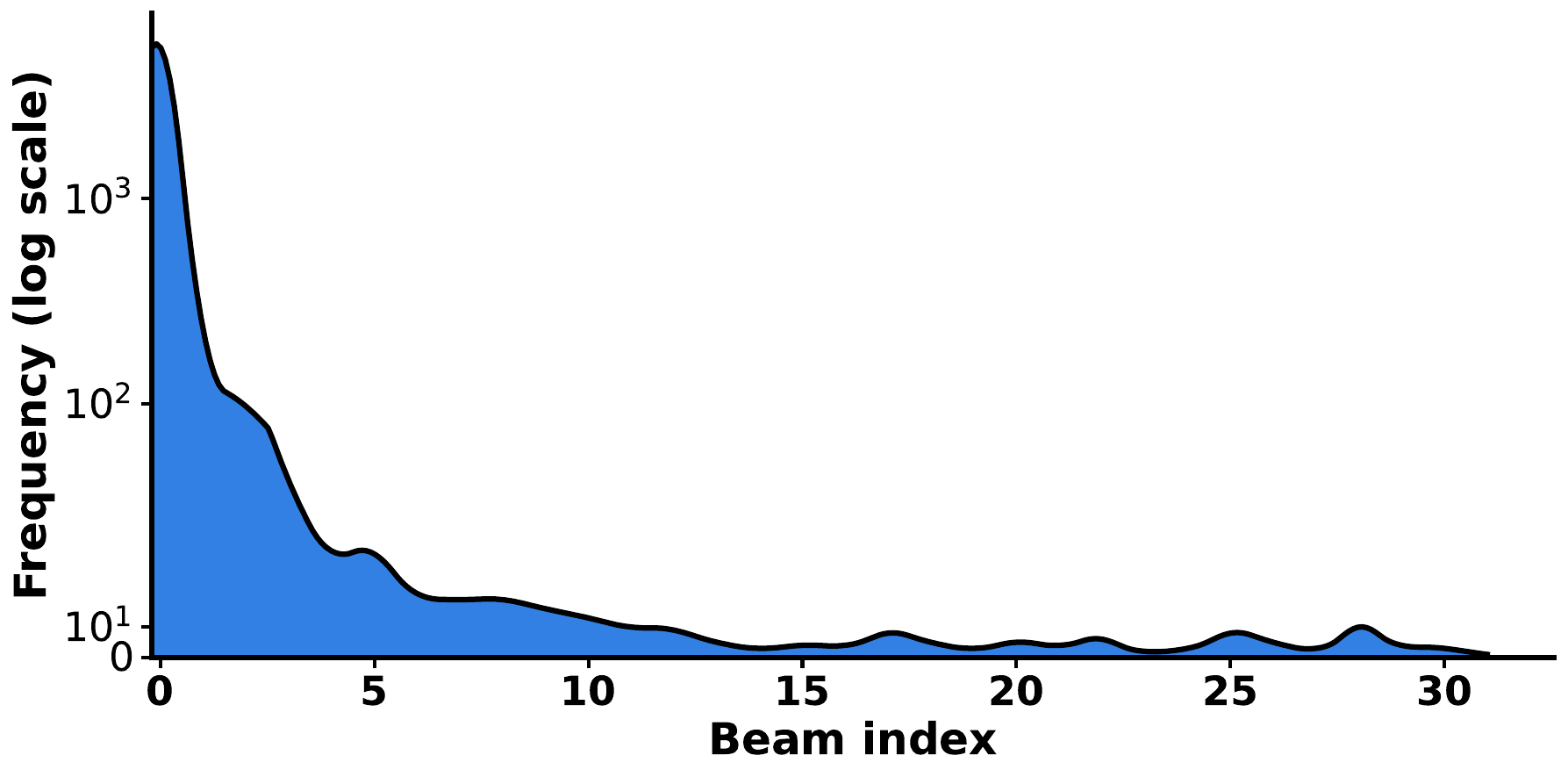}        
        \caption{}
        \label{fig:MP distribution}
    \end{subfigure}
    \caption{Comparison of model performance across beam sizes without constants. Each bar is further subdivided based on reconstruction strategies. \textbf{(a)} Performance on 1\,000 samples without floating-point constants. \textbf{(b)} Distribution of beam positions of correct solutions over 10\,000 samples without constants, using beam size 32, with 6\,125 correct solutions.}
    \label{fig:MP different beam sizes.}
\end{figure}
\fullcref{fig:MP across beam sizes 1000 samples} shows performance improves with larger beam sizes, including better skeleton reconstruction, but the overall benefit of beam search remains small. This means that while larger beams let the model consider more candidate sequences, most improvements come from relatively small increases in beam size.

The distribution plot in \fullcref{fig:MP distribution} shows that with a beam size of 32, most correct solutions appear within the first three beams. This indicates the top-ranked predictions are highly reliable, and exploring beyond these beams seldom improves results. Almost no correct solutions are found beyond beam 15, demonstrating that increasing beam size further offers limited benefit while increasing computational cost.

\subsection{Model Difficulties}
\label{Model Difficulties}

In addition to overall performance, analysing the operation composition of formulas the model can and cannot recreate reveals further strengths and limitations. Figures \ref{fig:MPa} (\(n=6125\)) and \ref{fig:MPb} (\(n=3125\)) compare true equations successfully recreated versus those not recreated (CNR). Notably, the \texttt{log} operation appears only 56 times in the recreated set versus 1,676 times in the CNR set, highlighting a major challenge. The \texttt{exp} operator shows similar occurrence counts in both figures; however, since the CNR set is half the size, its relative frequency is higher in the unable to recreate dataset, posing another limitation.

\begin{figure}[ht!]
    \centering
    \begin{subfigure}{0.48\textwidth}
        \centering
        \includegraphics[width=\linewidth]{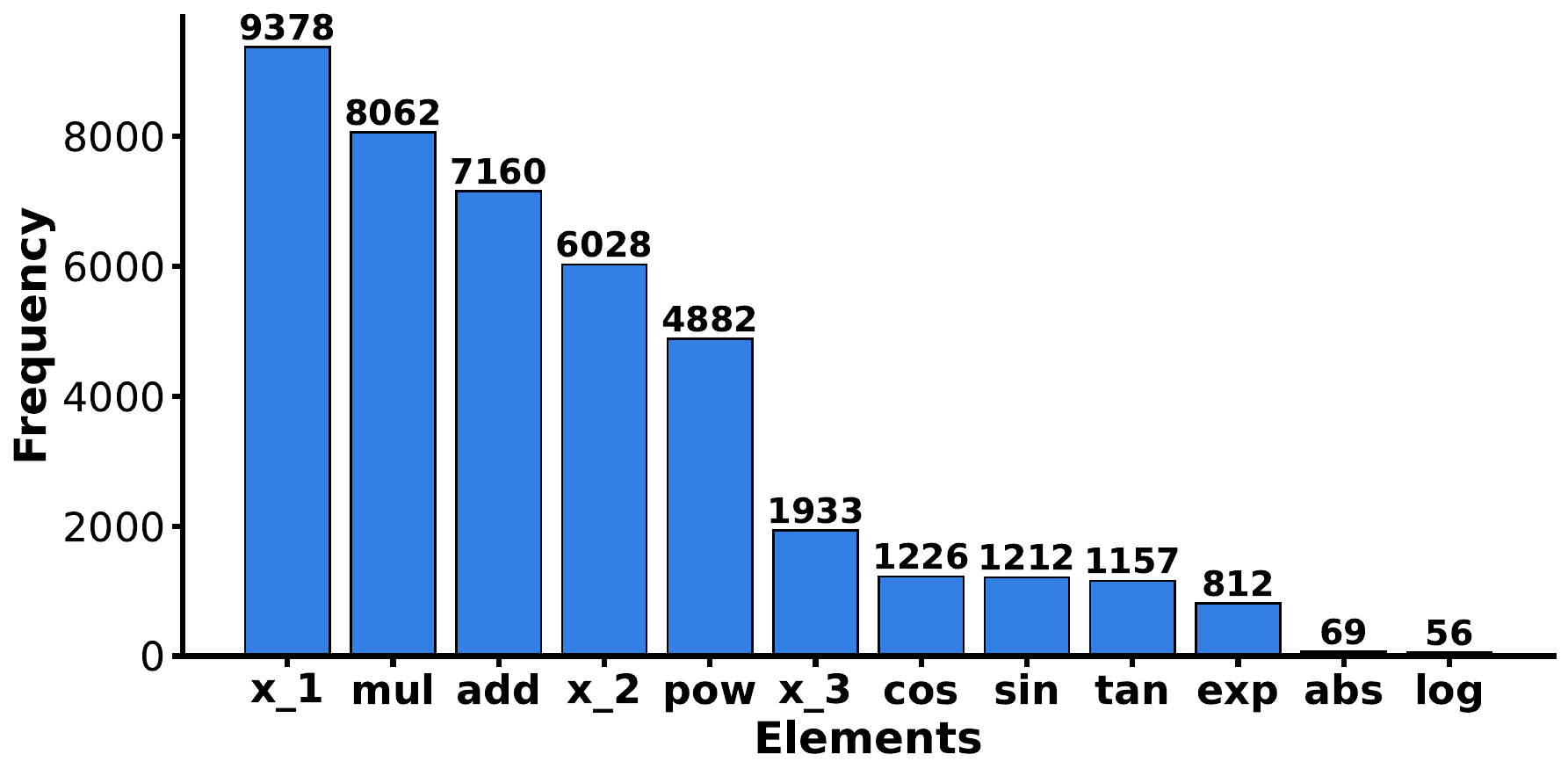}
        \caption{}
        \label{fig:MPa}
    \end{subfigure}
    \begin{subfigure}{0.48\textwidth}
        \centering
        \includegraphics[width=\linewidth]{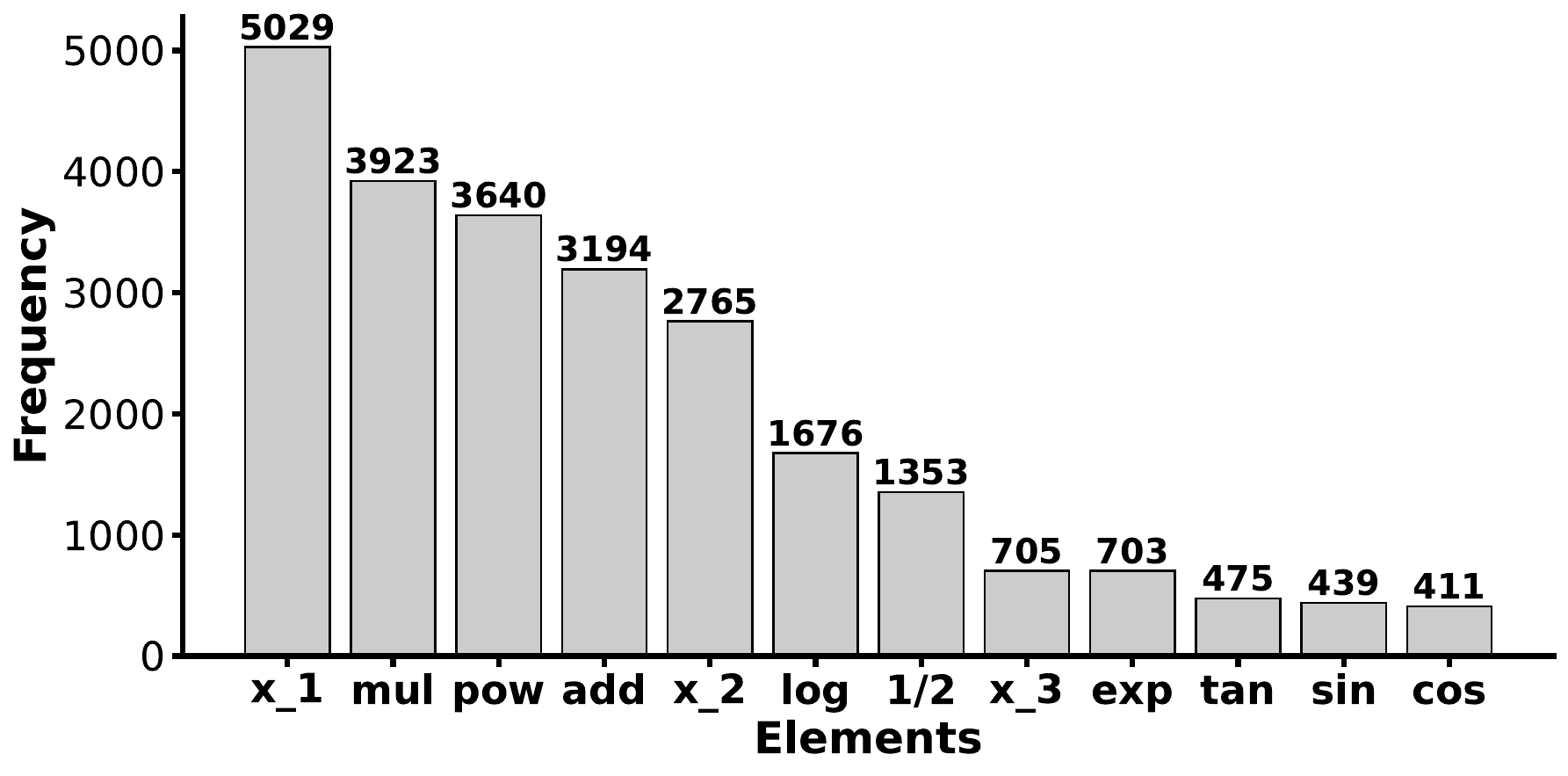}        
        \caption{}
        \label{fig:MPb}
    \end{subfigure}
    \begin{subfigure}{0.48\textwidth}
        \centering
        \includegraphics[width=\linewidth]{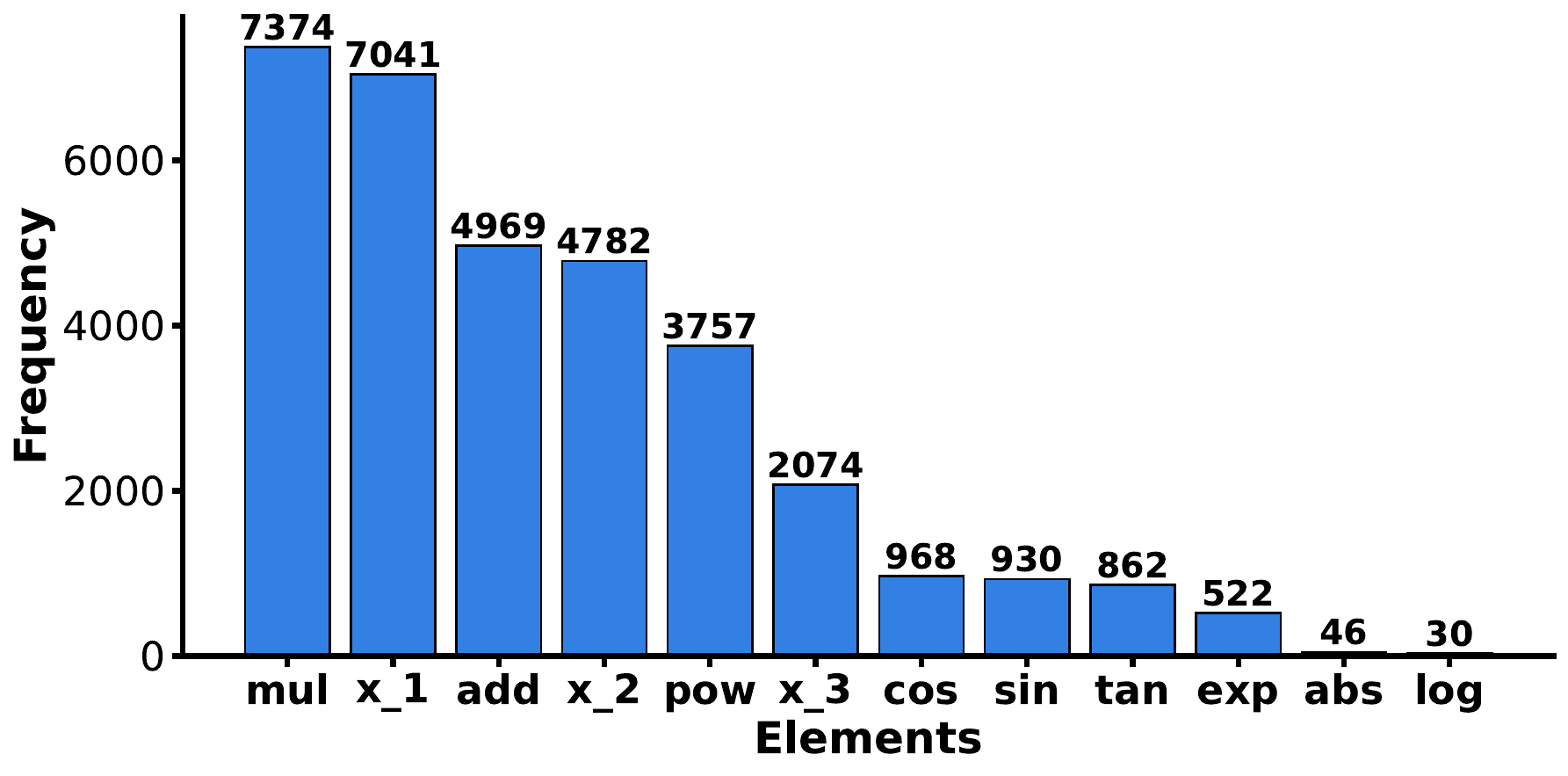}        
        \caption{}
        \label{fig:MPc}
    \end{subfigure}
    \begin{subfigure}{0.48\textwidth}
        \centering
        \includegraphics[width=\linewidth]{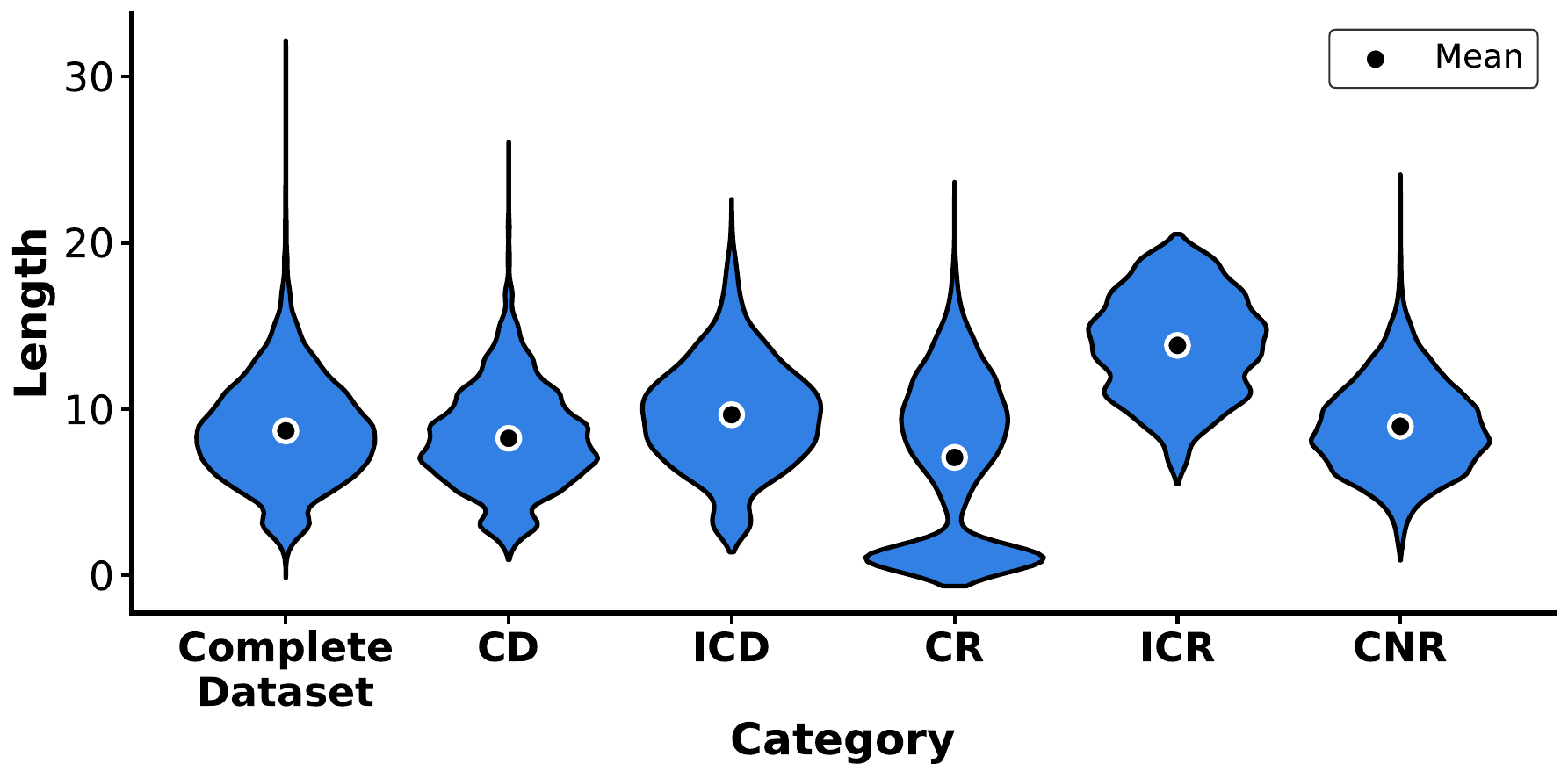}
        \caption{}
        \label{fig:MP violin}
    \end{subfigure}
   \caption{
    Distribution of elements and mean equation lengths for the NeSymReS model tested on 10,000 equations without constants.
    \textbf{(a)} Elements in correctly recreated true equations (\(n=6,125\)); counts above \(6,125\) indicate multiple occurrences per equation.
    \textbf{(b)} Elements in true equations the model failed to recreate (CNR, \(n=3,175\)).
    \textbf{(c)} Elements in the model’s correctly recreated equations.
    \textbf{(d)} Equation length distributions for different subsets: (I)CD = (In)Correct Dataset, (I)CR = (In)Correctly Recreated, CNR = Could Not Recreate.
    }
    \label{fig:MP element distributions}
\end{figure}

Comparing Figures \ref{fig:MPa} and \ref{fig:MPc} reveals similar operation distributions, however, recreated formulas contain relatively more multiplication and fewer \(x_1\) terms, suggesting the model tends to produce more concise expressions, as also observed in \fullcref{fig:MP violin}.

The violin plot in \fullcref{fig:MP violin} illustrates equation length distributions: correctly recreated equations average 5.7 elements, while incorrect ones average 11.2 tokens, indicating significant overgeneration when the model is uncertain. Longer equations also pose more difficulty, reflected in higher mean lengths for the Incorrect and CNR subsets.

\subsection{Feynman AI Dataset}

The Feynman AI Dataset is a collection of mathematical expressions and physical laws used to benchmark SR algorithms \cite{AI_Feynman}. We use this dataset to evaluate real-world performance and compare our results to those reported in the original paper.

\begin{figure}[ht!]
    \centering
    \begin{subfigure}{0.49\textwidth}
        \centering
        \includegraphics[width=\linewidth]{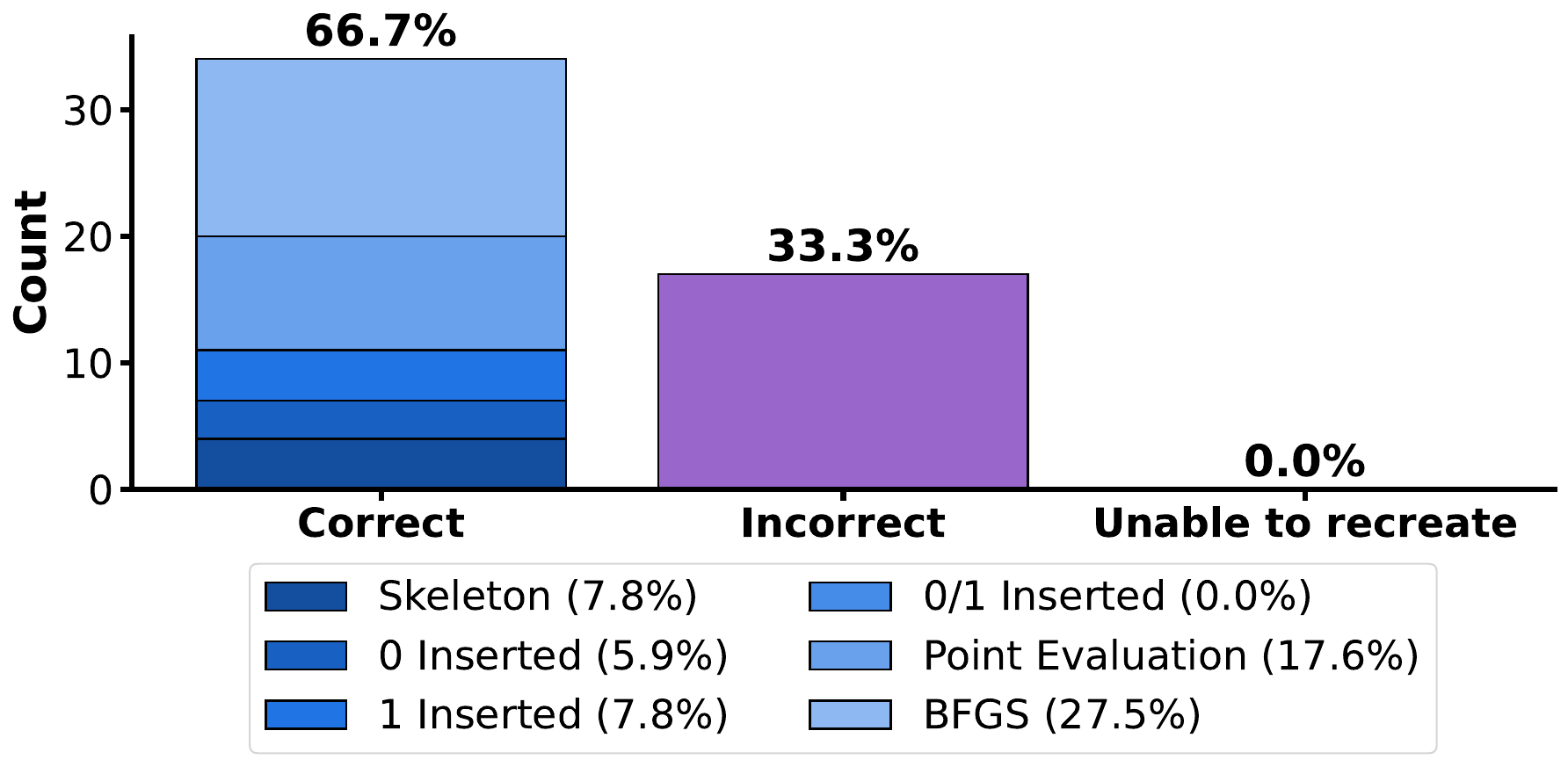}
        \caption{}
        \label{fig:MP Feyman 200}
    \end{subfigure}
    \begin{subfigure}{0.49\textwidth}
        \centering
        \includegraphics[width=\linewidth]{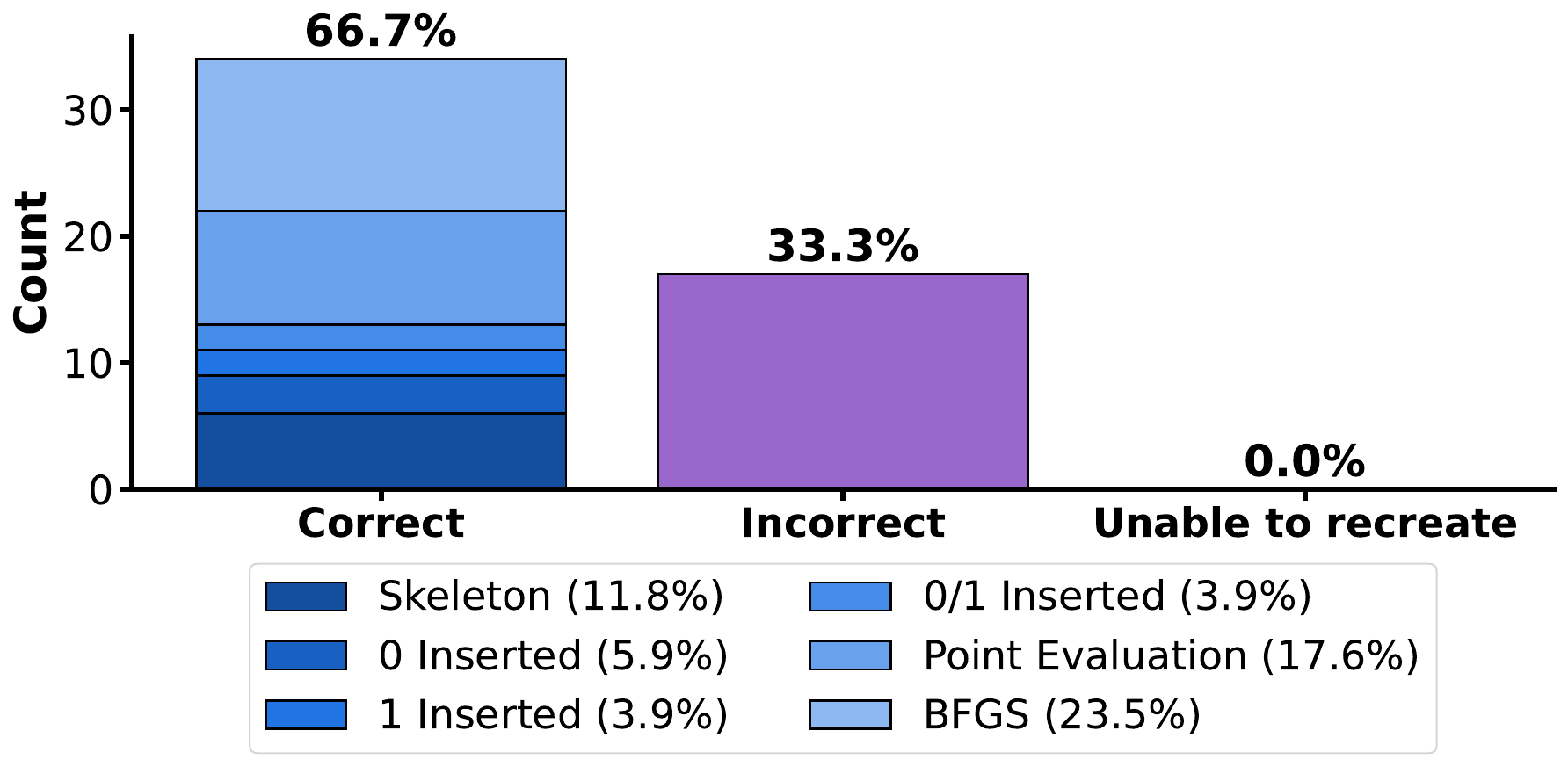}
        \caption{}
        \label{fig:MP Feyman 1000}
    \end{subfigure}
    \caption{Feynman AI Dataset, common benchmarking tool for SR. containing 100 physics based equations from which 51 are able to be used by this model. Others have too many variables. \textbf{(a)} Performance on 200 observations. \textbf{(b)} Performance on 1000 observations.}
    \label{fig:Feyman AI Datset}
\end{figure}
The original study done by NeSymReS reports an accuracy between \(65\%\) and \(75\%\) on this dataset, and \fullcref{fig:Feyman AI Datset} shows results within this margin~\cite{Neural_symbolic_regression_that_scales}. For this final experiment, we tested one remaining hyperparameter: the number of observations required for the encoder to effectively learn the underlying formula. Using \(200\) observations, chosen empirically as a trade-off between performance and runtime, our results are shown in \fullcref{fig:MP Feyman 200}. The model correctly generates the skeleton for \(7.8\%\) of the \(51\) expressions. Replacing the model’s constants with zero or one improves this by \(13.7\%\), and applying the BFGS algorithm solves an additional \(27.5\%\). Increasing the number of observations to \(1,000\) (\fullcref{fig:MP Feyman 1000}) slightly improves skeleton accuracy to \(11.8\%\), but does not enhance overall performance. This suggests that \(200\) observations are sufficient for the encoder to learn the equation's underlying structure.
\newpage
\section{Additional Circuit Discovery Results}
\label{APP: Additional Circuit Discovery Results}
Circuit configurations were selected based on practical constraints: for \texttt{Add} and \texttt{Mul}, mean patching produced trivial circuits of length zero and were excluded; for \texttt{Pow} and \texttt{Exp}, only mean patching was used due to limited compute and the prior discovery of correct circuits. All other operations include both mean and resample patching for comparison.

\patches\ runs for 250 generations with a population size of 40\footnote{This setting was selected as the best trade-off between computational efficiency and circuit quality during preliminary testing.}. To promote a diverse starting population, access to the covariance matrix is disabled for the first 10 generations, and the global step size is initialised to 0.5.

\begin{table}[ht]
\centering
\caption{Model and Patched Model baseline scores, measured in top-$k$ accuracy (T1, T2, T3) and normalised logit score (LS) for R(esample) and M(ean) patching.}
\begin{tabular}{@{}lcccccP{7px}cccc@{}}
\toprule
\textbf{Op} & \textbf{Patching} & 
\multicolumn{4}{c}{\textbf{Model $\uparrow$}} & &
\multicolumn{4}{c}{\textbf{Patched Model $\downarrow$}} \\
\cmidrule(lr){3-6} \cmidrule(lr){8-11}
& & {T1} & {T2} & {T3} & {LS} & & {T1} & {T2} & {T3} & {LS} \\
\midrule
\multirow{1}{*}{\textbf{Add}} & R   & 0.93 & 0.94 & 1.00 & 0.92 && 0.16 &\red{ 0.86} & \red{0.99} & 0.21 \\
\midrule
\multirow{3}{*}{\textbf{Cos}} 
 & M   & 0.95 & 0.97 & 1.00 & 0.92 && 0.00 & 0.00 & 0.00 & 0.00 \\
 & R   & 0.95 & 0.97 & 1.00 & 0.92 && 0.00 & 0.00 & 0.02 & 0.00 \\
 & Sin & 0.95 & 0.97 & 1.00 & 0.92 && 0.10 & \red{0.25} & \red{0.45} & 0.00 \\
\midrule
\multirow{1}{*}{\textbf{Exp}} & M   & 0.60 & 0.75 & 1.00 & 0.58 && 0.00 & 0.00 & 0.00 & 0.00 \\
\midrule
\multirow{3}{*}{\textbf{Log}} 
 & M   & 0.31 & 0.63 & 1.00 & 0.31 && 0.00 & 0.00 & 0.12 & 0.00 \\
 & R   & 0.31 & 0.63 & 1.00 & 0.31 && 0.00 & 0.06 & \red{0.54} & 0.00 \\
 & Exp & 0.31 & 0.63 & 1.00 & 0.31 && 0.00 & 0.20 & \red{0.33} & 0.00 \\
\midrule
\multirow{1}{*}{\textbf{Mul}} & R   & 0.87 & 0.98 & 1.00 & 0.87 && 0.15 & \red{0.95} & \red{1.00} & 0.19 \\
\midrule
\multirow{1}{*}{\textbf{Pow}} & M   & 0.92 & 0.96 & 1.00 & 0.91 && 0.00 & 0.00 & \red{1.00} & 0.00 \\
\midrule
\multirow{3}{*}{\textbf{Sin}} 
 & M   & 0.73 & 0.83 & 1.00 & 0.70 && 0.00 & 0.00 & 0.00 & 0.00 \\
 & R   & 0.73 & 0.83 & 1.00 & 0.70 && 0.03 & 0.12 & 0.24 & 0.04 \\
 & Cos & 0.73 & 0.83 & 1.00 & 0.70 && 0.00 & \red{0.50} & \red{0.70} & 0.03 \\
\midrule
\multirow{3}{*}{\textbf{Tan}} 
 & M   & 0.50 & 0.59 & 1.00 & 0.47 && 0.00 & 0.00 & 0.00 & 0.00 \\
 & R   & 0.50 & 0.59 & 1.00 & 0.47 && 0.00 & 0.03 & 0.08 & 0.01 \\
 & Sin & 0.50 & 0.59 & 1.00 & 0.47 && 0.10 & \red{0.30} & \red{0.42} & 0.00 \\
\bottomrule
\end{tabular}
\label{tab:baselines}
\end{table}

\begin{table}[p]
\vspace*{\fill}
\centering
\caption{Faithfulness and completeness test scores, measured by top-$k$ accuracy (T1–T3) and normalised Logit Score (LS). CFG indicates patching/evaluation strategies: Mean (M), Resample (R), or CoT patching (operation specified in functional evaluation). Evaluation types: Model (M), Functional (F). BL provides baselines for the operator. \textit{Circuit Length} (CL) should be minimised (↓); \textit{faithfulness} maximised (↑); \textit{completeness} minimised (↓). \textit{\#C} indicates the number of circuits found: 1 denotes a unique circuit; values greater than 1 indicate a circuit class. \textit{Correct} denotes whether the circuit reproduces the behaviour under the evaluation strategy. Green: above full model performance, red: below threshold.}
\begin{tabular}{llc@{\hspace{20pt}}ccccP{7px}ccc}
\toprule
\textbf{Op} & \textbf{CFG} & \textbf{CL $\downarrow$} & 
\multicolumn{4}{c}{\textbf{Faithful $\uparrow$}} & &
\multicolumn{2}{c}{\textbf{Complete $\downarrow$}} & \textbf{Correct}\\
\cmidrule(lr){4-7} \cmidrule(l){9-10}
\; \textbf{BL} & & & {T1} & {T2} & {T3} & {LS} & {} & {T3} & {LS} &\\
\midrule
\shortstack[l]{\textbf{Add}\\[-4pt]
\begin{tabular}[t]{@{}c@{}c@{}} 
\scriptsize T1: & \scriptsize \;0.93 \\[-6pt]
\scriptsize T2: & \scriptsize \;0.94 \\[-6pt]
\scriptsize LS: & \scriptsize \;0.92 
\end{tabular}}
& \raisebox{+5pt}{RF}    & \raisebox{+5pt}{57} & \raisebox{+5pt}{0.85} & \raisebox{+5pt}{0.94} & \raisebox{+5pt}{0.99} & \raisebox{+5pt}{\red{0.72}} && \raisebox{+5pt}{\red{0.99}} & \raisebox{+5pt}{0.23} &  \raisebox{+5pt}{\faTimes}\\[-19pt]
& RM & 67 & 0.92 & 0.94 & 1.00 & 0.92 && \red{0.99} & 0.21 &\faTimes \\[+4pt]
\midrule
\multirow{5}{*}{\shortstack[l]{\textbf{Cos}\\[-4pt]
\begin{tabular}[t]{@{}c@{}c@{}} 
\scriptsize T1: & \scriptsize \;0.95 \\[-6pt]
\scriptsize T2: & \scriptsize \;0.97 \\[-6pt]
\scriptsize LS: & \scriptsize \;0.91 
\end{tabular}}}
 &  MF & 61 & 0.87 & 0.92 & 0.99 & \red{0.49} && 0.00 & 0.00  &\faCheck \\
 & MM & 77 & 0.98 & \green{0.99} & 0.99 & 0.86 && 0.00 & 0.00  &\faCheck \\
 & RF & 77 & \red{0.79} & 0.94 & 0.95  & \red{0.51} &&0.00 & 0.00  &\faTimes \\
 & RM & 81 & \green{0.95} & \green{0.97} & 0.97 & 0.81 && 0.00 & 0.00  &\faCheck \\
 & Sin-F & 79 & \red{0.82} & 0.93 & 0.95 & \red{0.62} && \red{0.45} & 0.04  &\faTimes \\
\midrule
\shortstack[l]{\textbf{Exp}\\[-4pt]
\begin{tabular}[t]{@{}c@{}c@{}} 
\scriptsize T1: & \scriptsize \;0.60 \\[-6pt]
\scriptsize T2: & \scriptsize \;0.75 \\[-6pt]
\scriptsize LS: & \scriptsize \;0.58 
\end{tabular}}
& \raisebox{+5pt}{MF}    & \raisebox{+5pt}{58} & \raisebox{+5pt}{0.52} & \raisebox{+5pt}{\green{0.84}} & \raisebox{+5pt}{0.90} & \raisebox{+5pt}{\red{0.38}} && \raisebox{+5pt}{0.00} & \raisebox{+5pt}{0.00} & \raisebox{+5pt}{\faCheck} \\[-19pt]
& MM & 61 & 0.54 & 0.68 & 0.73 & \red{0.45} && 0.00 & 0.00 &\faTimes \\[+4pt]
\midrule

\multirow{5}{*}{\shortstack[l]{\textbf{Log}\\[-4pt]
\begin{tabular}[t]{@{}c@{}c@{}} 
\scriptsize T1: & \scriptsize \;0.30 \\[-6pt]
\scriptsize T2: & \scriptsize \;0.62 \\[-6pt]
\scriptsize LS: & \scriptsize \;0.31 
\end{tabular}}}
 & MF    & 50 & \green{0.66} & \green{1.00} & 1.00 & \green{0.44} && 0.15 & 0.01  & \faCheck \\
 & MM    & 47 & \red{0.00} & \green{0.96} & 1.00 & 0.23 && 0.12 & 0.00  & \faCheck \\
 & RF    & 84 & 0.26 & 0.58 & 0.90 & 0.28 && \red{0.52} & 0.02 & \faTimes \\
 & RM    & 47 & 0.23 & \red{0.41} & \red{0.79} & 0.21 && \red{0.53} & 0.05 & \faTimes \\
 & Exp-F & 70 & 0.24 & 0.60 & 0.92 & 0.30 && \red{0.45} & \red{0.63} & \faTimes \\
\midrule

\shortstack[l]{\textbf{Mul}\\[-4pt]
\begin{tabular}[t]{@{}c@{}c@{}} 
\scriptsize T1: & \scriptsize \;0.86 \\[-6pt]
\scriptsize T2: & \scriptsize \;0.97 \\[-6pt]
\scriptsize LS: & \scriptsize \;0.87 
\end{tabular}}
 & \raisebox{+5pt}{RF}    & \raisebox{+5pt}{61} & \raisebox{+5pt}{\red{0.70}} & \raisebox{+5pt}{0.94} & \raisebox{+5pt}{0.99} & \raisebox{+5pt}{\red{0.59}} && \raisebox{+5pt}{\red{1.00}} & \raisebox{+5pt}{0.19} & \raisebox{+5pt}{\faTimes} \\[-19pt]
 & RM    & 65 & 0.78 & 0.94 & 0.99 & 0.72 && \red{1.00} & 0.18 &\faTimes \\[+4pt]

\midrule

\shortstack[l]{\textbf{Pow}\\[-4pt]
\begin{tabular}[t]{@{}c@{}c@{}} 
\scriptsize T1: & \scriptsize \;0.92 \\[-6pt]
\scriptsize T2: & \scriptsize \;0.95 \\[-6pt]
\scriptsize LS: & \scriptsize \;0.91 
\end{tabular}}
& \raisebox{+5pt}{MF}    & \raisebox{+5pt}{88} & \raisebox{+5pt}{0.86} & \raisebox{+5pt}{0.90} & \raisebox{+5pt}{0.95} & \raisebox{+5pt}{\red{0.58}} && \raisebox{+5pt}{0.00} & \raisebox{+5pt}{0.00} &  \raisebox{+5pt}{\faCheck} \\[-19pt]
 & MM    & 91 & 0.89 & \green{0.96} & 0.97 & 0.82 && 0.10 & 0.00 & \faCheck \\[+4pt]
\midrule
\multirow{5}{*}{\shortstack[l]{\textbf{Sin}\\[-4pt]
\begin{tabular}[t]{@{}c@{}c@{}} 
\scriptsize T1: & \scriptsize \;0.73 \\[-6pt]
\scriptsize T2: & \scriptsize \;0.83 \\[-6pt]
\scriptsize LS: & \scriptsize \;0.70
\end{tabular}}}
 & MF    & 57 & 0.70 & 0.79 & 1.00 & \red{0.51} && 0.00 & 0.00 &  \faCheck \\
 & MM    & 52 & 0.79 & 0.83 & 1.00 & 0.63 && 0.00 & 0.00 & \faCheck \\
 & RF    & 63 & 0.65 & 0.81 & \red{0.87} & \red{0.59} && \red{0.28} & 0.04 &  \faTimes \\
 & RM    & 62 & 0.65 & 0.80 & \red{0.86} & 0.60 && 0.24 & 0.04 & \faCheck \\
 & Cos-F & 69 & \red{0.55} & 0.74 & 0.91 & 0.39 && \red{0.70} & 0.03 &  \faTimes \\
\midrule
\multirow{5}{*}{\shortstack[l]{\textbf{Tan}\\[-4pt]
\begin{tabular}[t]{@{}c@{}c@{}} 
\scriptsize T1: & \scriptsize \;0.50 \\[-6pt]
\scriptsize T2: & \scriptsize \;0.58 \\[-6pt]
\scriptsize LS: & \scriptsize \;0.47 
\end{tabular}}}
 & MF    & 65 & 0.38 & \green{0.60} & 0.92 & \red{0.29} && 0.00 & 0.00 &  \faCheck \\
 & MM    & 66 & 0.41 & \green{0.65} & \red{0.75} & 0.38 && 0.00 & 0.00 & \faCheck \\
 & RF    & 76 & \red{0.36} & \green{0.60} & \red{0.87} & \red{0.30} && 0.08 & 0.01 & \faTimes \\
 & RM    & 63 & \red{0.35} & 0.54 & \red{0.71} & \red{0.33} && 0.08 & 0.01 & \faTimes \\
 & Sin-F & 79 & 0.38 & 0.53 & \red{0.88} & \red{0.30} && 0.23 & 0.02 & \faTimes \\
\bottomrule

\end{tabular}
\vspace*{\fill}
\label{tab:circuits test}
\end{table}

\begin{table}[ht!]
\vspace*{\fill}
\centering
\caption{Faithfulness and completeness train scores, Same structure as \fullcref{tab:circuits test}.}
\begin{tabular}{llc@{\hspace{20pt}}ccccP{7px}cccc}
\toprule
\textbf{Op} & \textbf{CFG} & \textbf{CL $\downarrow$} & 
\multicolumn{4}{c}{\textbf{Faithful $\uparrow$}} & &
\multicolumn{2}{c}{\textbf{Complete $\downarrow$}} & \textbf{Correct}\\
\cmidrule(lr){4-7} \cmidrule(l){9-10}
\; \textbf{BL} & & & {T1} & {T2} & {T3} & {LS} & {} & {T3} & {LS}  &\\
\midrule
\shortstack[l]{\textbf{Add}\\[-4pt]
\begin{tabular}[t]{@{}c@{}c@{}} 
\scriptsize T1: & \scriptsize \;0.91 \\[-6pt]
\scriptsize T2: & \scriptsize \;0.93 \\[-6pt]
\scriptsize LS: & \scriptsize \;0.91 
\end{tabular}}
& RF & 57 & 0.84 & 0.90 & 0.99 & 0.70 && 0.95 & 0.19   & \faTimes \\[-15pt]
& RM & 67 & 0.88 & 0.94 & 0.99 & 0.87 && 0.95 & 0.17   & \faTimes \\ [+4pt]
\midrule
\multirow{5}{*}{\shortstack[l]{\textbf{Cos}\\[-4pt]
\begin{tabular}[t]{@{}c@{}c@{}} 
\scriptsize T1: & \scriptsize \;0.92 \\[-6pt]
\scriptsize T2: & \scriptsize \;0.93 \\[-6pt]
\scriptsize LS: & \scriptsize \;0.92 
\end{tabular}}}
 &  MF & 61 & 0.87 & 0.92 & 0.99 & 0.48 && 0.00 & 0.00   & \faCheck  \\
& MM & 77 & 0.97 & 0.97 & 0.98 & 0.82 && 0.00 & 0.00   & \faCheck \\
& RF & 77 & 0.84 & 0.94 & 0.95 & 0.53 && 0.01 & 0.00   & \faCheck \\
& RM & 81 & 0.94 & 0.96 & 0.96 & 0.82 && 0.01 & 0.00   & \faCheck \\
 & Sin-F & 79 & 0.85 & 0.94 & 0.96 & 0.61 && 0.53 & 0.04 &\faTimes \\
\midrule
\shortstack[l]{\textbf{Exp}\\[-4pt]
\begin{tabular}[t]{@{}c@{}c@{}} 
\scriptsize T1: & \scriptsize \;0.59 \\[-6pt]
\scriptsize T2: & \scriptsize \;0.74 \\[-6pt]
\scriptsize LS: & \scriptsize \;0.59 
\end{tabular}}
& MF & 58 & 0.53 & 0.83 & 0.90 & 0.38 && 0.00 & 0.00 & \faCheck  \\[-15pt]
& MM & 61 & 0.61 & 0.73 & 0.75 & 0.49 && 0.00 & 0.00 & \faCheck \\ [+4pt]
\midrule

\multirow{5}{*}{\shortstack[l]{\textbf{Log}\\[-4pt]
\begin{tabular}[t]{@{}c@{}c@{}} 
\scriptsize T1: & \scriptsize \;0.30 \\[-6pt]
\scriptsize T2: & \scriptsize \;0.62 \\[-6pt]
\scriptsize LS: & \scriptsize \;0.31 
\end{tabular}}}
& MF & 50 & 0.61 & 1.00 & 1.00 & 0.44 && 0.15 & 0.00 & \faCheck \\
& MM & 47 & 0.00 & 0.96 & 1.00 & 0.23 && 0.05 & 0.00  & \faCheck \\
& RF & 84 & 0.29 & 0.55 & 0.90 & 0.30 && 0.46 & 0.00 & \faCheck \\
& RM & 47 & 0.24 & 0.35 & 0.74 & 0.21 && 0.44 & 0.00& \faCheck \\
&Sin-F  & 70 & 0.32 & 0.55 & 0.90  & 0.23  && 0.42 & 0.00  & \faTimes \\
\midrule
\shortstack[l]{\textbf{Mul}\\[-4pt]
\begin{tabular}[t]{@{}c@{}c@{}} 
\scriptsize T1: & \scriptsize \;0.86 \\[-6pt]
\scriptsize T2: & \scriptsize \;0.97 \\[-6pt]
\scriptsize LS: & \scriptsize \;0.87 
\end{tabular}}
& RF & 61 & 0.76 & 0.95 & 1.00 & 0.60 && 1.00 & 0.24 & \faTimes  \\[-15pt]
& RM & 65 & 0.83 & 0.96 & 1.00 & 0.77 && 1.00 & 0.24 & \faTimes \\[+4pt]
\midrule
\shortstack[l]{\textbf{Pow}\\[-4pt]
\begin{tabular}[t]{@{}c@{}c@{}} 
\scriptsize T1: & \scriptsize \;0.92 \\[-6pt]
\scriptsize T2: & \scriptsize \;0.95 \\[-6pt]
\scriptsize LS: & \scriptsize \;0.91 
\end{tabular}}
& MF & 88 & 0.84 & 0.92 & 0.95 & 0.57 && 0.00 & 0.00 & \faTimes \\[-15pt]
& MM & 91 & 0.89 & 0.98 & 0.98 & 0.81 && 0.10 & 0.00 &\faTimes  \\[+4pt]
\midrule
\multirow{5}{*}{\shortstack[l]{\textbf{Sin}\\[-4pt]
\begin{tabular}[t]{@{}c@{}c@{}} 
\scriptsize T1: & \scriptsize \;0.65 \\[-6pt]
\scriptsize T2: & \scriptsize \;0.78 \\[-6pt]
\scriptsize LS: & \scriptsize \;0.62 
\end{tabular}}}
& MF & 57 & 0.63 & 0.78 & 1.00 & 0.48 && 0.00 & 0.00 & \faCheck \\
& MM & 52 & 0.76 & 0.80 & 1.00 & 0.60 && 0.00 & 0.00 & \faCheck \\
& RF & 63 & 0.63 & 0.80 & 0.90 & 0.57 && 0.21 & 0.03 &\faCheck \\
& RM & 62 & 0.66 & 0.78 & 0.82 & 0.59 && 0.17 & 0.03 &\faCheck \\
&Cos-F & 69 & 0.59 & 0.67 & 0.90 & 0.50 && 0.65 & 0.07& \faTimes \\
 \midrule
\multirow{5}{*}{\shortstack[l]{\textbf{Tan}\\[-4pt]
\begin{tabular}[t]{@{}c@{}c@{}} 
\scriptsize T1: & \scriptsize \;0.50 \\[-6pt]
\scriptsize T2: & \scriptsize \;0.58 \\[-6pt]
\scriptsize LS: & \scriptsize \;0.47 
\end{tabular}}}
& MF & 65 & 0.40 & 0.59 & 0.92 & 0.29 && 0.00 & 0.00  & \faCheck \\
& MM & 66 & 0.43 & 0.64 & 0.75 & 0.39 && 0.00 & 0.00  & \faCheck \\
& RF & 76 & 0.40 & 0.61 & 0.90 & 0.32 && 0.12 & 0.02  & \faCheck \\
& RM & 63 & 0.41 & 0.55 & 0.66 & 0.37 && 0.12 & 0.02 & \faCheck \\
&Sin-F & 79 & 0.45 & 0.50 & 0.91 & 0.21 && 0.25 & 0.00  & \faTimes \\
\bottomrule
\end{tabular}
\label{tab:circuits Train}
\end{table}

\begin{figure}
    \centering
    \includegraphics[width=0.8\linewidth]{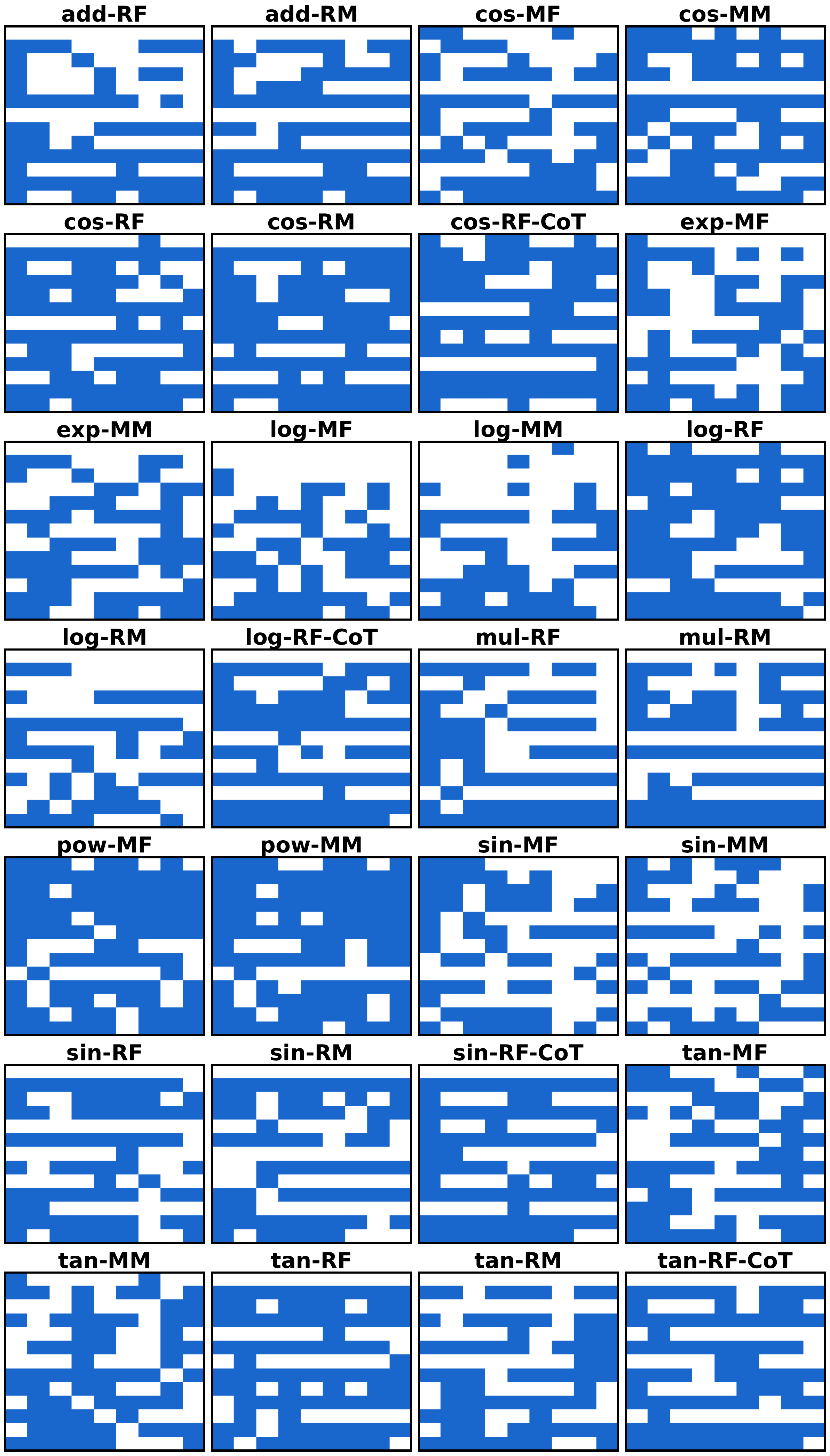}
    \caption{Circuits across all methods and operations. While usage patterns vary across components, no clear or consistent correlations emerge between component usage and specific operations or discovery methods. This suggests that circuit composition is not tightly tied to the type of symbolic operation or the patching strategy used. Vertical: layers, Horizontal MLP + heads 1-8.}
    \label{fig:all_circuits}
\end{figure}

\newpage
\section{Multi-Token Circuits}
\label{Multi-Token Circuits}
\fullcref{tab:mulitokencircuits} displays baseline, train, and test performance on two multitoken circuits namely monomial and posynomial function classes. We chose to focus on the mean functional configuration due to limited compute and mean patching requiring fewer resources, with functional evaluation typically leading to better end-user performance.

\subsection{Monomial and Posynomial Function Classes}

In SR, some methods constrain the search space to specific function classes to improve tractability or reflect domain knowledge. Two such classes are \textit{monomial} and \textit{posynomial} functions, which are particularly relevant in domains such as engineering, physics, and control systems, where functional relationships often follow power laws or multiplicative structures \cite{boyd2007tutorial, AI_Feynman}. therefore we choose to find circuits for these function classes.

A \textbf{monomial function} is defined as:
\[
f(x) = c \cdot x_1^{a_1} x_2^{a_2} \cdots x_n^{a_n},
\]
where \( c > 0 \) and \( a_i \in \mathbb{R} \). These functions are positive for all \( x_i > 0 \).

For instance, in physics education, the period \( T \) of a pendulum is commonly expressed as a monomial: \( T = 2\pi \sqrt{\frac{L}{g}} \), where \( L \) is the length of the pendulum and \( g \) is the gravitational constant \cite{openstax2023pendulum}. This expression fits the monomial form, as it consists of a constant ($2\pi$) multiplied by variables ($L/g$) raised to a power ($\frac{1}{2}$). If prior knowledge suggests such a relationship, restricting the model to monomials can lead to faster and more accurate discovery.

A \textbf{posynomial function} generalizes this form:
\[
f(x) = \sum_{k=1}^{K} c_k \cdot x_1^{a_{1k}} x_2^{a_{2k}} \cdots x_n^{a_{nk}},
\]
where \( c_k > 0 \) and \( a_{ik} \in \mathbb{R} \). Posynomials are essentially weighted sums of monomials and preserve the interpretability of individual terms while capturing more complex relationships.

Posynomial functions, in contrast, are useful when modelling more complex systems that involve sums of monomials. For instance, in designing battery-powered drones, engineers may want to model flight time as a function of several factors like battery capacity, weight, and motor efficiency. Each of these factors contribute multiplicatively to the total, and their combined effect may be expressed as a sum of monomials \cite{tyto2023flighttime}.

\subsection{Results}

\begin{table}[h]
\centering
\caption{Evaluation scores for Monomial and Posynomial circuits on MF configuration}
\begin{tabular}{@{}llcccccP{7px}ccc@{}}
\toprule
\textbf{Func Class} & \textbf{Eval} & \textbf{CL $\downarrow$} & 
\multicolumn{4}{c}{\textbf{Faithful $\uparrow$}} & &
\multicolumn{2}{c}{\textbf{Complete $\downarrow$}} &  \textbf{Correct}\\
\cmidrule(lr){4-7} \cmidrule(l){9-10}
 & & & {T1} & {T2} & {T3} & {LS} & {} & {T3} & {LS} & \\
\midrule
\multirow{3}{*}{\textbf{Monomial}} 
  & Baseline & \multirow{3}{*}{69} & 0.79 & 0.88 & 1.00 & 0.87 && 0.15 & 0.07 & \multirow{3}{*}{\faCheck}\\
  & Train    &                      & 0.69 & 0.85 & 0.95 & \textcolor{red}{0.56} && 0.00 & 0.00 & \\
  & Test     &                      & \textcolor{green!60!black}{0.82} & \textcolor{green!60!black}{0.89} & 1.00 & \red{0.60} && 0.09 & 0.05 & \\
\midrule
\multirow{3}{*}{\textbf{Posynomial}} 
  & Baseline & \multirow{3}{*}{55} & 0.66 & 1.00 & 1.00 & 0.65 && 0.05 & 0.00 & \multirow{3}{*}{\faTimes}\\
  & Train    &                      & 0.56 & 0.96 & 1.00 & \textcolor{red}{0.47} && 0.00 & 0.00 & \\
  & Test     &                      & \textcolor{red}{0.55} & 0.93 & 1.00 & 0.60 && 0.09 & 0.05 & \\
\bottomrule
\end{tabular}
\label{tab:mulitokencircuits}
\end{table}

The results displayed in \fullcref{tab:mulitokencircuits} illustrate a correct circuit for monomial, even outperforming train and baseline performance on the test set for Top-1 and Top-2 accuracy. The circuit is not unique where we can replace 2 heads with 3 other heads thus having a circuit class of 5 (one combination gives an incorrect circuit) circuits. The posynomial circuit is incorrect as it has a 1\% too low top-1 accuracy but is complete and unique. We are thus able to find circuits of not only a single component, but extent it to multi-token circuits.

\section{Additional Verification Experiments}
\label{Additional Verification Experiments}
\subsection{Component Usage}

As an additional verification experiment that circuits are not merely subsets of one another but rely on distinct components, \fullcref{fig:component usage} shows how many circuits use each model component, including MLPs and attention heads across all layers. One component appears in every circuit: the output (OUT) MLP, highlighted with yellow borders. This is expected, as the OUT MLP maps the final internal representation to the model’s output token distribution and is thus always involved in generating the predicted behaviour.

\begin{figure}[ht!]
    \centering
    \includegraphics[width=0.5\linewidth]{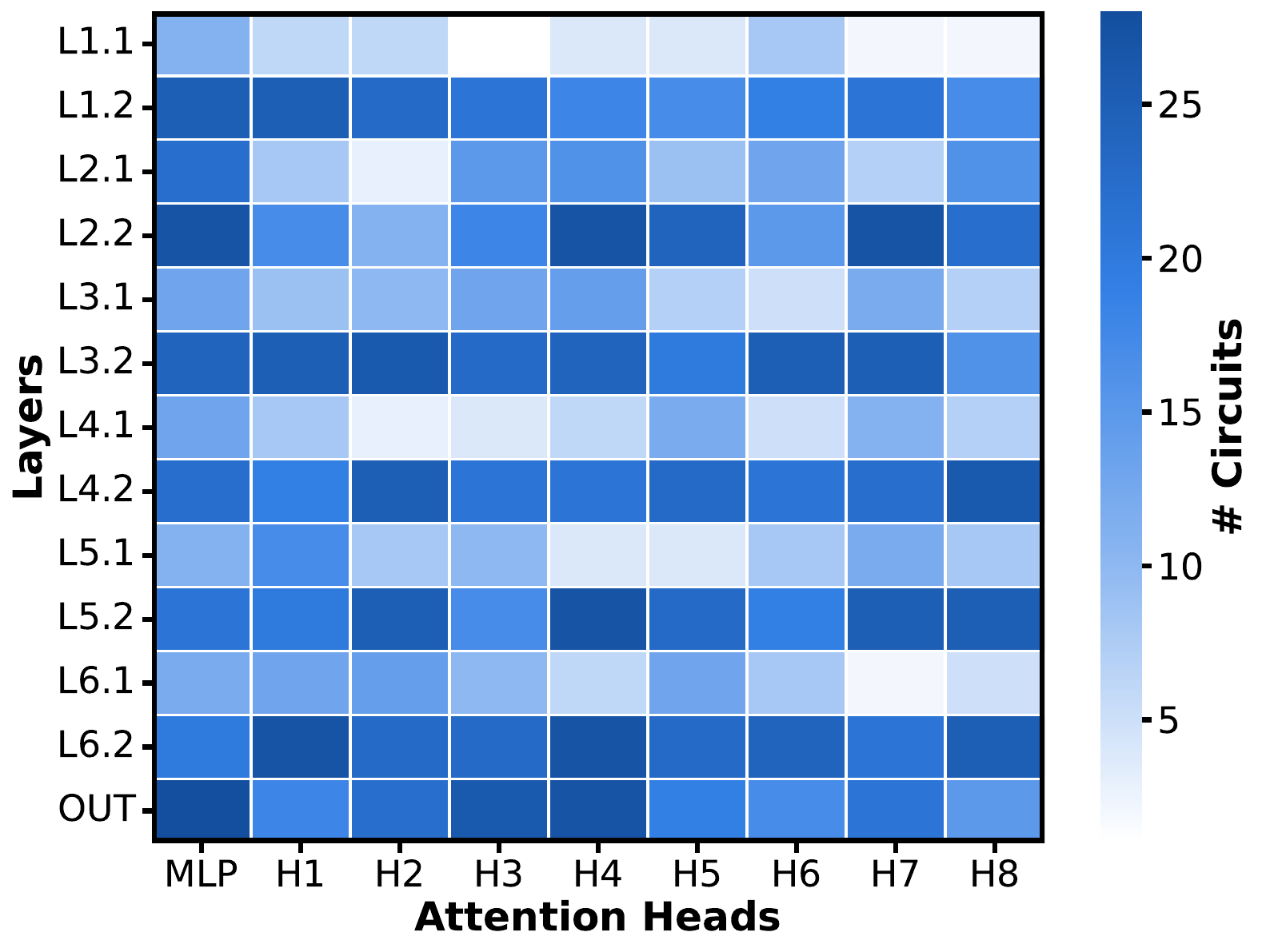}
    \caption{Usage of components in all circuits. Darker blue means more circuits use this component. Only the OUT MLP is used by all found circuits meaning all components have their own }
    \label{fig:component usage}
\end{figure}

A notable trend is the higher utilisation of the second Multi-Head Attention Block (MAB2) compared to the first (MAB1) across layers. This can be attributed to the presence of a residual connection over the first MAB, which allows the model to bypass its outputs entirely, whereas no such residual exists over the second within a layer, only over the entire ISAB (see \fullcref{fig:set_encoder_diagram}).

Interestingly, there is no clear preference for either early or late layers; component usage appears distributed relatively evenly across the encoder. This suggests that the model leverages information from various levels of abstraction throughout. In addition, there is no apparent correlation between function class and component usage; circuits for \texttt{sin}, \texttt{cos}, and \texttt{tan} do not more closely resemble each other than they do circuits for \texttt{add}, \texttt{mul}, or \texttt{pow} (see \fullcref{fig:all_circuits}).
The output projection layer (OUT) is frequently used, which is expected as it maps internal representations back into the vocabulary embedding space for token generation.

\subsection{Overlap Percentages}
\begin{figure}[ht!]
    \centering
    \begin{subfigure}{0.48\textwidth}
        \centering
        \includegraphics[width=\linewidth]{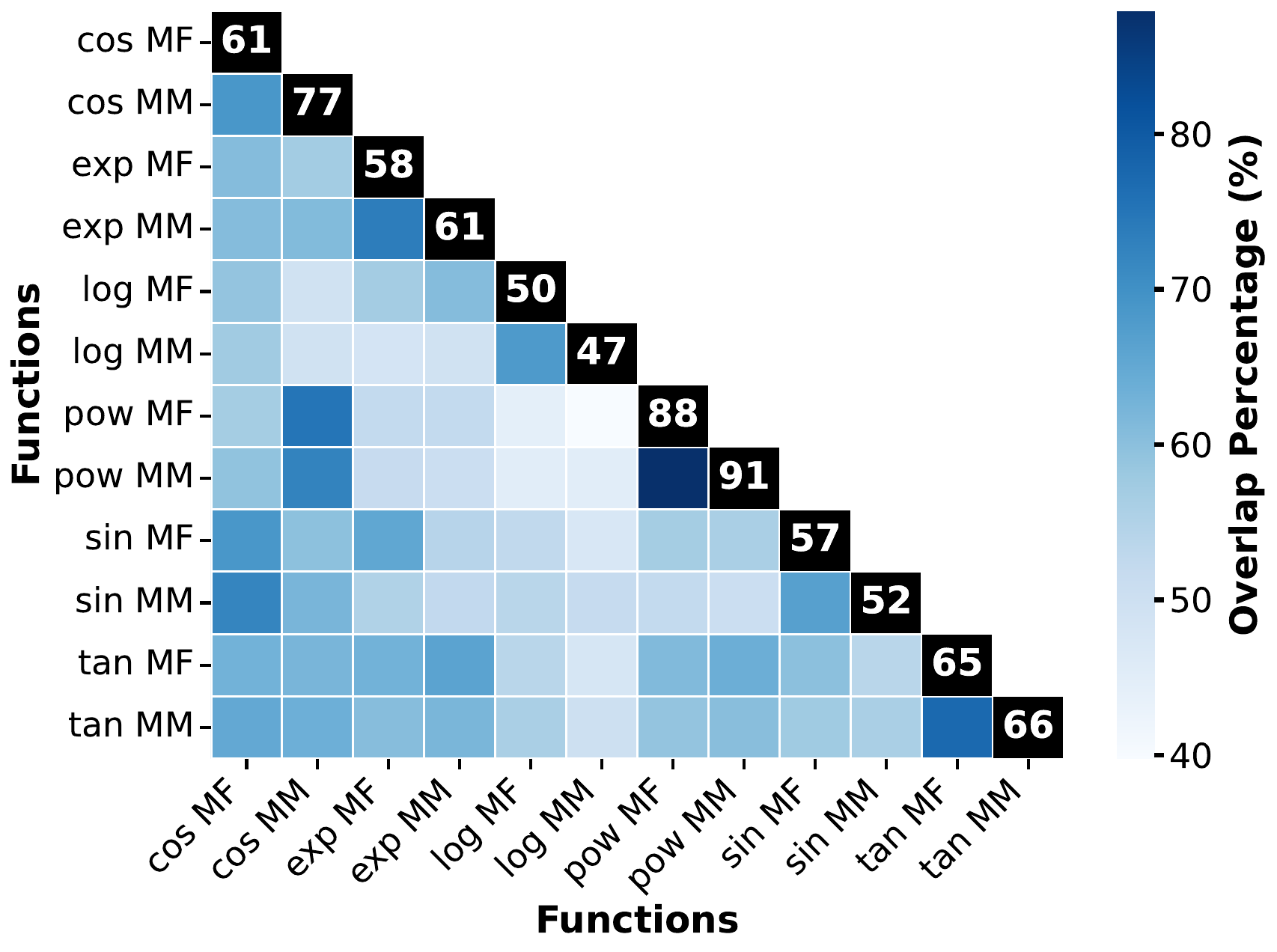}
        \caption{}
        \label{fig:Overlap MvsFF}
    \end{subfigure}
    \begin{subfigure}{0.48\textwidth}
        \centering
        \includegraphics[width=\linewidth]{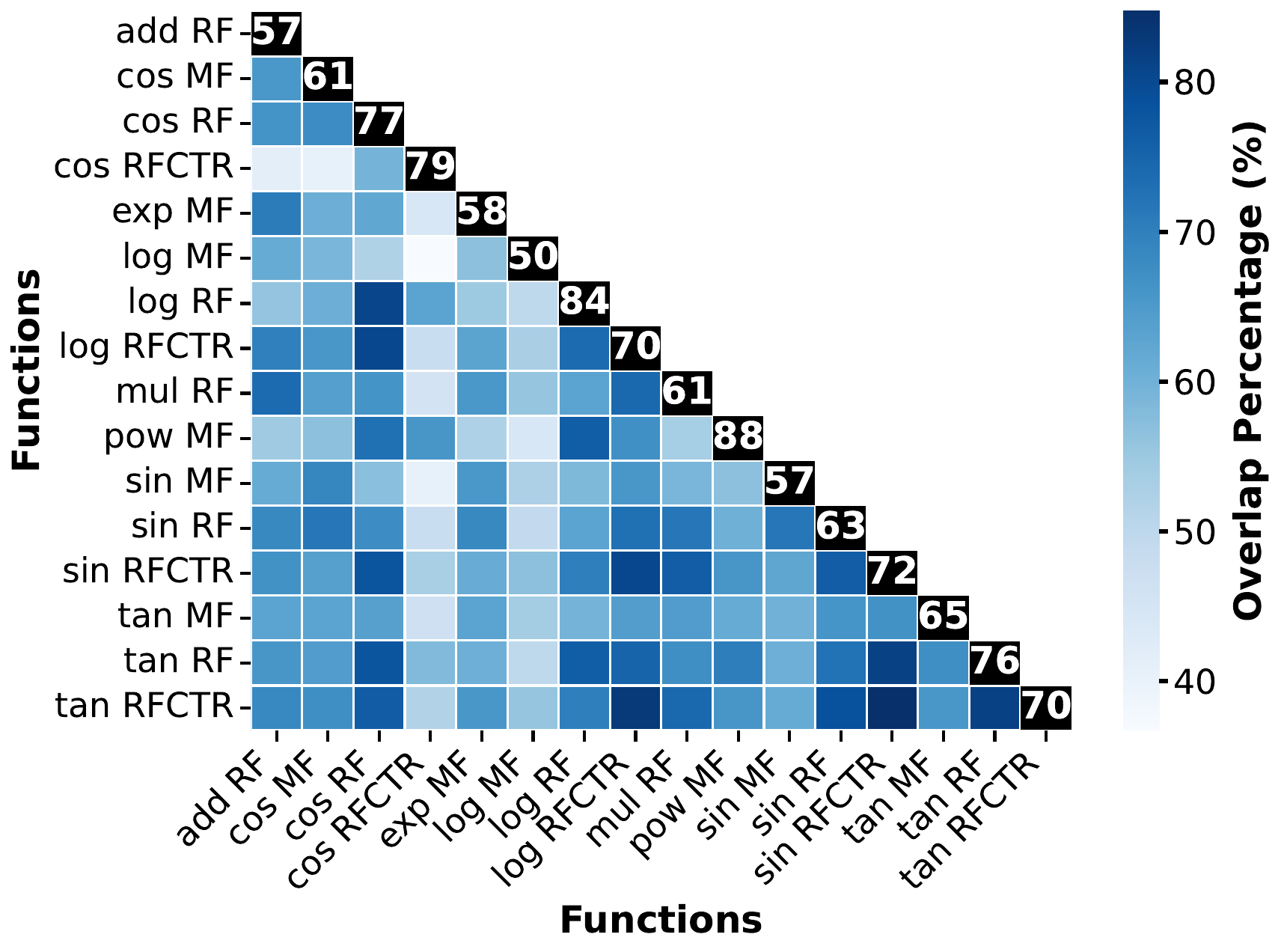}
        \caption{}
        \label{fig:Overlap MvsRP}
    \end{subfigure}
    \caption{Overlap matrices comparing circuit similarity across methods. Each cell reports the maximum overlap percentage between two circuits; diagonal entries indicate circuit lengths (number of components). \textbf{(a)} Overlap between model-faithful (MM) and functionally-faithful (MF) circuits using mean patching. \textbf{(b)} Overlap between mean-patched (MF) and resample-patched (RF) circuits, both evaluated for functional faithfulness.}
    \label{fig:Overlap}
\end{figure}

\begin{figure}
    \centering
    \includegraphics[width=0.99\linewidth]{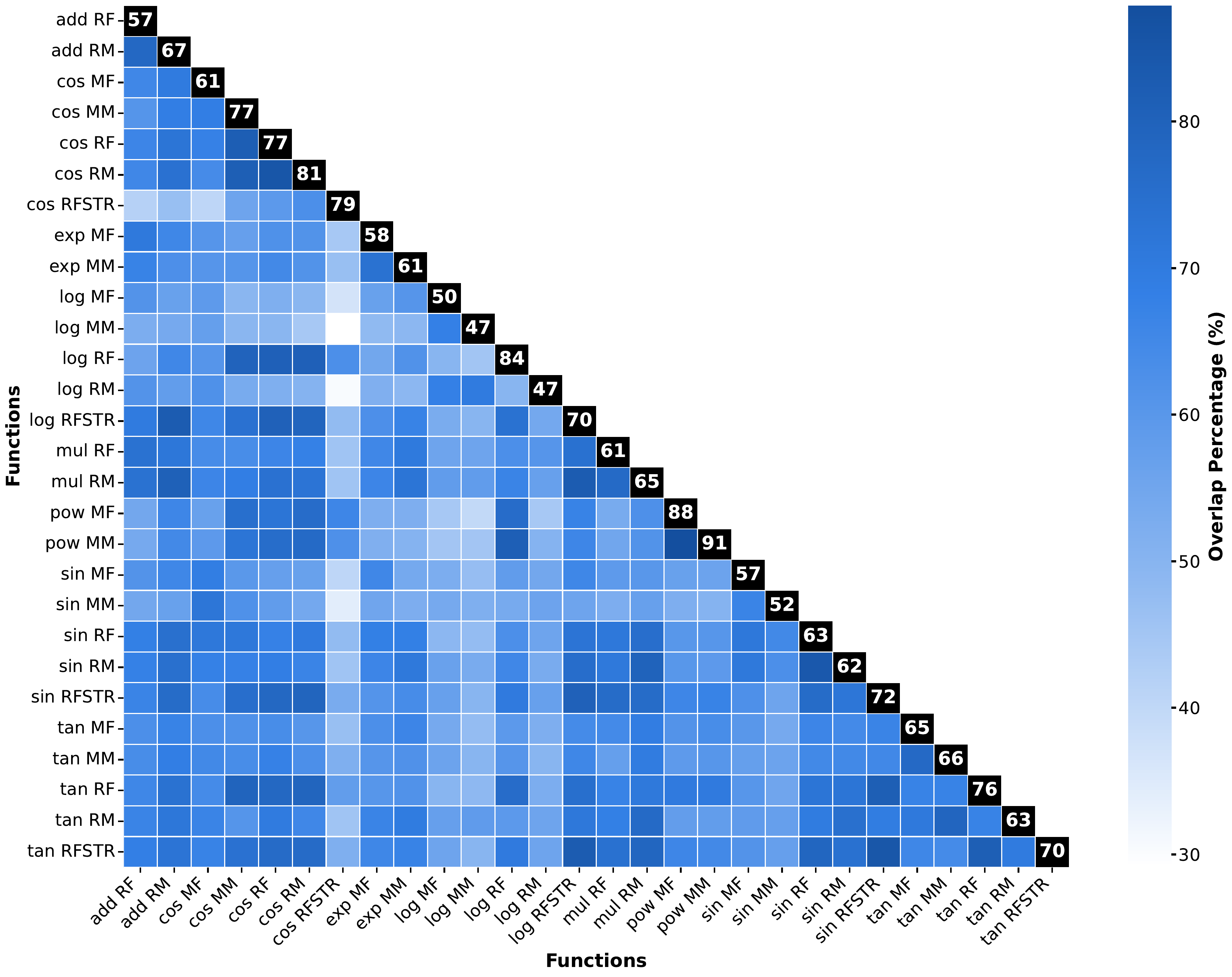}
    \caption{Overlap matrix comparing circuit similarity across all methods. Each cell reports the maximum overlap percentage between two circuits; diagonal entries indicate circuit lengths (number of components)}
    \label{fig:overlapAll}
\end{figure}

To confirm that real circuits are discovered, not just active components, \fullcref{fig:Overlap MvsFF} displays the component overlap between model- and functionally evaluated circuits. Circuits for the same operation share most components, confirming that different operations yield distinct circuits. Log circuits show little overlap with others, likely due to weak model performance and the distinct structure of log operations. There is no consistent similarity pattern across strategies; model- and functionally faithful circuits do not cluster more closely with their own type

\fullcref{fig:Overlap MvsRP} displays the overlap in circuits between resample and mean patching strategies, showing overlap within circuits of the same operation are highest. Additionally, circuits derived from resample patching exhibit greater overlap than those from mean patching, likely due to their increased size. A complete overlap matrix can be found in \fullcref{fig:overlapAll}.

\subsection{Recovery Scores}
To further assess the circuits, we test whether they can outperform the full model in reproducing specific behaviours. To assess this, we used an experimental setting where the baseline model fails to accurately recreate the TgT. The complement of each circuit was patched out, leaving only the circuit intact, and top-3 accuracy was measured on a 100-example test set. Results are shown in \fullcref{fig:improvement scores}.

\begin{figure}[ht!]
    \centering
    \includegraphics[width=0.8\linewidth]{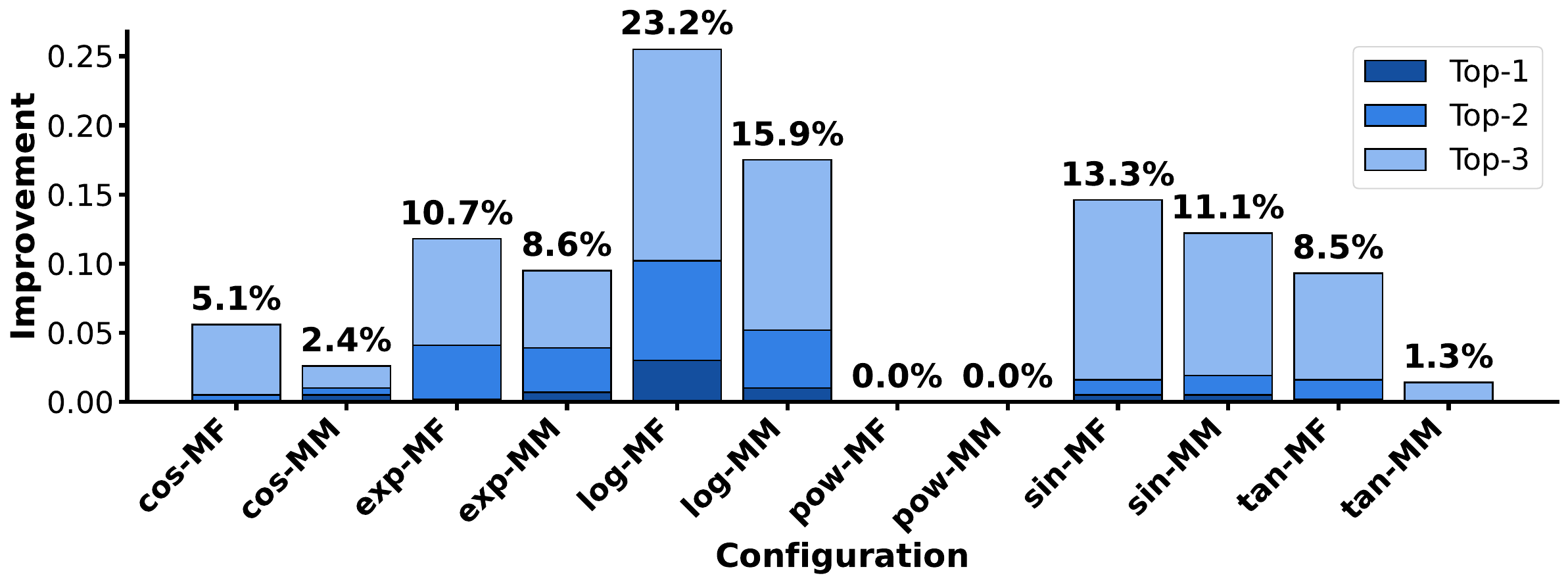}
    \caption{Improvement score of each circuit over the full model. For 100 samples where the full model failed to predict the correct behaviour in the top-3, the circuit model was tested.}
    \label{fig:improvement scores}
\end{figure}

Improvement scores vary from zero to 23.2\%. We show moderate improvements for \texttt{cos} and \texttt{tan} operators and no improvement for \texttt{pow}. As observed in \fullcref{Model Difficulties} these operators already perform well with pow performing best. For \texttt{log}, \texttt{exp}, and \texttt{sin} we display higher performance increases also corresponding to weaker performing operators as seen in \fullcref{fig:MP element distributions}. Addition and multiplication operators are not displayed because they observe the same trends as \texttt{pow}.

These improvements provide further evidence that the circuits accurately capture the intended behaviour. The fact that performance improves when the rest of the model is ablated supports the hypothesis that the retained subcomponents are not only sufficient but also more specialised and less confused by unrelated computation. This behaviour resembles that of a mixture of experts model (MoE), in which different submodules specialise in different tasks. However, unlike traditional MoE systems where routing is learned explicitly, here the specialisation emerges implicitly, and is revealed through circuit extraction. This suggests that even within densely connected models, modular and interpretable substructures are recoverable.

\newpage
\section{Iterative Patching: Additional Results}
\label{Iterative Patching: Additional Results}

\begin{table}[ht!]
\vspace*{\fill}
\centering
\caption{Faithfulness and completeness test scores for Iterative Patching.}
\begin{tabular}{llc@{\hspace{20pt}}ccccP{7px}cccc}
\toprule
\textbf{Op} & \textbf{CFG} & \textbf{CL $\downarrow$} & 
\multicolumn{4}{c}{\textbf{Faithful $\uparrow$}} & &
\multicolumn{2}{c}{\textbf{Complete $\downarrow$}}  & \textbf{Correct}\\
\cmidrule(lr){4-7} \cmidrule(l){9-10}
\; \textbf{BL} & & & {T1} & {T2} & {T3} & {LS} & {} & {T3} & {LS} & &\\
\midrule
\shortstack[l]{\textbf{Add}\\[-4pt]
\begin{tabular}[t]{@{}c@{}c@{}} 
\scriptsize T1: & \scriptsize \;0.93 \\[-6pt]
\scriptsize T2: & \scriptsize \;0.94 \\[-6pt]
\scriptsize LS: & \scriptsize \;0.92 
\end{tabular}}
& RF & 87 & 0.87 & 0.93 & 1.00 & 0.73 && 0.99 & 0.19 & \faTimes   \\[-14pt]
& RM & 94 & 0.87 & 0.93 & 1.00 & 0.83 && 0.99 & 0.19 & \faTimes   \\[+4pt]
\midrule
\multirow{5}{*}{\shortstack[l]{\textbf{Cos}\\[-4pt]
\begin{tabular}[t]{@{}c@{}c@{}} 
\scriptsize T1: & \scriptsize \;0.95 \\[-6pt]
\scriptsize T2: & \scriptsize \;0.97 \\[-6pt]
\scriptsize LS: & \scriptsize \;0.91 
\end{tabular}}}
 & MF & 75 & 0.87 & 0.94 & 0.98 & 0.46 && 0.00 & 0.00   & \faCheck \\
 & MM & 87 & 0.97 & 0.98 & 0.99 & 0.812 && 0.00 & 0.00   & \faCheck  \\
 & RF & 90 & 0.85 & 0.95 & 0.96 & 0.63 && 0.02 & 0.00   & \faCheck  \\
 & RM & 103 & 0.89 & 0.97 & 0.97 & 0.81 && 0.02 & 0.00   & \faTimes  \\
\midrule
\shortstack[l]{\textbf{Exp}\\[-4pt]
\begin{tabular}[t]{@{}c@{}c@{}} 
\scriptsize T1: & \scriptsize \;0.60 \\[-6pt]
\scriptsize T2: & \scriptsize \;0.75 \\[-6pt]
\scriptsize LS: & \scriptsize \;0.58 
\end{tabular}}
& MF & 75 & 0.52 & 0.64 & 0.86 & 0.40 && 0.00 & 0.00 &  \faTimes \\ [-14pt]
& MM & 86 & 0.55 & 0.63 & 0.72 & 0.47 && 0.00 & 0.00 &  \faTimes  \\
\midrule
\multirow{5}{*}{\shortstack[l]{\textbf{Log}\\[-4pt]
\begin{tabular}[t]{@{}c@{}c@{}} 
\scriptsize T1: & \scriptsize \;0.30 \\[-6pt]
\scriptsize T2: & \scriptsize \;0.62 \\[-6pt]
\scriptsize LS: & \scriptsize \;0.31 
\end{tabular}}}
&  MF & 87 & 0.24 & 0.47 & 0.94 & 0.22 && 0.00 & 0.00   & \faTimes  \\
 & MM & 76 & 0.24 & 0.30 & 0.95 & 0.22 && 0.12 & 0.00   & \faCheck  \\
& RF & 86 & 0.24 & 0.57 & 0.86 & 0.23 && 0.46 & 0.00 &  \faTimes \\
& RM & 77 & 0.23 & 0.44 & 0.80 & 0.21 && 0.53 & 0.00 &  \faTimes  \\
\midrule

\shortstack[l]{\textbf{Mul}\\[-4pt]
\begin{tabular}[t]{@{}c@{}c@{}} 
\scriptsize T1: & \scriptsize \;0.86 \\[-6pt]
\scriptsize T2: & \scriptsize \;0.97 \\[-6pt]
\scriptsize LS: & \scriptsize \;0.87 
\end{tabular}}
& RF & 92 & 0.73 & 0.96 & 0.99 & 0.66 && 1.00 & 0.19 &  \faTimes \\[-14pt]
& RM & 99 & 0.79 & 0.97 & 1.00 & 0.74 && 1.00 & 0.19 &  \faTimes \\[+4pt]
\midrule
\shortstack[l]{\textbf{Pow}\\[-4pt]
\begin{tabular}[t]{@{}c@{}c@{}} 
\scriptsize T1: & \scriptsize \;0.92 \\[-6pt]
\scriptsize T2: & \scriptsize \;0.95 \\[-6pt]
\scriptsize LS: & \scriptsize \;0.91 
\end{tabular}}
& MF & 99 & 0.87 & 0.93 & 0.96 & 0.49 && 1.00 & 0.00 & \faTimes   \\[-14pt]
 & MM & 110 & 0.90 & 0.96 & 0.99 & 0.83 && 1.00 & 0.00 & \faTimes   \\[+4pt]
\midrule
\multirow{5}{*}{\shortstack[l]{\textbf{Sin}\\[-4pt]
\begin{tabular}[t]{@{}c@{}c@{}} 
\scriptsize T1: & \scriptsize \;0.73 \\[-6pt]
\scriptsize T2: & \scriptsize \;0.83 \\[-6pt]
\scriptsize LS: & \scriptsize \;0.70
\end{tabular}}}
&  MF & 56 & 0.68 & 0.80 & 0.90 & 0.52 && 0.00 & 0.00 & \faTimes   \\
&  MM & 60 & 0.70 & 0.86 & 0.87 & 0.64 && 0.00 & 0.00 & \faCheck   \\
&  RF & 72 & 0.70 & 0.82 & 0.90 & 0.58 && 0.19 & 0.02 & \faTimes  \\
&  RM & 72 & 0.67 & 0.80 & 0.87 & 0.59 && 0.24 & 0.03 & \faTimes    \\
\midrule
\multirow{5}{*}{\shortstack[l]{\textbf{Tan}\\[-4pt]
\begin{tabular}[t]{@{}c@{}c@{}} 
\scriptsize T1: & \scriptsize \;0.50 \\[-6pt]
\scriptsize T2: & \scriptsize \;0.58 \\[-6pt]
\scriptsize LS: & \scriptsize \;0.47 
\end{tabular}}}
&  MF & 79 & 0.45 & 0.60 & 0.90 & 0.30 && 0.00 & 0.00 & \faTimes    \\
&  MM & 66 & 0.48 & 0.69 & 0.87 & 0.36 && 0.00 & 0.00 & \faTimes  \\
& RF & 80 & 0.44 & 0.64 & 0.88 & 0.33 && 0.08 & 0.01 & \faTimes    \\
& RM & 91 & 0.37 & 0.58 & 0.72 & 0.34 && 0.08 & 0.01 & \faTimes   \\
\bottomrule

\end{tabular}
\vspace*{\fill}
\label{tab:circuits IterativeVSPatchesComplete}
\end{table}

\newpage
\section{Probing: Additional Results and Hyperparameters}
\label{APP:Probing}

\begin{figure}[ht]
    \centering
    \includegraphics[width=1.00\linewidth]{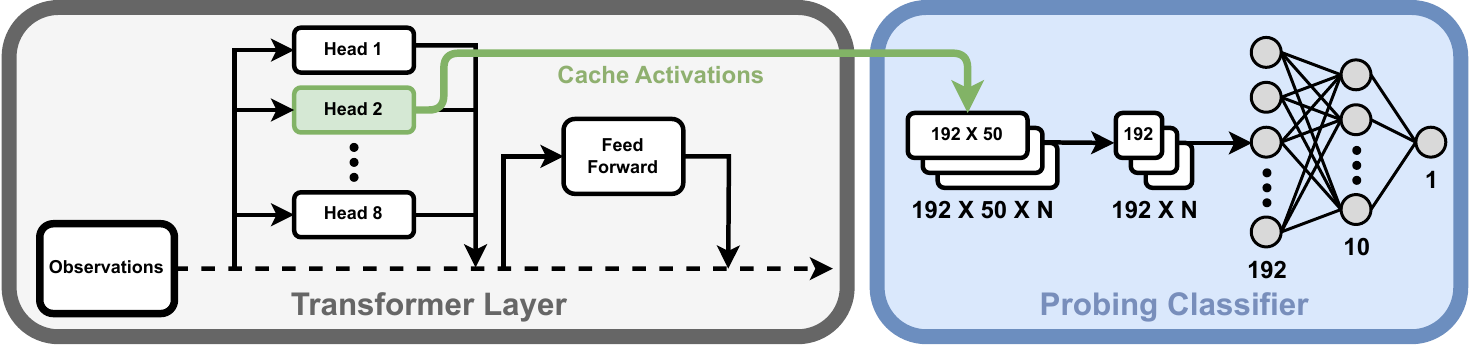}
    \caption{Probing process overview. Activations from a selected attention head (head 2) are cached (green arrow), mean-pooled over the induction dimension, and passed to a feedforward neural network (FFNN) to predict whether the equation contains the target element. The left panel shows a simplified transformer layer with residual flow (dashed arrow), and the right panel shows the probing classifier.}
    \label{fig:diagram_Probing}
\end{figure}

Our probing model \( P \) is implemented as a multilayer perceptron (MLP) with a single linear layer followed by a ReLu activation. The 64-dimensional input is projected to a hidden layer of 10 units and then to a single scalar output constrained with a sigmoid to be between \([0,1]\). Each attention head in the target model outputs activations of shape $(192,50)$, corresponding to 50 induction points per input and 64-dimensional representations per operation. To prepare the input $x_p$ for the probe $P(x_p)$, we aggregate these token-level activations into a single vector of shape $(192,)$ by applying mean pooling across the induction dimension (see \fullcref{fig:diagram_Probing}).

The model is trained on a balanced dataset of 1\,000 samples, with a stratified split of 70\% training, 10\% validation, and 20\% test data. We use a batch size of 32, a learning rate of $1\times10^{-4}$, and train for 200 epochs using the Adam optimizer and binary cross-entropy loss. The best-performing model on the validation set is selected for evaluation. Given that training takes approximately 12 seconds, no early stopping or patience scheduling was necessary.

To ensure that the probing classifier had enough data in training (without overfitting) we provide an exemplary loss and accuracy curve in \fullcref{fig:probing_curves}. The smooth and consistent trend in both curves suggests stable learning dynamics, with no visible signs of overfitting (e.g., training accuracy improving while validation accuracy degrades) or underfitting (e.g., both loss and accuracy stagnating at poor values).

\begin{figure}[ht!]
\centering
\includegraphics[width=0.99\linewidth]{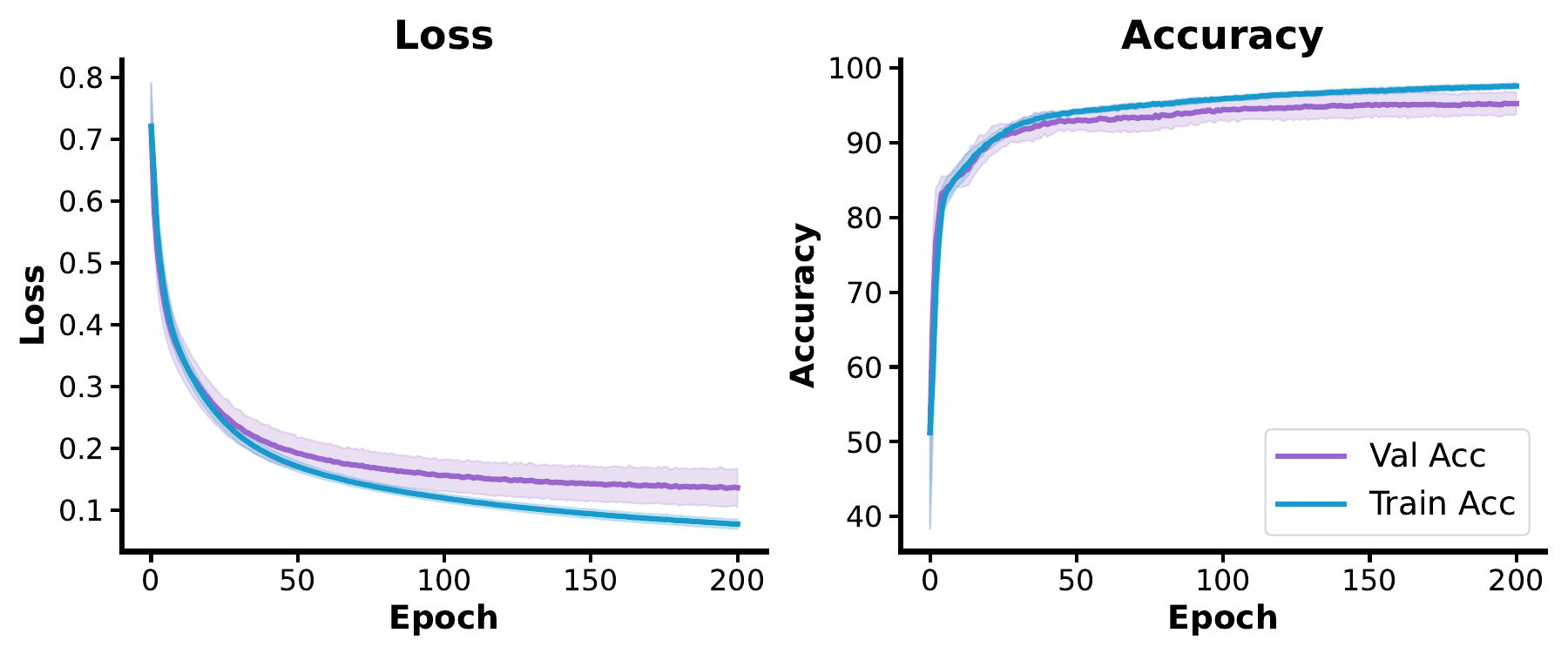}
\caption{Loss and accuracy curves of the probing classifier trained for 200 epochs.}
\label{fig:probing_curves}
\end{figure}

\begin{table}[ht]
\caption{Accuracy and statistical comparison between Circuit and Complement across different operations and setups. Bold are the higher mean accuracy scores out of every setup, green indicates a statistically significant difference between both.}
\centering
\small
\begin{tabular}{@{}cc
    cc
    >{\hspace{1px}}c<{\hspace{1px}}  
    cc
    c@{}}
\toprule
\textbf{Operation} & \textbf{Setup} & 
\multicolumn{2}{c}{\textbf{Circuit}} & &
\multicolumn{2}{c}{\textbf{Complement}} & 
\textbf{P-value $\downarrow$} \\
\cmidrule(lr){3-4} \cmidrule(lr){6-7}
 & & Mean $\uparrow$ & Std $\downarrow$ & & Mean $\uparrow$ & Std $\downarrow$ & \\
\midrule
\multirow{2}{*}{\textbf{Add}}  
& RF & \textbf{0.95} & 0.05 && 0.90 & 0.09 & 0.208 \\
& RM & \textbf{0.94} & 0.05 && 0.90 & 0.08 & 0.236 \\
\midrule
\multirow{5}{*}{\textbf{Cos}}  
& MF & \textbf{0.98} & 0.01 && 0.97 & 0.01 & 0.402 \\
& MM & \textbf{0.97} & 0.01 && 0.97 & 0.01 & 0.541 \\
& RF & \textbf{0.97} & 0.01 && 0.97 & 0.01 & 0.764 \\
& RM & \textbf{0.97} & 0.01 && 0.97 & 0.01 & 0.495 \\
& Sin & \textbf{0.97} & 0.01 && 0.97 & 0.01 & 0.532 \\
\midrule
\multirow{2}{*}{\textbf{Exp}}  
& MF & \textbf{0.96} & 0.02 && 0.95 & 0.02 & 0.186 \\
& MM & \textbf{0.96} & 0.02 && 0.95 & 0.02 & 0.617 \\
\midrule
\multirow{5}{*}{\textbf{Log}}  
& MF & \textbf{0.96} & 0.02 && 0.88 & 0.08 & \green{0.015} \\
& MM & \textbf{0.96} & 0.02 && 0.88 & 0.09 & \green{0.029} \\
& RF & \textbf{0.93} & 0.06 && 0.90 & 0.09 & 0.404 \\
& RM & \textbf{0.96} & 0.02 && 0.90 & 0.08 & 0.051 \\
& Exp & \textbf{0.90} & 0.07 && 0.90 & 0.08 & 0.938 \\
\midrule
\multirow{2}{*}{\textbf{Mul}}  
& RF & \textbf{0.92} & 0.03 && 0.90 & 0.04 & 0.206 \\
& RM & \textbf{0.91} & 0.04 && 0.89 & 0.04 & 0.189 \\
\midrule
\multirow{2}{*}{\textbf{Pow}}  
& MF & \textbf{0.98} & 0.02 && 0.98 & 0.01 & 0.482 \\
& MM & 0.98 & 0.02 && \textbf{0.98} & 0.01 & 0.628 \\
\midrule
\multirow{5}{*}{\textbf{Sin}}
& MF & \textbf{0.94} & 0.03 && 0.93 & 0.03 & 0.661 \\
& MM & \textbf{0.94} & 0.03 && 0.93 & 0.03 & 0.565 \\
& RF & \textbf{0.93} & 0.03 && 0.93 & 0.03 & 0.878 \\
& RM & 0.93 & 0.03 && \textbf{0.94} & 0.03 & 0.597 \\
& Cos & 0.93 & 0.03 && \textbf{0.94} & 0.03 & 0.766 \\
\midrule
\multirow{5}{*}{\textbf{Tan}}
& MF & 0.92 & 0.06 && \textbf{0.94} & 0.05 & 0.569 \\
& MM & \textbf{0.95} & 0.03 && 0.90 & 0.06 & 0.067 \\
& RF & \textbf{0.93} & 0.04 && 0.92 & 0.06 & 0.580 \\
& RM & \textbf{0.95} & 0.03 && 0.90 & 0.06 & 0.061 \\
& Sin & \textbf{0.93} & 0.04 && 0.92 & 0.06 & 0.580 \\
\bottomrule
\end{tabular}
\vspace*{\fill}
\label{tab:probing}
\end{table}

\newpage
\section{Direct Logit Attribution Additional Results}
\label{Direct Logit Attribution Additional Results}
To assess whether the findings in \fullcref{fig:DLAPaper} generalise across different patching types and operators, we report additional results in \fullcref{fig:DLAAPP}. These results confirm that the observed pattern is consistent: circuits constructed using direct logit attribution are larger than those identified using \patches\ or iterative patching techniques. This procedure is applied to 100 training samples for the different configurations to provide examples of both patching methods.

 We observe the same trends across operators, indicating that the tendency of direct logit attribution to produce overly large circuits is not specific to a particular setting. Consequently, irrespective of the operator or patching strategy employed, direct logit attribution appears to be a less suitable method for circuit discovery, as it systematically includes substantially more components than necessary. Moreover, it is unclear how an appropriate cut-off for such circuits should be defined. In this paper, we apply the same thresholding procedure used for \patches\ circuits to enable a fair comparison; however, this choice remains inherently arbitrary, and alternative thresholds could lead to markedly different circuit sizes.

\begin{figure*}[ht!]
    \centering
    \begin{subfigure}{0.49\textwidth}
        \centering
        \includegraphics[width=\linewidth]{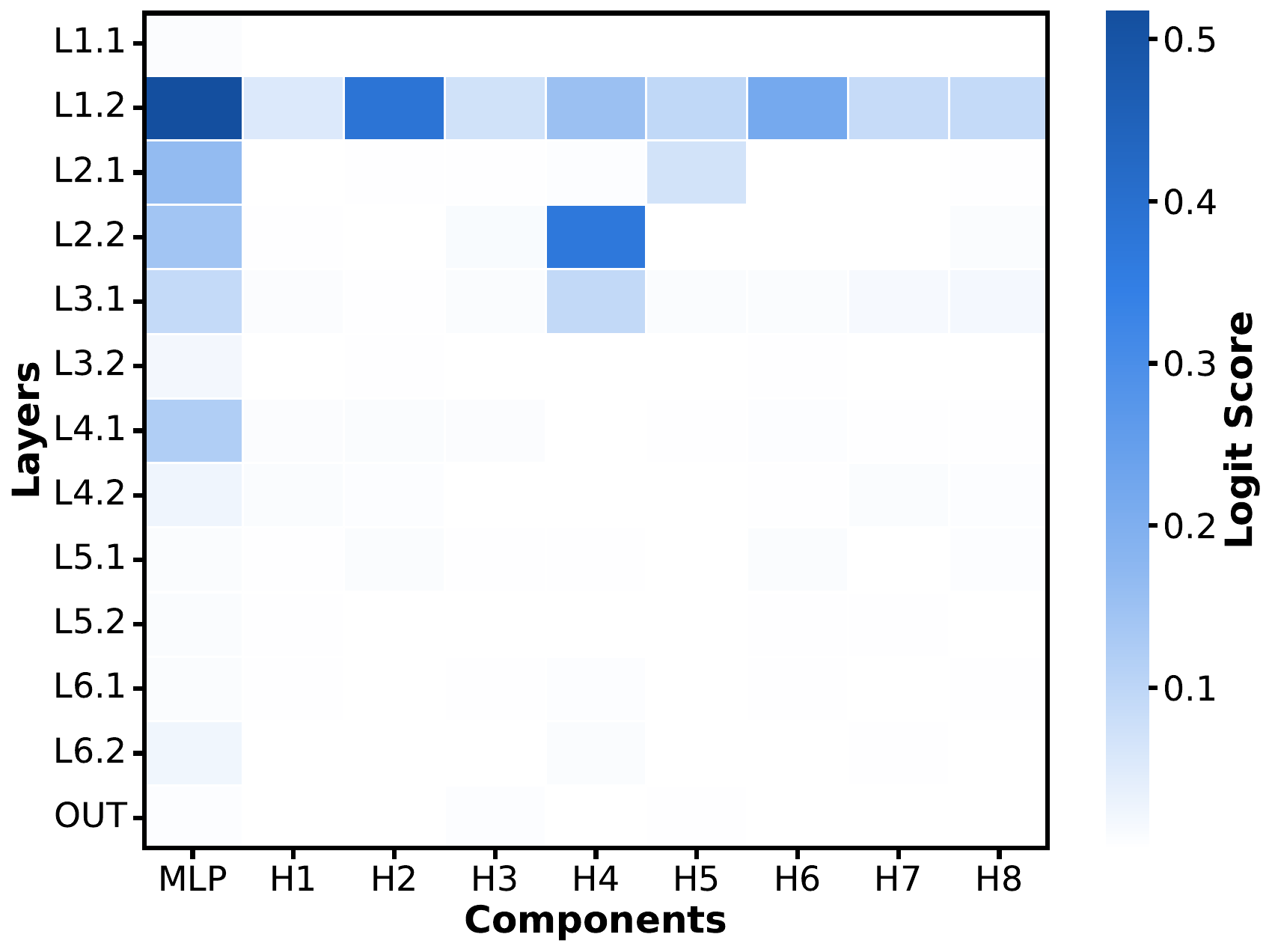}
        \caption{}
        \label{fig:DLA Sin resample attention}
    \end{subfigure}
    \begin{subfigure}{0.49\textwidth}
        \centering
        \includegraphics[width=\linewidth]{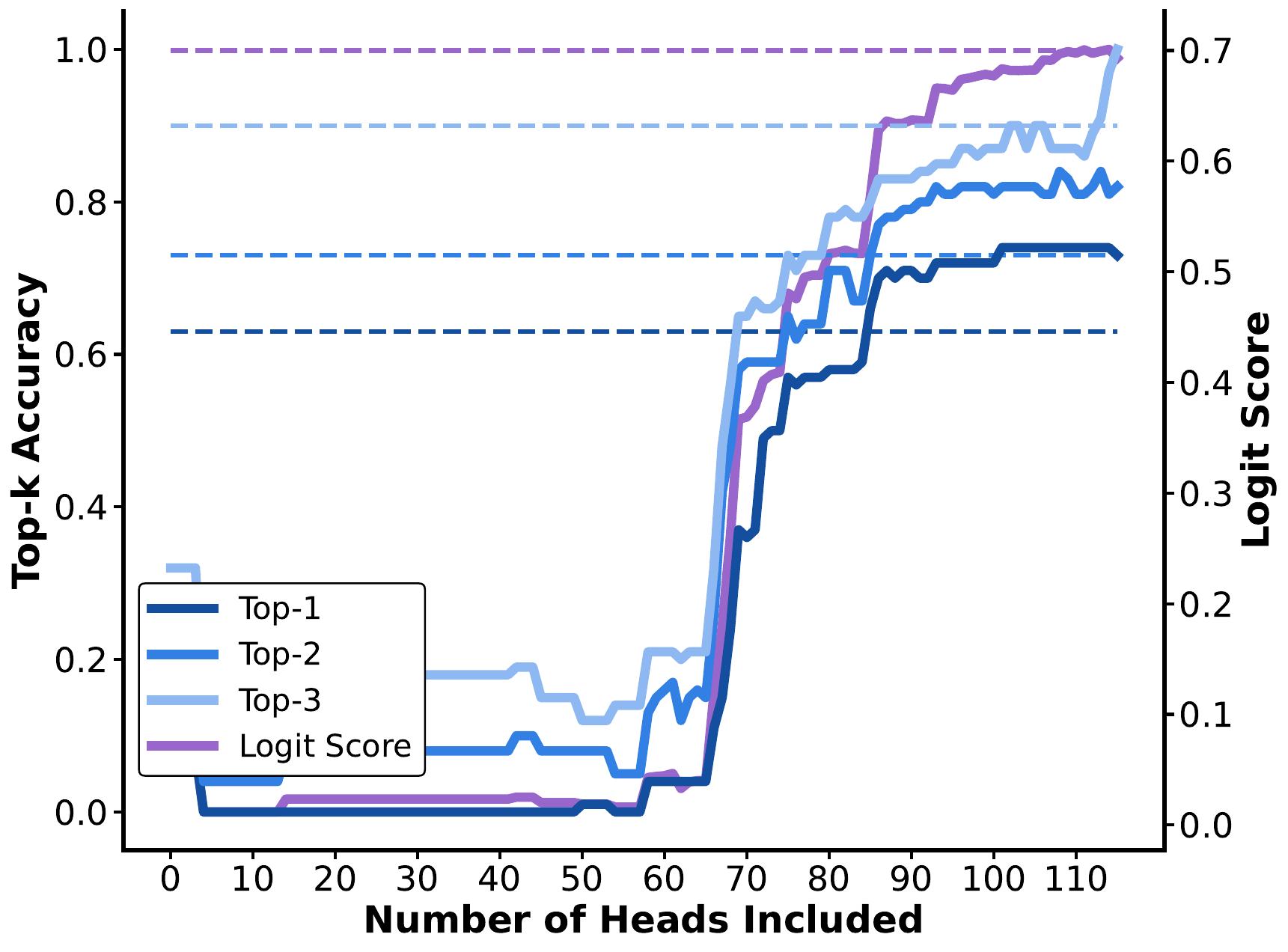}        
        \caption{}
         \label{fig:DLA Sin resample top}
    \end{subfigure}

    \begin{subfigure}{0.49\textwidth}
        \centering
        \includegraphics[width=\linewidth]{Pictures/DirectLogitAttribution/importance_exp_mean.pdf}
        \caption{}
        \label{fig:DLA exp Mean attention}
    \end{subfigure}
    \begin{subfigure}{0.49\textwidth}
        \centering
        \includegraphics[width=\linewidth]{Pictures/DirectLogitAttribution/topk_exp_mean.pdf}        
        \caption{}
         \label{fig:DLA exp mean top}
    \end{subfigure}

    \begin{subfigure}{0.49\textwidth}
        \centering
        \includegraphics[width=\linewidth]{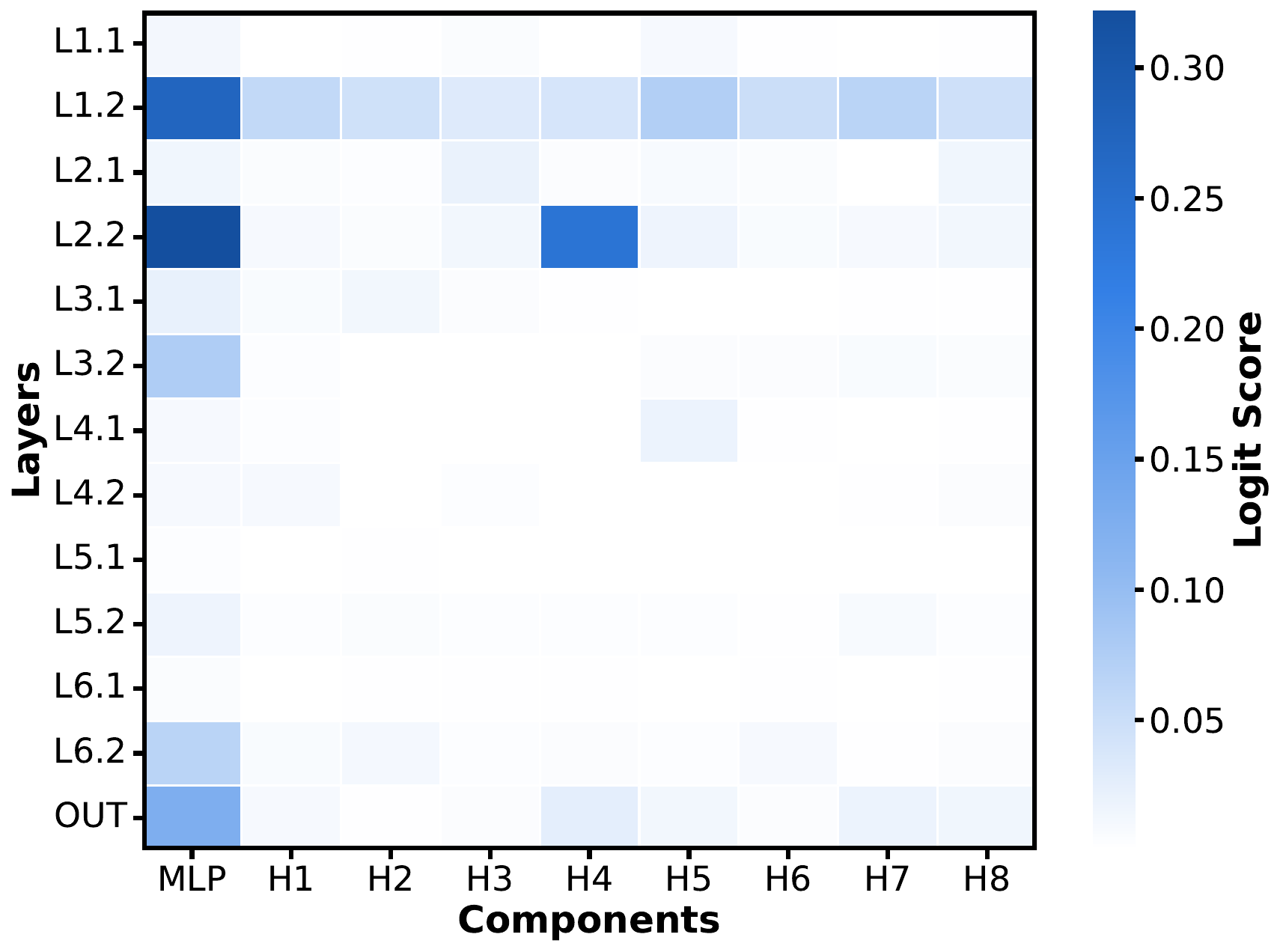}
        \caption{}
        \label{fig:DLA exp resample attention}
    \end{subfigure}
    \begin{subfigure}{0.49\textwidth}
        \centering
        \includegraphics[width=\linewidth]{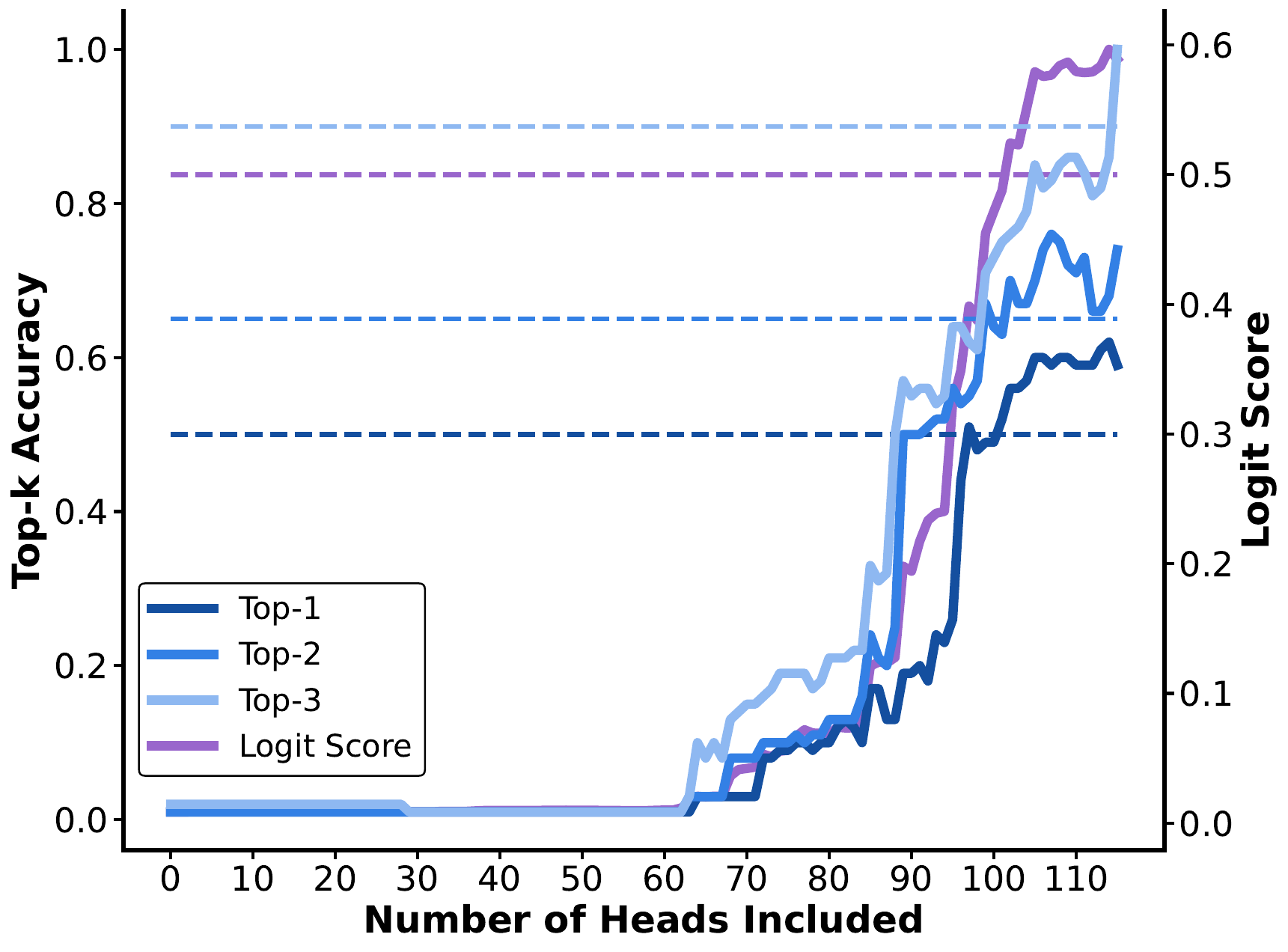}        
        \caption{}
         \label{fig:DLA exp resample top}
    \end{subfigure}
    \caption{\textbf{Direct attribution and faithfulness evaluation Additional Results}. \textit{(a,c,e)} Change in logit score when patching individual heads, averaged over 100 samples. \textit{(b, d, e)} Faithfulness evaluation of importance ranking; thresholds (dashed lines) from \fullcref{tab:baselines}. \textit{(a, b)} Sin Resample; \textit{(c, d)} Exp Mean, \textit{(e, f)} Exp Resample.}
\label{fig:DLAAPP}
\end{figure*}


\end{document}